\documentclass[10pt,twocolumn,letterpaper]{article}

\usepackage{iccv}
\usepackage{times}
\usepackage{epsfig}
\usepackage{graphicx}
\usepackage{amsmath}
\usepackage{amssymb}
\usepackage{enumerate}
\usepackage{enumitem}
\usepackage{tabularx}
\usepackage{makecell}
\usepackage[accsupp]{axessibility} 
\usepackage{times}
\usepackage{enumitem}
\usepackage{epsfig}
\usepackage{algorithm}
\usepackage{algorithmic}
\usepackage[caption=false]{subfig}
\usepackage[algo2e]{algorithm2e}
\usepackage{graphicx}
\usepackage{amsmath}
\usepackage{tabularx}
\usepackage{amssymb}
\usepackage{makecell}
\usepackage{enumerate}
\usepackage{booktabs}
\usepackage[usenames, dvipsnames]{color}
\usepackage{array}
\usepackage{makecell}
\usepackage[table]{xcolor}

\usepackage{url}            
\usepackage{booktabs}       
\usepackage{amsfonts} 
\usepackage{xr}
\usepackage{nicefrac}       
\usepackage{capt-of}
\usepackage{microtype}      
\usepackage[switch]{lineno}
\usepackage[caption=false]{subfig}
\usepackage{bm}

\usepackage{pifont}
\newcommand{\red}[1]{\textcolor{red}{#1}}
\usepackage[pagebackref=true,breaklinks=true,letterpaper=true,colorlinks,bookmarks=false]{hyperref}

\iccvfinalcopy 

\def\iccvPaperID{9984} 

\ificcvfinal\fi

\begin{document}

\title{PASS: Protected Attribute Suppression System for Mitigating Bias in Face Recognition}

\author{Prithviraj Dhar*\textsuperscript{1}, Joshua Gleason*\textsuperscript{2}, Aniket Roy\textsuperscript{1}, Carlos D. Castillo\textsuperscript{1}, Rama Chellappa\textsuperscript{1}\\
\textsuperscript{1}Johns Hopkins University,
\textsuperscript{2}University of Maryland, College Park\\
{\tt\small \{pdhar1,aroy28,carlosdc,rchella4\}@jhu.edu, gleason@umd.edu}
}

\maketitle
\ificcvfinal\thispagestyle{empty}\fi

\begin{abstract}
\vspace{-0.4cm}
   Face recognition networks encode information about sensitive attributes  while being trained for identity classification. Such encoding has two major issues: (a) it makes the face representations susceptible to privacy leakage (b) it appears to contribute to bias in face recognition. However, existing bias mitigation approaches generally require end-to-end training and are unable to achieve high verification accuracy. Therefore, we present a descriptor-based adversarial de-biasing approach called `Protected Attribute Suppression System (PASS)'. PASS can be trained on top of descriptors obtained from any previously trained high-performing network to classify identities and simultaneously reduce encoding of sensitive attributes. This eliminates the need for end-to-end training. As a component of PASS, we present a novel discriminator training strategy that discourages a network from encoding protected attribute information.  We show the efficacy of PASS to reduce gender and skintone information in descriptors from SOTA face recognition networks like Arcface. As a result, PASS descriptors outperform existing baselines in reducing gender and skintone bias on the IJB-C dataset, while maintaining a high verification accuracy.
   
\end{abstract}

\vspace{-0.25cm}
\section{Introduction}
{\let\thefootnote\relax\footnote{{*These authors have contributed equally to this work. }}}
\label{sec:intro}
Over the past few years, the accuracy of face recognition networks has significantly improved \cite{schroff2015facenet,taigman2014deepface,ranjan2019fast,deng2018arcface,bansal2018deep,dhar2019measuring}. These improvements have led to the deployment of face recognition systems in a large number of applications. However, recent studies \cite{dhar2020attributes, hill2019deep, terhorst2020beyond} have also shown that face recognition networks encode information about protected attributes such as race, gender, and age, while being trained for identity classification. Encoding of sensitive attributes raises concerns regarding privacy and bias. \\
\begin{figure}
\centering
{\includegraphics[width=\linewidth]{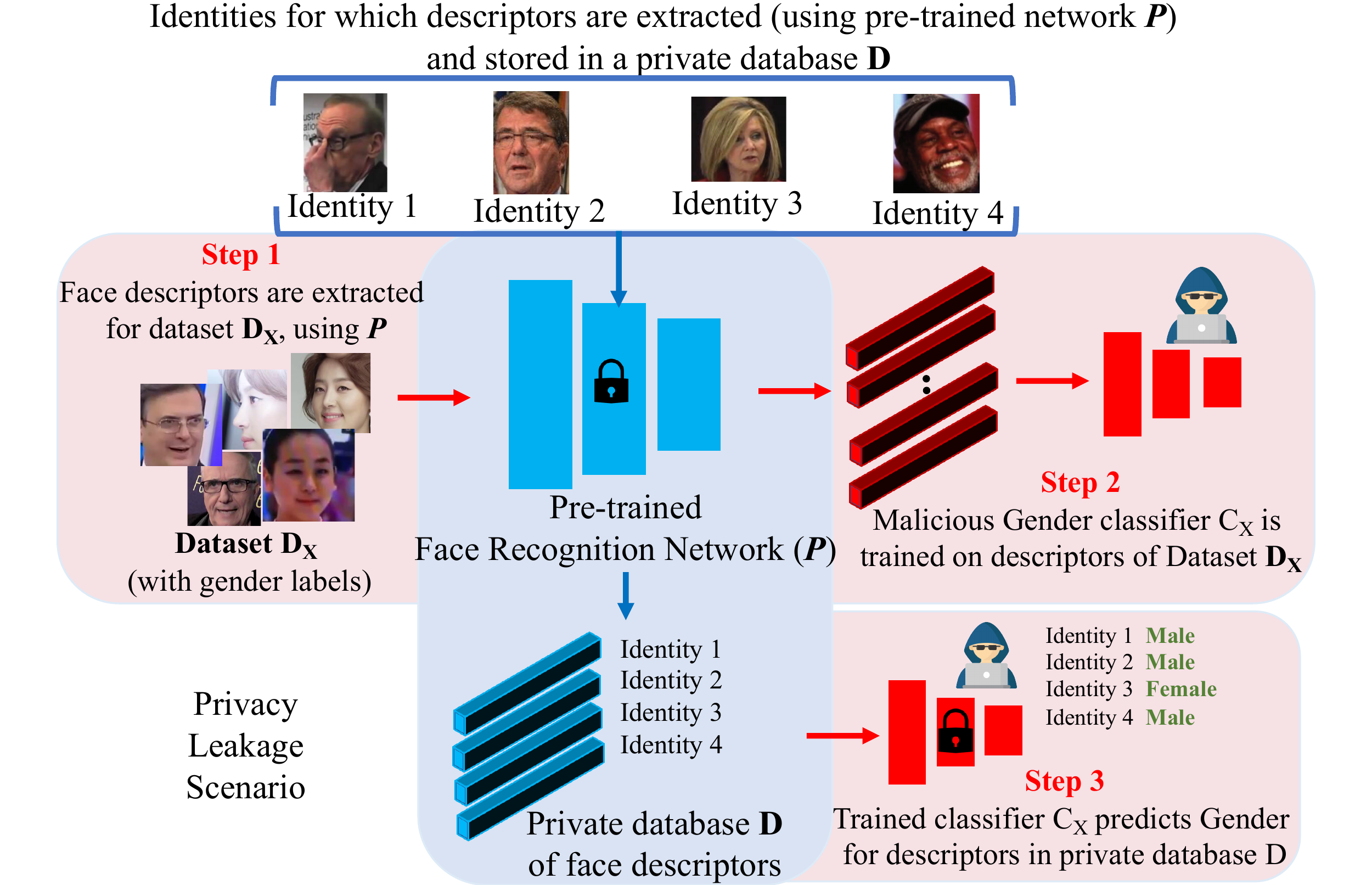}}
\caption{\small Suppose a malicious agent $X$ has gained access to a private database $D$ (blue) which consists of a pre-trained network $P$ and face descriptors of four identities. The agent can use $P$ to extract descriptors (red) for a gender-labeled dataset $D_X$ (Step 1). Using these descriptors, the agent can train a gender classifier $C_X$ (Step 2). Using the trained $C_X$, the agent can predict the gender of the descriptors in $D$ (Step 3) and thus cause privacy breach.}
\label{fig:teaser}
\vspace{-15pt}
\end{figure}
\textbf{Privacy concerns:}  Many large-scale face verification and identification systems employ a database that stores face descriptors of identities, as opposed to face images. Face descriptors refer to the features extracted from the penultimate layer of a previously trained face recognition network.
Storing descriptors, rather than images, allows for very fast gallery lookup and verification against known subjects. This also acts as an additional layer of security by not storing potentially sensitive information present in the original face images. However, since some sensitive information is still encoded in these descriptors (e.g. race, gender, age), a malicious agent with access to these descriptors can potentially extract this information and use it for nefarious purposes. An example scenario is presented in Figure~\ref{fig:teaser}. \\
\textbf{Bias concerns:} 
Encoding of protected attributes such as gender or race in face descriptors results in bias w.r.t. these attributes when used for face recognition. A recent study from NIST \cite{grother2019face} found evidence that characteristics such as gender and ethnicity impact verification and matching performance of face descriptors. Similarly, it has been shown that most face-based gender classifiers perform significantly better on male faces with light skintone than female faces with dark skintone~\cite{buolamwini2018gender}.

One method of addressing privacy and bias issues is by producing face descriptors that are independent of the protected attribute(s).
For instance, Debface \cite{gong2020jointly} proposes an end-to-end method for producing face descriptors that are disentangled from protected attributes using an adversarial approach.
Another common strategy for mitigating bias is to train face recognition systems using training datasets that are balanced in terms of sensitive attributes. However, building large datasets that are balanced in terms of the attributes we want to protect is difficult, expensive, and time-consuming. Moreover, once such a `fair' dataset is constructed, we still need to perform the costly operation of training a large recognition network from scratch. 

End-to-end training of a large-scale network requires access to a large dataset and computing power, and is time-consuming. Application of adversarial losses while training (as done in \cite{gong2020jointly}), also slows down the training process. Several works \cite{wu2018towards,bortolato2020learning,li2019deepobfuscator} show that reducing the information of sensitive attributes while training a network results in a drop in overall performance. Even if a new network is trained to generate attribute-agnostic face descriptors, we need to replace the existing network (say, $P$ in Fig \ref{fig:teaser}), and re-compute the descriptors for all the identities by feeding in the respective face images.

In this work, we propose a solution that addresses the following four points: (i) reduces the opportunity for leakage of protected attributes in face descriptors. (ii) mitigates bias with respect to multiple attributes (gender and skintone). (iii) operates on existing descriptors and does not require expensive end-to-end training. (iv) does not require a balanced training dataset.

The proposed method trains a lightweight model that transforms face descriptors obtained from an existing face recognition model, and maps them to an attribute agnostic representation. We achieve this using a novel adversarial training procedure called \textbf{P}rotected \textbf{A}ttribute \textbf{S}uppression \textbf{S}ystem (PASS). Unlike other works that adversarially suppress protected attributes~\cite{gong2020jointly, wu2018towards} using end-to-end training, we operate on descriptor space. Once trained, PASS may be easily applied to other existing face descriptors. In summary, we make the following contributions in this paper: 
\begin{enumerate}[leftmargin=*]
     \item We present PASS, an adversarial method that aims to reduce the information of sensitive attributes in face descriptors from any face recognition network, while maintaining high face verification performance. We show the efficacy of PASS to reduce gender and skintone information in face descriptors, and thus considerably reduce the associated biases. Moreover, PASS can be used on top of face descriptors obtained from \textit{any} face recognition network. We show these results on two SOTA pre-trained networks: Arcface~\cite{deng2018arcface} and Crystalface~\cite{ranjan2019fast}.
    \item Our descriptor-based model cannot include CNN-based discriminators, which poses new challenges. We present a novel discriminator training strategy in PASS, to enforce the removal of sensitive information in the descriptors.
    \item We extend PASS to reduce information of multiple attributes simultaneously, and show that such a framework (known as `MultiPASS') also performs well in terms of reducing the leakage of sensitive attributes and bias in face descriptors, while maintaining reasonable face verification performance.
    \item Since reducing the information of protected attributes in face descriptors also reduces their identity-classifying capability, we introduce a new metric called Bias Performance Coefficient (BPC), that measures the trade-off between bias reduction and drop in verification performance. We show that our PASS framework achieves better BPC values than existing baselines.
\end{enumerate}
\vspace{-0.2cm}
\section{Related work}
\vspace{-0.2cm}
\textbf{Bias in face recognition:} Several empirical studies \cite{grother2019face, buolamwini2018gender, drozdowski2020demographic} have shown that many publicly available face recognition systems demonstrate bias towards attributes such as race and gender. \cite{wang2019racial, wang2020mitigating, gac} highlight the issue of racial bias in face recognition, and propose strategies to mitigate the same. In the context of gender bias \cite{albiero2020does, lu2019experimental}, most experiments show that the performance of face recognition on females is lower than that of males. Use of cosmetics by females \cite{cook2019fixed, klare2012face} and gendered hairstyles \cite{albiero2020face} has been assumed to play a major role in the resulting gender bias. However, \cite{albiero2020analysis} shows that cosmetics only play a minor role in the gender gap. \cite{lu2019experimental} shows that face verification systems perform better on lighter skintones than darker skintones.\cite{albiero2020does, wang2019balanced} show that the gender bias is not mitigated even if the training dataset is gender-balanced. \cite{robinson2020face, wang2019racial} presents an evaluation datasets that is balanced in terms of race and provide the verification protocols for the same.\\
\textbf{Adversarial techniques to suppress attributes:}
\begin{table}
\centering
\scriptsize
\begin{tabular}{ccc}
\hline
Method & Target task & Sensitive attribute\\ 
\hline
\cite{zhang2018mitigating, madras2018learning}& Analogy completion &Gender\\
\cite{wang2019balanced} & Object classification & Gender \\
\cite{wu2018towards} & Action classification & Identity, private attributes \\
\cite{choi2019can}& Action recognition & Scene \\ 
\hline
\cite{alvi2018turning} & Gender/Age prediction & Age/Gender\\
\cite{gafni2019live}&Preserve pose/illumination/expresssion&Identity\\
\cite{li2019deepobfuscator} & Smile, high-cheekbones & Gender, make-up\\
\cite{amini2019uncovering} & Face detection & Skintone\\
\cite{quadrianto2019discovering} & Face attractiveness & Gender\\
\cite{wang2019racial} & Face recognition & Race\\
\cite{gong2020jointly}&Face recognition&Age,gender,race\\
\hline
PASS (Ours) & Face recognition & Gender, skintone\\
\end{tabular}
\caption{\small Methods that adversarially remove sensitive attributes in general vision/NLP tasks (top) and face-related tasks (bottom)}
\label{tab:rel}
\vspace{-10pt}
\end{table}
A summary of works that adversarially remove sensitive attributes, while performing a target task is provided in Table \ref{tab:rel}. Most of these works do not operate on descriptor space. Also, in some of the these experiments, the attribute under consideration is ephemeral to the target task. For example, in \cite{wu2018towards}, an action is not specific to an identity. In contrast, attributes like gender and race may not be ephemeral to face recognition. A given identity can be generally tied to a single gender/skintone. Because of the dependence between identity and gender/skintone, disentangling them is harder.\\
\textbf{Attribute privacy in face recognition:} \cite{mirjalili2018gender,mirjalili2017soft} introduce techniques to synthesize perturbed face images using an adversarial approach so that gender classifiers are confounded, but the performance of a commercial face-matcher is preserved. \cite{terhorst2019suppressing, bortolato2020learning, terhorst2019unsupervised} introduce techniques to suppress protected attributes like race, age and gender in face representations (as opposed to face images). However, the effect of such privacy preserving techniques on bias in face recognition is currently unclear.
\section{Problem Statement}
\vspace{-0.25cm}
\begin{figure}
{
\vspace{-10pt}
\centering
\subfloat[]{\includegraphics[width = 0.5\linewidth]{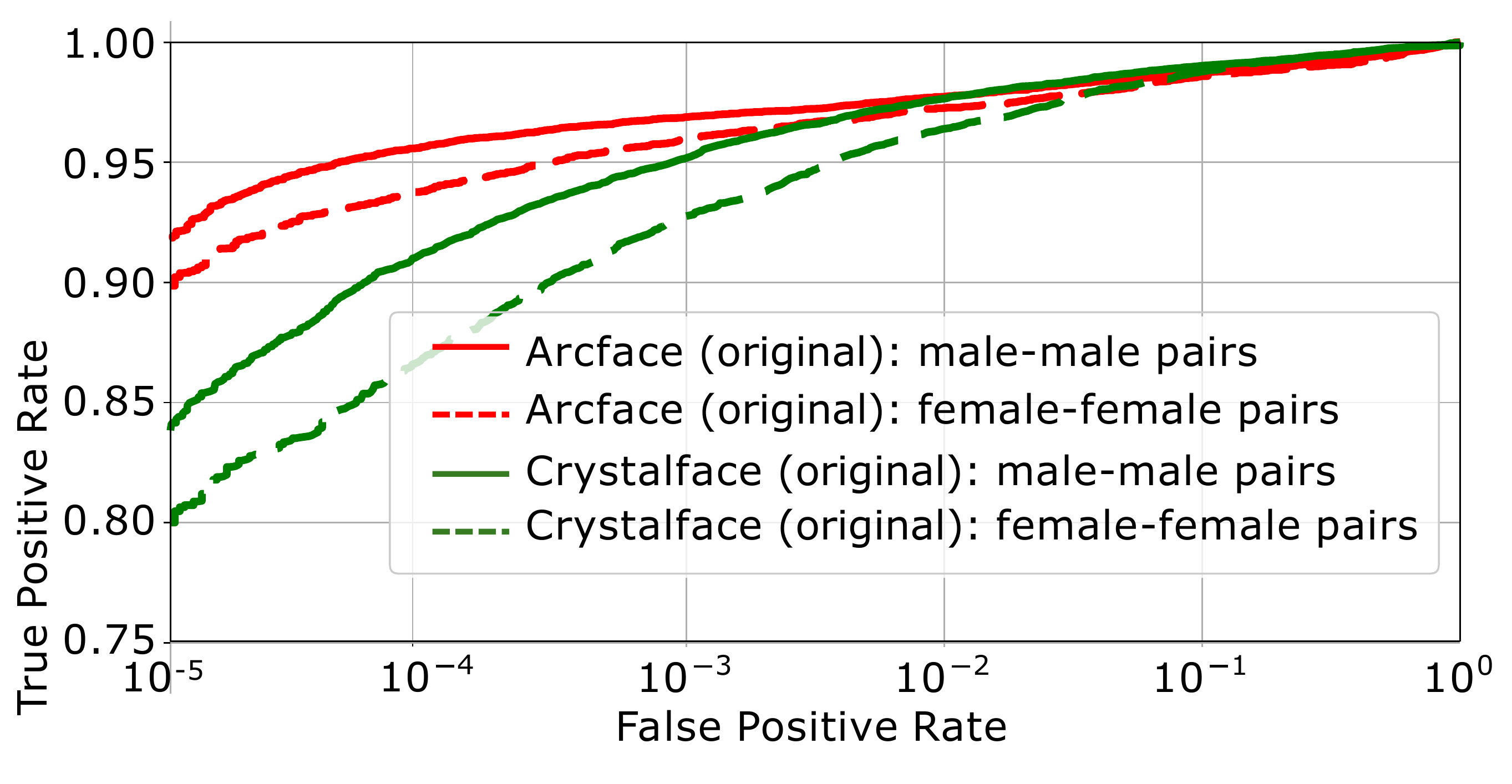}\label{fig:hypgen}}
\subfloat[]{\includegraphics[width = 0.5\linewidth]{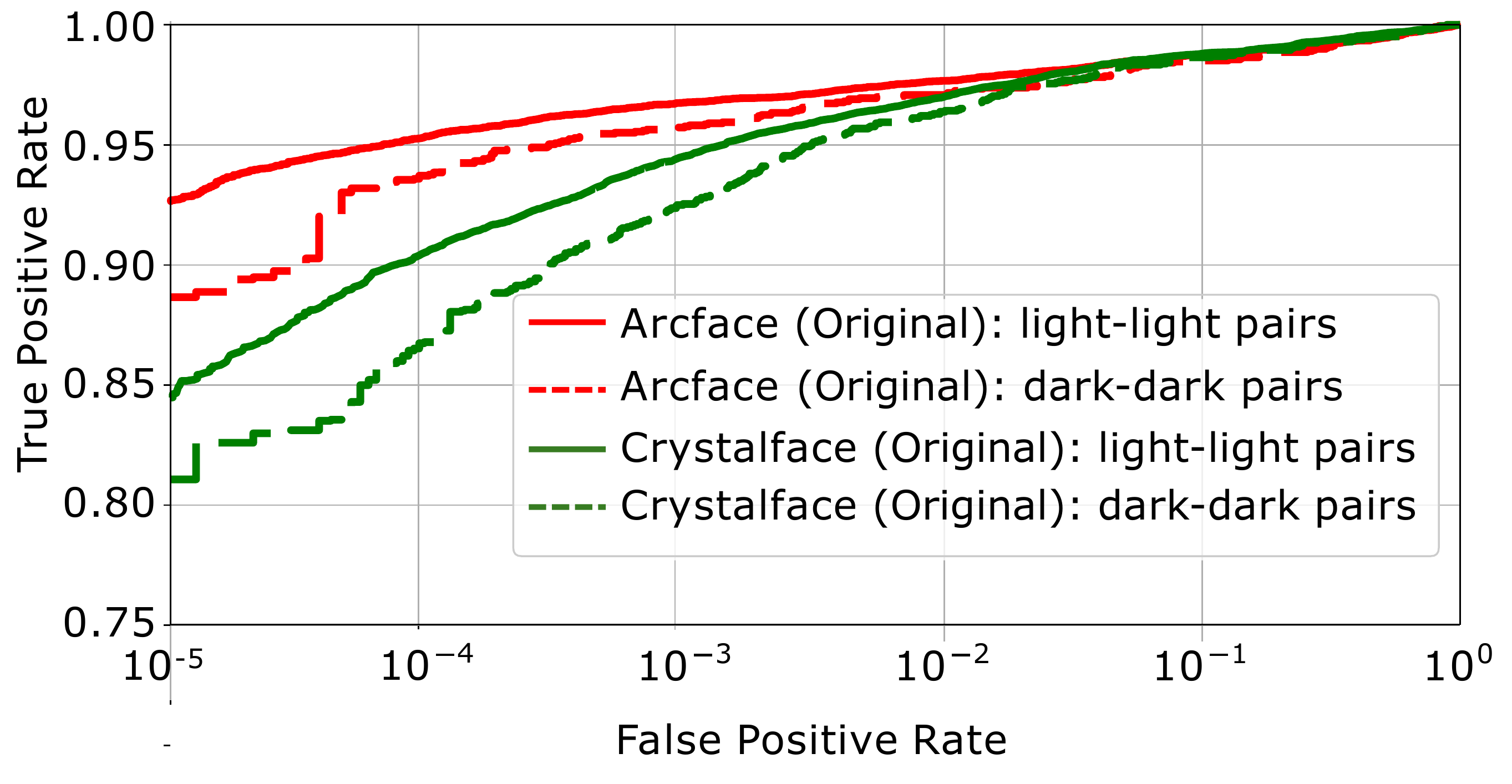}\label{fig:hypst}}
\caption{\small (a) Gender-wise and (b) Skintone-wise verification plot for Arcface and Crystalface networks, on IJB-C dataset. We define  bias as the difference between TPRs of males and females (or dark and light skintones) at a fixed FPR.}
\label{fig:cfafroc}
}
\vspace{-15pt}
\end{figure}

Our goal is to reduce gender and skintone information in face descriptors so that the ability of a classifier to predict gender and skintone from these descriptors is reduced. As an additional requirement, we constrain the gender and skintone-agnostic face descriptors to encode sufficient identity information, so that they can be used to perform face verification. We hypothesize that \textit{reducing the ability to predict protected attributes (gender and skintones) in face descriptors will reduce gender/skintone bias in face verification tasks}. This hypothesis is built on the results of \cite{gong2020jointly}, which shows that adversarially removing sensitive information from face representations reduces bias. However, unlike \cite{gong2020jointly}, we approach the problem in descriptor space.\\
\textbf{Mesuring bias:} At this point, we quantitatively describe gender and skintone bias in the context of face verification. Most work on face verification \cite{deng2018arcface, ranjan2019fast, liu2017sphereface,dhar2019measuring} report performance of a system by using an ROC (TPR vs FPR) curve, similar to Fig \ref{fig:cfafroc}. Hence, we define gender and skintone bias, at a given false positive rate (FPR) as follows: \vspace{-6pt}
\begin{equation}
    \text{Gender Bias}^{(F)} = |\text{TPR}_{m}^{(F)} - \text{TPR}_{f}^{(F)}|
\label{eq:gbias}\vspace{-15pt}
\end{equation}
\begin{equation}
    \text{Skintone Bias}^{(F)} = |\text{TPR}_{l}^{(F)} - \text{TPR}_{d}^{(F)}|
\label{eq:stbias}\vspace{-6pt}
\end{equation}
where $(\text{TPR}_{m}^{(F)}, \text{TPR}_{f}^{(F)}, \text{TPR}_{l}^{(F)}, \text{TPR}_{d}^{(F)})$ denote the true positive rates for the verification of male-male, female-female, light-light and dark-dark pairs respectively at FPR $F$. In some works such as \cite{gong2020jointly}, bias is evaluated as the difference between area under ROC curves (AUC). While this can be viewed as an aggregate of our measure, such an aggregation fails to meaningfully capture the bias at realistic operating points as it marginalizes the performance at low FPR. In our experience, most real world verification systems tend to operate at very low FPR, i.e. less than $10^{-4}$, which is not meaningfully captured with AUC. In this work, we focus on FPR values that we consider to be realistic operating conditions.\\
\textbf{Measuring bias/performance trade-off:}
Several methods that reduce the information of sensitive attributes in images or representations demonstrate a slight drop in overall performance of the system \cite{bortolato2020learning, li2019deepobfuscator, wu2018towards, 8869910}. So, reducing gender/skintone information in descriptors for de-biasing may lead to a slight drop in face verification performance. Inspired by the metric in \cite{bortolato2020learning}, we introduce a new metric called bias performance coefficient (BPC) to measure the trade-off between bias reduction and drop in verification performance. 
\vspace{-0.7cm}
\begin{equation}
\label{eq:bpc}
\text{BPC}^{(F)}= \frac{\text{Bias}^{(F)}-\text{Bias}^{(F)}_{deb} }{\text{Bias}^{(F)}}-\frac{\text{TPR}^{(F)} - \text{TPR}^{(F)}_{deb} }{\text{TPR}^{(F)}}
\vspace{-0.1cm}
\end{equation}
 Here, ($\text{TPR}^{(F)},\text{Bias}^{(F)}$) refer to the \textit{overall} TPR obtained by original descriptors and the corresponding bias (Gender/Skintone bias) at FPR of $F$.  ($\text{TPR}^{(F)}_{deb}, \text{Bias}^{(F)}_{deb}$) denote their de-biased counterparts. We prefer an algorithm that obtains higher BPC since a higher BPC denotes high bias reduction and low drop in verification performance. The original face descriptors (without any de-biasing) would have a zero BPC (since $\text{Bias}^{(F)}=\text{Bias}^{(F)}_{deb}$ and $\text{TPR}^{(F)}=\text{TPR}^{(F)}_{deb}$). Note that a negative BPC denotes that the percentage drop in TPR is higher than the percentage reduction in bias. In our work, we denote the BPC for skintone as `BPC\textsubscript{st}' and that for gender as `BPC\textsubscript{g}'. In summary, we aim \textit{to build systems that achieve high BPC values}.
\begin{figure}
\centering
{\includegraphics[width=\linewidth]{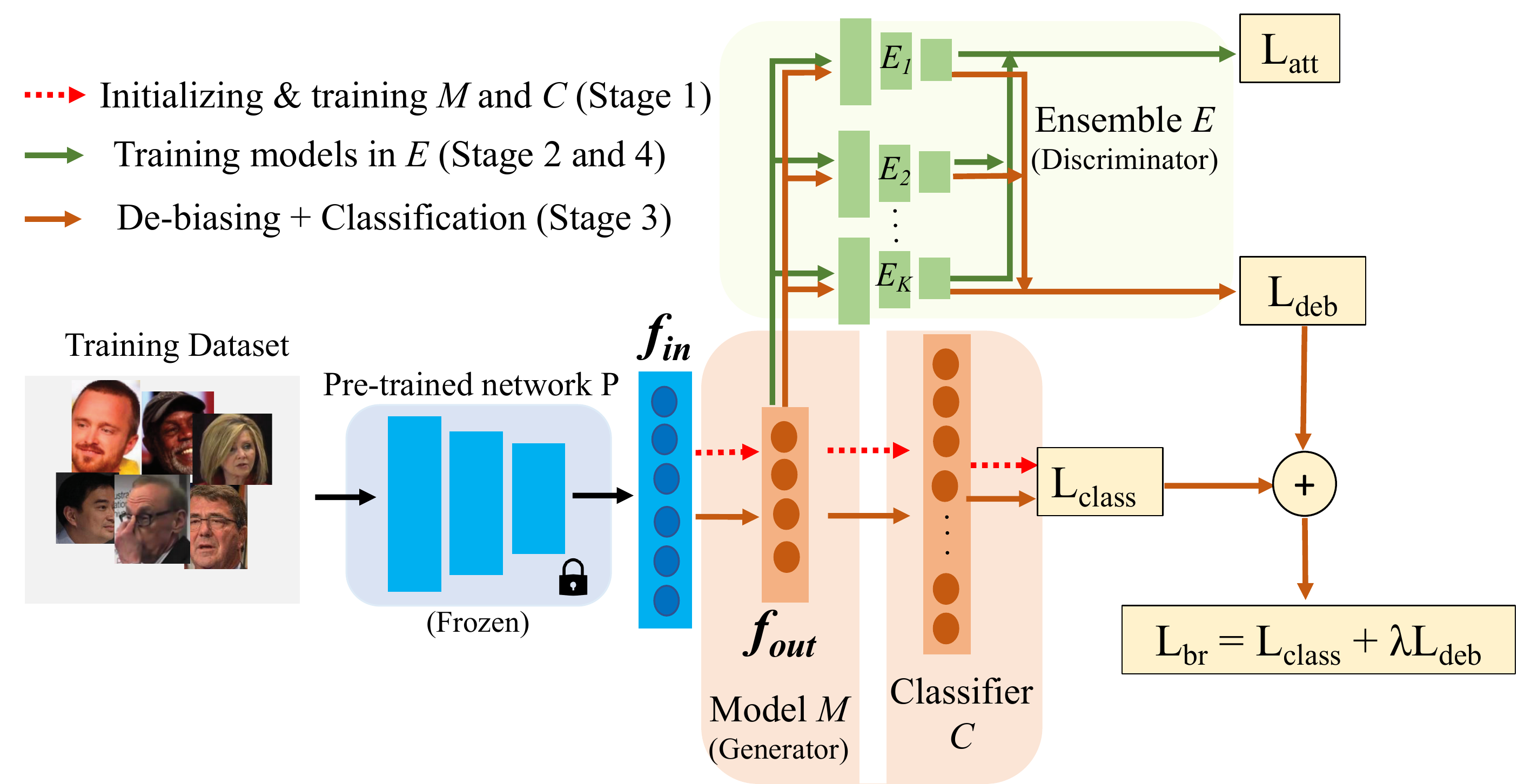}}
\caption{\small \textbf{PASS architecture}. Face descriptors $f_{in}$ are extracted from a previously trained network $P$ and are fed to a model $M$. $M$ consists of a single linear layer with PReLU activation that outputs transformed face descriptor $f_{out}$. This is then fed to classifier $C$ and ensemble $E$. The arrows indicate the dataflow at various training stages. In stage 1, $M$ and $C$ are initialized and trained to classify identity using the gradients of $L_{class}$. In stage 2, $E$ is initialized and trained to classify attribute using gradients of $L_{att}$. In stage 3, $M$ and $C$ are trained using the gradients of $L_{br}$ to debias  $f_{out}$ with respect to the target attribute, while simultaneously being able to classify identity. In stage 4, one member of ensemble $E$ is trained to classify attribute from $f_{out}$ using the gradients of $L_{att}$. Stages 3 and 4 are repeated in alternating fashion, where the ensemble member of $E$ being trained in stage 4 changes at each iteration.}
\vspace{-0.4cm}
\label{fig:pipeline}
\end{figure}
\section{Proposed Approach}
\vspace{-0.1cm}
\label{sec:approach}
\subsection{PASS}
\vspace{-0.1cm}
The key idea in our proposed approach - PASS, is to train a model to classify identities while discouraging it from predicting a specific protected attribute.  Firstly, for a given image $I$, we extract a face descriptor $f_{in}$ using a pre-trained network $P$.
\vspace{-4pt}
\begin{equation}
    f_{in} = P(I)
\end{equation}

\vspace{-6pt}
We present the PASS architecture in Fig. \ref{fig:pipeline}. This architecture is inspired by the adversarial framework in \cite{wu2018towards}. PASS is composed of three components:\\
(1) \textbf{Generator model $M$}: A model that accepts face descriptor $f_{in}$ from a pre-trained network $P$, and generates a lower dimensional descriptor $f_{out} \in \mathbb{R}^{256}$. $M$ consists of a single linear layer with 256 units, followed by a PReLU \cite{he2015deep} layer. The weights of $M$ are denoted as $\phi_M$. \\
(2) \textbf{Classifier} $C$: A classifier that takes in $f_{out}$ and generates a prediction vector for identity classification. The weights of $C$ are denoted as $\phi_C$.\\
(3) \textbf{Ensemble of attribute classifiers $E$}: An ensemble of $K$ attribute prediction models represented as $E_1, E_2 \ldots E_K$ that take $f_{out}$ as input. Each of these models is a two layer MLP with 128 and 64  hidden units respectively with SELU activations, followed by a sigmoid activated output layer with $N_{att}$ units. Here, $N_{att}$ denotes the number of classes in the attribute being considered. We collectively denote the weights of all the models in $E$ as $\phi_E$ and weights of $k^{th}$ model $E_k$ as $\phi_{E_k}$. Note that the attribute classifiers in $E$ are simple MLP networks (and not CNNs as used in \cite{wu2018towards}). This is because the input to $E$  are low-dimensional descriptors $f_{out}$ and not images.

We now explain PASS as an adversarial approach. $M$ can be viewed as a generator that should generate descriptors $f_{out}$ that are agnostic to the attribute under consideration. $f_{out}$ is fed to the ensemble $E$ of attribute prediction models which acts as a discriminator and tries to predict the protected attribute. The objective of $M$ is to generate descriptors $f_{out}$ that can fool $E$ in terms of attribute prediction, and can also be used to classify identities. Therefore, we impose two constraints on $f_{out}$: (i) a penalty for misidentification, and (ii) a penalty for attribute predictability from $f_{out}$. To this end, we propose a bias reducing classification loss $L_{br}$ described in section~\ref{sec:bias}.
\vspace{-0.3cm}
\subsubsection{Bias reducing classification loss $\bm{L_{br}}$}
\vspace{-0.3cm}
\label{sec:bias}
\setlength{\belowdisplayskip}{1pt}
\setlength{\abovedisplayskip}{1.2pt}
After extracting the descriptor $f_{in}$ from a pre-trained face recognition network, we pass it through $M$ to obtain a lower dimensional descriptor $f_{out}$.
\begin{equation}
    f_{out} = M(f_{in}, \phi_M)
\end{equation}
\textbf{First constraint:} To make $f_{out}$ proficient at classifying identities, we provide it to classifier $C$ and use cross-entropy classification loss $L_{class}$ to train both $C$ and $M$.
\begin{equation}
\label{eq:lclass}
    L_{class}(\phi_M, \phi_C) = -\mathbf{y_{id}} . \text{log}(C(f_{out}, \phi_C))
\end{equation}
$\mathbf{y_{id}}$ is a one hot identity label and classifier $C$ produces softmaxed outputs.
\\
\textbf{Training discriminators:} $M$ generates $f_{out}$ which is fed to ensemble $E$. Each of the attribute prediction models in $E$, denoted as $E_k$, is used for computing the cross entropy loss $L^{(E_k)}_{att}$ for attribute classification. $L_{att}$ is computed as the sum of cross-entropy losses for each $E_k$.
\begin{equation}
\label{eq:gc}
L_{att}(\phi_M, \phi_E)=-\sum^{K}_{k=1}\sum_{i=1}^{N_{att}} y_{att,i} \text{log } y_{att,i}^{(k)}    
\end{equation}
$y_{att,i}$ is the binary attribute label for the $i^{th}$ attribute category associated with the input face descriptor, and $y^{(k)}_{att,i}$ represents the respective softmaxed outputs of $E_k$ in the ensemble. $N_{att}$ denotes the number of categories associated with the attribute under consideration. \\ 
\textbf{Training generator (second constraint):} After training $E$, $M$ is trained to transform $f_{in}$ into attribute-agnostic descriptor $f_{out}$. We then provide $f_{out}$ to each model in $E$:
\begin{equation}
    o_{k} = E_{k}(f_{out}, \phi_{E_{k}})~~\text{for}~ k=1\ldots K
    \label{eq:ok}
\end{equation}
The outputs $o_k$ are $N_{att}$-dimensional and represent the probability scores for different categories associated with the attribute. We refer to the $i^{th}$ element of $o_k$ as $o_{k,i}$.

If an optimal classifier operating on $f_{out}$ were to always produce a posterior probability of $\frac{1}{N_{att}}$ for all categories in the attribute, then this implies that no attribute information is present in the descriptor. To this end, we define the adversarial loss $L^{(E_{k})}_{adv}$ for the $k^{th}$ model in $E$ to be:
\begin{equation}
\label{eq:adv}
    L^{(E_{k})}_{adv}(\phi_M,\phi_{E_{k}}) = -\sum_{i=1}^{N_{att}}\frac{1}{N_{att}}\text{log}(o_{k,i})
\end{equation}
Here, we use an ensemble of attribute prediction models, rather than a single model because, we want $f_{out}$ to be constructed such that \textit{no} model can predict the protected attribute. This approach was motivated by the work of \cite{wu2018towards} to solve `the $\forall$ challenge'.
After computing the adversarial loss for model $M$ with respect to all the models in $E$, we select the one for which the loss is maximum. We term this loss as debiasing loss $L_{deb}$.
\begin{equation}
\label{eq:ldeb}
    L_{deb}(\phi_M, \phi_E) = \text{max}\{L^{(E_{k})}_{adv}(\phi_M, \phi_{E_k}) |^{K}_{k=1} \}
\end{equation}
This loss function penalizes $M$ with respect to the strongest attribute predictor which it was not able to fool. This approach was introduced in \cite{wu2018towards}.  $L_{deb}$ is then combined with $L_{class}$ to compute a bias reducing classification loss $L_{br}$.
\begin{equation}
\label{eq:lbr}
    L_{br}(\phi_C,\phi_M,\phi_E) = L_{class}(\phi_C, \phi_M)+\lambda L_{deb}(\phi_M, \phi_E)
\end{equation}

\vspace{-0.30cm} \hspace{-15pt}
Here, $\lambda$ is used to weight the de-biasing loss.
\vspace{-0.4cm}
\subsubsection{Stage-wise Training}
\vspace{-0.2cm}
\label{sec:stagewise}
We now discuss the various stages of training PASS. \\
\textbf{Stage 1 - Initializing and training $M$ and $C$}: Using input descriptors $f_{in}$ from a pre-trained network, we train $M$ and $C$ from scratch for $T_{fc}$ iterations using $L_{class}$ (Eq. \ref{eq:lclass}). \\
\textbf{Stage 2 - Initializing and training $E$}: Once $M$ is trained to perform classification, we feed the outputs $f_{out}$ of $M$ to an ensemble $E$ of $K$ attribute prediction models. $E$ is trained from scratch to classify attribute for $T_{atrain}$ iterations using $L_{att}$ (Eq. \ref{eq:gc}). $\phi_M, \phi_C$ remain unchanged in this stage.\\ 
\textbf{Stage 3 - Update model $M$ and classifier $C$}: Here, $M$ is trained to generate descriptors $f_{out}$ that are proficient in classifying identities and are relatively attribute-agnostic. $f_{out}$ is fed to the ensemble $E$ and the classifier $C$, the outputs of which result in $L_{deb}$ (Eq. \ref{eq:ldeb}) and $L_{class}$ (Eq. \ref{eq:lclass}) respectively. We combine them to compute $L_{br}$ (Eq. \ref{eq:lbr}) for training $M$ and $C$ for $T_{deb}$ iterations, while $\phi_E$ remains locked.  While computing $L_{br}$, the gradient updates for $L_{deb}$ are propagated to $\phi_M$ and those for $L_{class}$ are propagated to $\phi_M$ and $\phi_C$.\\
\textbf{Stage 4 - Update ensemble $E$ (discriminator)}:
In stage 4, members of $E$ are trained to classify attribute using $f_{out}$. Therefore, we run stages 3 and 4 alternatively, for $T_{ep}$ episodes, after which we re-initialize and re-train all the models in $E$ (as done in stage 2). This re-initialization follows from \cite{wu2018towards}, in order to prevent trivial overfitting between $M$ and $E$. Here, one episode indicates an instance of running stages 3 and 4 consecutively. In stage 4, we choose one of the models in $E$, and train it for $T_{plat}$ iterations or until it reaches an accuracy of $A^*$ on the validation set. $\phi_M$ and $\phi_C$ remain locked in this stage. The detailed PASS algorithm is provided in the supplementary material.
\begin{figure}
\centering
{\includegraphics[width=\linewidth]{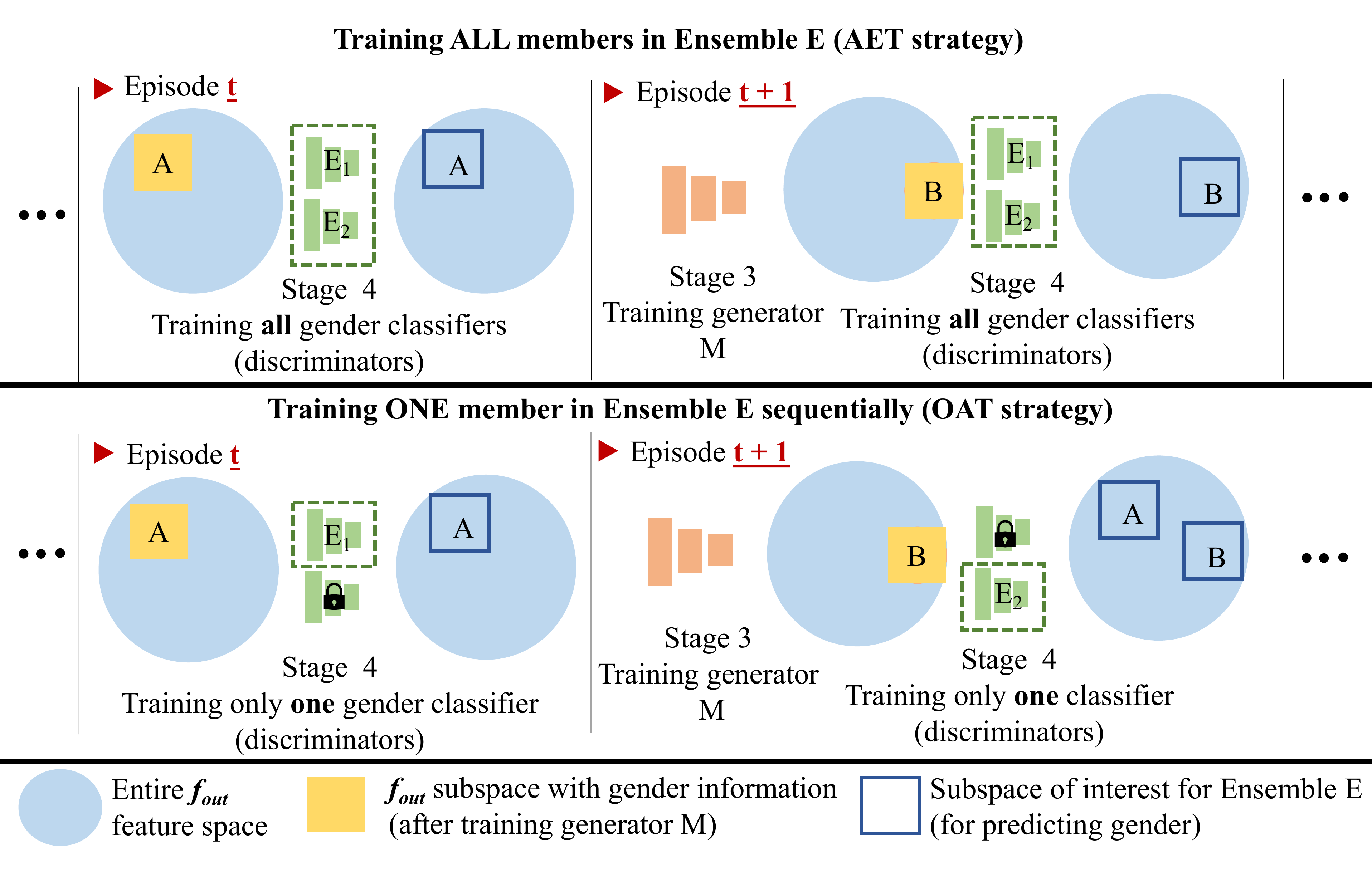}}
\caption{\small Descriptor space for AET (top) versus OAT (bottom) strategies (example using 2 member ensemble). Using OAT, $M$ is more restricted in how it may represent protected attribute information in descriptor space, encouraging it to instead remove information about the protected attribute all-together.} 
\label{fig:oat_aet_eg}
\vspace{-14pt}
\end{figure}
\vspace{-0.4cm}
\subsubsection{One-At-a-time (OAT) vs All-Every-Time (AET)}
\label{sec:oat_vs_aet}
\vspace{-0.2cm}
We note that the method on which PASS is based \cite{wu2018towards}, trains all the discriminators during stage 4 training. We call this `All-Every-Time (AET)' strategy. However, in this section we present a conceptual argument describing how AET could produce descriptors that still contain sensitive information. The key ideas of this argument are visualized in Fig~\ref{fig:oat_aet_eg}.

Consider the case where PASS consists of an ensemble $E$ with two gender classifiers, and suppose that model $M$ has distilled all gender information into a subspace, $A$, of descriptor space after stage 3 of episode $t$. Following the AET strategy, all classifiers in $E$ are trained to classify gender, thus, encouraging them to focus on subspace $A$. In episode $t+1$, suppose $M$ re-organizes the descriptor space to distill gender information into a new subspace $B$ (orthogonal to $A$) in order to fool the classifiers in $E$.  In stage 4 of episode $t+1$, all the gender classifiers will then be trained again to extract gender information, causing them to focus on subspace $B$ and forget subspace $A$. Thus, in stage $t+2$, $M$ could revert to its episode $t$ state, once again distilling gender information back into subspace $A$ without penalty.


To address this issue, we propose a novel discriminator training strategy that we call `\textbf{O}ne-\textbf{A}t-a-\textbf{T}ime (OAT)', where, during stage 4 we train one member in $E$, and freeze the rest. Using the same example from Fig~\ref{fig:oat_aet_eg} (bottom row), we describe how this encourages $M$ to remove gender.

As before, suppose that after stage 3 of episode $t$, $M$ has distilled all gender information into subpace $A$. However, unlike in the AET example, suppose only member $E_1$ of ensemble $E$ is trained during stage 4. In stage 3 of episode $t+1$, suppose $M$ again distills gender information into subspace $B$. During stage 4 of episode $t+1$, $E_2$ is trained, and the weights of $E_1$ are held constant. Thus, after 2 episodes the prediction of ensemble $E$ depends on both subspace $A$ and $B$ (since $E_1$ is still dependent on subspace $A$). Our conclusion is that this strategy restricts $M$ from reverting back to its episode $t$ state after stage 3 of episode $t+2$, thus improving  the chance that $M$ removes gender information all-together.



For the PASS architecture with $K$ classifiers in ensemble $E$, at episode $i$, we train the $j^{th}$ classifier in the ensemble, where $j=i$ mod $K$, and freeze the rest (thus sequentially choosing one discriminator). We conduct experiments to compare OAT and AET (in Section \ref{sec:novcomp}) and show that OAT leads to better attribute-removal as compared to AET.

\vspace{-0.15cm}
\subsection{MultiPASS}
\begin{figure}
\centering
{\includegraphics[width=\linewidth]{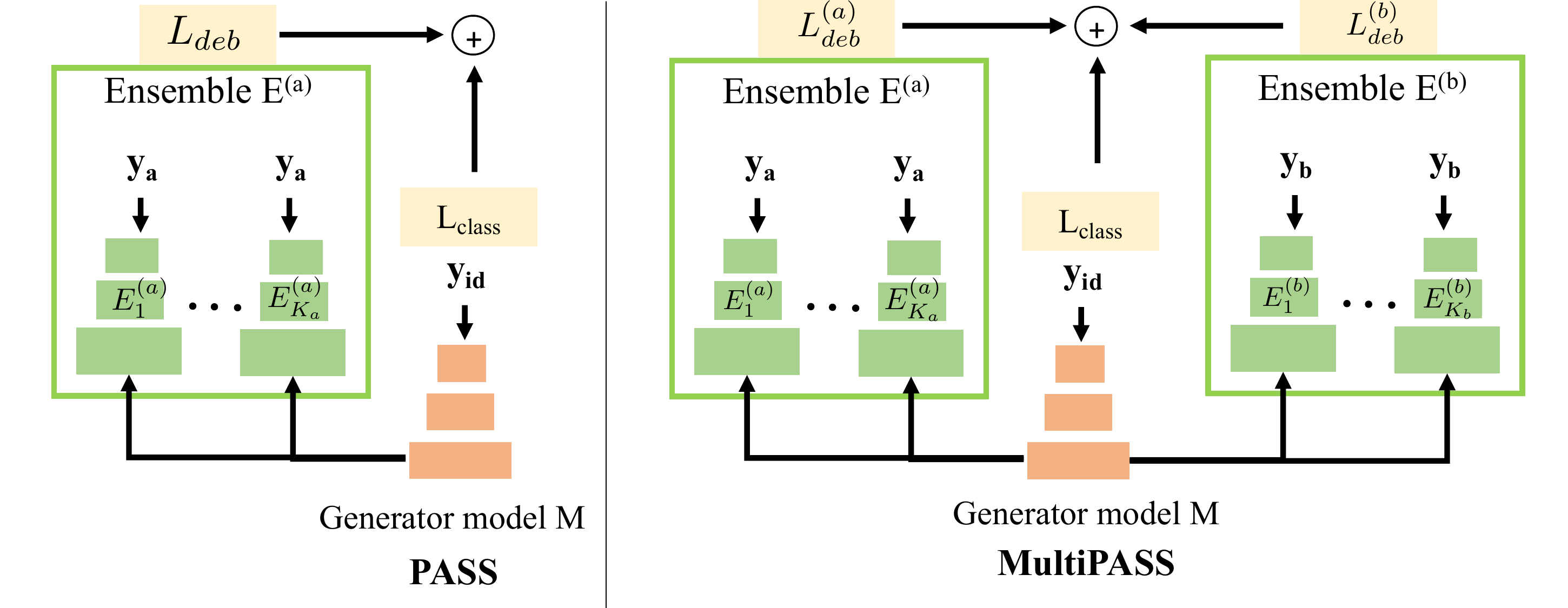}}
\caption{\small We build MultiPASS by extending PASS to tackle two attributes simultaneously.}
\label{fig:multipass}
\vspace{-18pt}
\end{figure}

\vspace{-0.2cm}
We also propose MultiPASS (Fig \ref{fig:multipass}), by extending PASS to reduce the information of several sensitive attributes simultaneously. Here, we describe how to extend PASS to tackle two attributes. 

We consider two attributes : Attribute $a$, with $N^{(a)}_{att}$ categories and attribute $b$, with $N^{(b)}_{att}$ categories. In contrast to PASS, we include two ensembles of discriminators in MultiPASS: one for attribute $a$, denoted as $E^{(a)}$ and one for attribute $b$, denoted as $E^{(b)}$. Let $E^{(a)}$ and $E^{(b)}$ consist of $K_a$ and $K_b$ adversary classifiers respectively. The stage 1 training for model $M$ in MultiPASS is same as that in PASS. In stage 2, we train both $E^{(a)}$ and $E^{(b)}$. 
In stage 3, we compute the outputs $o^{(a)}_{k}$ from all the classifiers in $E^{(a)}$ by extending Eq \ref{eq:ok}.
\begin{equation}
    o^{(a)}_{k} = E_{k}(f_{out}, \phi_{E^{(a)}_{k}})~~\text{for}~ k=1\ldots K_a
\end{equation}
\vspace{-0.15cm}
Using $o^{(a)}_{k}$ and extending Eq \ref{eq:adv} and \ref{eq:ldeb}, we compute the adversarial loss $L^{(E^{(a)}_{k})}_{adv}$ and debiasing loss $ L^{(a)}_{deb}$  with respect to  $E^{(a)}$ as follows:
\vspace{-0.1cm}
\begin{equation}
    L^{(E^{(a)}_{k})}_{adv} = -\sum_{i=1}^{N^{(a)}_{att}}\frac{1}{N^{(a)}_{att}}\text{log}(o^{(a)}_{k,i})
\end{equation}
\begin{equation}
\label{eq:ldeba}
    L^{(a)}_{deb} = \text{max}\{L^{(E^
    {(a)}_{k})}_{adv}|^{K_a}_{k=1} \}
\end{equation}
\vspace{-0.08cm}
\hspace{-5pt} We compute the adversarial loss $L^{(b)}_{deb}$ with respect to $E^{(b)}$ in a similar way. Using weights $\lambda_a$ for $L^{(a)}_{deb}$ and $\lambda_b$ for $L^{(b)}_{deb}$, we compute the bias reducing classification as follows:
\begin{equation}
\label{eq:lbrmulti}
    L_{br} = L_{class}+ \lambda_a L^{(a)}_{deb} + \lambda_b L^{(b)}_{deb}
\end{equation}
We provide the detailed MultiPASS algorithm in the supplementary material.
\section{Experiments}
\vspace{-4pt}
\subsection{Pre-trained networks and evaluation dataset}
\vspace{-4pt}
\label{subsec:netnds}
We evaluate the face descriptors obtained from the penultimate layer of following two pre-trained networks:\\
\textbf{Arcface\cite{afmodel} }: Resnet-101 trained on MS1MV2\cite{ms1mv2} with Additive Angular margin (Arcface) loss \cite{deng2018arcface}. \\
\textbf{Crystalface} : Resnet-101 trained on a mixture of UMDFaces\cite{bansal2017umdfaces}, UMDFaces-Videos\cite{bansal2017s} and MS1M \cite{guo2016ms}, with crystal loss \cite{ranjan2019fast}. \\
The aforementioned Arcface \cite{deng2018arcface} network achieves state-of-the-art performance in face verification and identification. Hence, we construct the baselines and our PASS framework on top of the Arcface descriptors, and provide detailed analysis for the same (in Sec. \ref{sec:results}). To evaluate the generalizability of PASS and baselines, we also perform similar experiments with Crystalface \cite{ranjan2019fast}  descriptors (in Sec. \ref{subsec:cryres}).

For evaluation, we use aligned faces from IJB-C, and follow the 1:1 face verification protocol defined in \cite{maze2018iarpa}. The alignment is done using \cite{ranjan2017all}. This dataset provides gender (male/female) and skintone labels. There are six classes for the skintone attribute which we reorganize into three groups, (i) \textit{Light} (`light pink' $\cup$ `light yellow'), (ii) \textit{Medium} (`medium pink' $\cup$ `medium yellow'), (iii) \textit{Dark} (`medium dark' $\cup$ `dark brown'). For evaluating gender bias, we compute the verification performance of face descriptors for male-male and female-female pairs separately (out of all the pairs defined in the IJB-C protocol \cite{maze2018iarpa}). To compute skintone bias, we compute the verification performance of face descriptors for dark-dark and light-light pairs.


Using Arcface and Crystalface, we extract 512 dimensional descriptors for the aligned faces in the IJB-C dataset which are then used for gender-wise and skintone-wise verification, the plots for which are provided in Fig. \ref{fig:cfafroc}.
\vspace{-0.11cm}
\subsection{PASS for gender and skintone}
\vspace{-0.2cm}
In Section \ref{sec:approach}, we present PASS as a general approach to de-bias face descriptors with respect to any attribute. Here, we show the effectiveness of PASS by using it to reduce information about gender and skintone (separately). We term the PASS framework trained to reduce gender information from descriptors as PASS-g, and its skintone counterpart as PASS-s. Additionally, we build another variant of PASS (called `MultiPASS') to reduce the predictability of gender and skintone simultaneously. To train PASS-g, PASS-s and MultiPASS, we first need to extract $f_{in}$ from a pre-trained face recognition network on a training dataset that consists of appropriate labels. $f_{in}$ is extracted using the Arface network, described in Section \ref{subsec:netnds}. \\
\textbf{PASS-g }: For training PASS-g, we extract $f_{in}$ for a combination of UMDFaces\cite{bansal2017umdfaces}, UMDFaces-Videos\cite{bansal2017s} and MS1M\cite{guo2016ms}.  There are 39,712 male and 18,308 female identities in the dataset. Face alignment and gender labels are obtained using \cite{ranjan2017all}. For PASS-g, $N_{att}=2$ (male/female). \\
\textbf{PASS-s} : To the best of our knowledge, we currently do not have a large dataset with skintone labels. So, we train PASS-s using $f_{in}$ extracted for a dataset with race labels instead, since there is some correlation between race and skintone \cite{norwood2014color}. We use the BUPT-BalancedFace \cite{wang2020mitigating} for training PASS-s (aligned using \cite{ranjan2017all}). The dataset consists of 1.3 million images for 28k identities. Each identity is associated with one of the four races : African, Asian, Indian and Caucasian. So, for PASS-s, $N_{att}=4$.\\
\textbf{MultiPASS}: We design MultiPASS by combining the adversarial ensembles in PASS-s and PASS-g. MultiPASS is trained using the descriptors for BUPT-BalancedFace dataset, which consists of race labels. The gender labels for this dataset are predicted using \cite{ranjan2017all}.

After training PASS/MultiPASS, we feed the 512-dimensional descriptor $f_{in}$ for test (IJB-C) images to the trained model $M$ which generates 256-dimensional $f_{out}$. $f_{out}$ is then used for face verification. Additional information on the hyperparameters required for training PASS is provided in the supplementary material, where we also analyze the effect of important hyperparameters on bias mitigation and verification performance. The code for implementing PASS will be made publicly available upon publication.
\vspace{-6pt}
\subsection{Baseline methods}
\vspace{-6pt}
\subsubsection{Incremental Variable Elimination (IVE)}
\vspace{-4pt}
IVE \cite{terhorst2019suppressing} is an attribute suppression algorithm that excludes variables in the face representation that affect attribute classification. We build a two variants of IVE: IVE(g) and IVE(s).
IVE(g) is trained to reduce gender information using Arcface descriptors descriptors from MS1M and gender labels predicted using \cite{ranjan2017all}.
Similarly, IVE(s) is trained to reduce skintone information using Arcface descriptors and labels from BUPT-BalancedFace \cite{wang2020mitigating}.
Additional training details are provided in the supplementary material.
\vspace{-14pt}
\subsubsection{Obscuring hair - similar to \cite{albiero2020face}}
\vspace{-6pt}
It is shown in \cite{albiero2020face} that obscuring hair in facial images during evaluation helps to reduce gender bias by improving the similarity scores of genuine female-female pairs. We construct a similar pipeline for gender-bias mitigation. We compute the face border keypoints using \cite{ranjan2017all} for the images in the evaluation dataset (IJB-C) and obscure all hair regions using these keypoints. Finally, we extract Arcface descriptors for these hair-obscured images. More details for \cite{albiero2020face} are provided in the supplementary material. 
\vspace{-0.1cm}
\subsection{Results}
\label{sec:results}
\vspace{-4pt}
\subsubsection{Evaluating leakage of gender and skintone}
\label{sec:leakage}
\vspace{-6pt}


To evaluate gender-leakage, we train an MLP classifier on Arcface descriptors and its de-biased counterparts (PASS variants/IVE). These descriptors are extracted for a training set with 60k images (30k males and females), sampled from IJB-C. The MLP classifier is a two hidden layer MLP with 128 and 64 hidden units respectively with SELU activations, followed by a sigmoid activated output layer. Subsequently, we test the MLP on descriptors extracted for 20k non-training images (10k males and females) in IJB-C. Finally, we compute the gender classification accuracy of the MLP. Using the same experimental setup with respect to skintone, we also train an MLP (with the same architecture) to predict skintone (dark/medium/light). In Tables \ref{tab:arcg} and \ref{tab:arcst}, we find that for both gender and skintone, \textit{the classification accuracy is lowest when the face descriptors are produced using MultiPASS}. We also find that classifiers trained on PASS-g and PASS-s descriptors obtain the second lowest classification accuracy.
This indicates that PASS variants are capable of reducing gender and skintone information in face descriptors.

\vspace{-8pt}
\subsubsection{Evaluating bias}
\vspace{-6pt}

\newcommand{\tabarcgender}{
\begin{tabular}{cc|ccccc|ccccc|ccccc}
\toprule
FPR & & \multicolumn{5}{c|}{$10^{-5}$} & \multicolumn{5}{c|}{$10^{-4}$} & \multicolumn{5}{c}{$10^{-3}$}\\
\midrule
Method& Acc-g $(\downarrow)$ & TPR\textsubscript{m}& TPR\textsubscript{f}& TPR&Bias$(\downarrow)$&BPC\textsubscript{g}$(\uparrow)$  & TPR\textsubscript{m} & TPR\textsubscript{f}& TPR&Bias$(\downarrow)$&BPC\textsubscript{g}$(\uparrow)$ & TPR\textsubscript{m} & TPR\textsubscript{f} & TPR&Bias$(\downarrow)$&BPC\textsubscript{g}$(\uparrow)$\\
\midrule
Arcface\cite{deng2018arcface}&82.06& 0.921&0.900&0.929 & 0.021&0.000 &0.962&0.947&0.953 &\textbf{0.015}& \textbf{0.000}&0.969&0.956&0.974 & 0.013 & 0.000\\
W/o hair\cite{albiero2020face}&80.77& 0.418&0.833&0.616 & 0.415&-19.099
&0.788&0.889&0.864&0.101&-5.827
&0.933&0.928&0.925 & \textbf{0.005} &\textbf{0.565}\\
IVE(g\cite{terhorst2019suppressing})&80.20& 0.922&0.881&0.925&0.041&-0.957
&0.962&0.947&0.950&\textbf{0.015}&\underline{-0.003}
&0.969&0.956&0.966&0.013&-0.008 \\
\midrule
PASS-g (ours)&\underline{73.65}& 0.900&0.881&0.919&\underline{0.019}&\underline{0.084}&0.948&0.925&0.946&0.023&-0.541&0.957&0.947&0.962&\underline{0.010}&\underline{0.218}\\
MultiPASS (ours) &\textbf{68.43}&0.871&0.874&0.881&\textbf{0.003}&\textbf{0.805}&0.934&0.919&0.934&\textbf{0.015}&-0.019&0.953&0.936&0.950&0.017&-0.332\\
\bottomrule
\end{tabular}}

\newcommand{\tabarcsgenderbpc}{
\begin{tabular}{ccc|cc|cc}
\toprule
FPR  & \multicolumn{2}{c|}{$10^{-5}$} & \multicolumn{2}{c|}{$10^{-4}$} & \multicolumn{2}{c}{$10^{-3}$} \\
\midrule
Method & TPR\textsubscript{*} & BPC\textsubscript{g} $(\uparrow)$ &  TPR\textsubscript{*}& BPC\textsubscript{g}$(\uparrow)$ &  TPR\textsubscript{*} & BPC\textsubscript{g}$(\uparrow)$\\
\midrule
Arcface\cite{deng2018arcface}& 0.929&0.000&0.953&0.000&0.974&0.000 \\
W/o hair\cite{albiero2020face} & 0.616&-19.099&0.864&-26.760&0.962&0.603 \\
IVE(g) \cite{terhorst2019suppressing} & 0.925&-0.957&0.950&-1.736&0.966&-0.008\\
\midrule
PASS-g (ours)& 0.919&0.799&0.946&\textbf{0.726}&0.925&\textbf{0.180}  \\
MultiPASS (ours) & 0.881&\textbf{0.805}&0.934&-0.019&0.950&-0.332  \\
\bottomrule
\end{tabular}}

\newcommand{\tabarcskintone}{
\begin{tabular}{cc|ccccc|ccccc|ccccc}
\toprule
FPR & & \multicolumn{5}{c|}{$10^{-4}$} & \multicolumn{5}{c|}{$10^{-3}$} & \multicolumn{5}{c}{$10^{-2}$}\\
\midrule
Method&Acc-st $(\downarrow)$ & TPR\textsubscript{l}& TPR\textsubscript{d}&TPR& Bias$(\downarrow)$&BPC\textsubscript{st}$(\uparrow)$  & TPR\textsubscript{l} & TPR\textsubscript{d} &TPR& Bias$(\downarrow)$ &BPC\textsubscript{st}$(\uparrow)$ & TPR\textsubscript{l} & TPR\textsubscript{d} & TPR&  Bias$(\downarrow)$&BPC\textsubscript{st}$(\uparrow)$\\
\midrule
Arcface \cite{deng2018arcface}&87.15& 0.951&0.938&0.953&0.013&0.000&0.974&0.968&0.974&0.006&0.000&0.976&0.974&0.976&0.002&0.000\\
IVE(s)\cite{terhorst2019suppressing}&88.23& 0.951&0.938&0.953&0.013&0.000&0.973&0.967&0.974&0.006&0.000&0.976&0.974&0.976&0.002&0.000 \\
\midrule
PASS-s (ours)&\underline{83.86}& 0.925&0.919&0.934&\textbf{0.006}&\textbf{0.519}&0.949&0.949&0.950&\textbf{0.000}&\textbf{0.975}&0.974&0.974&0.973&\textbf{0.000}&\textbf{0.997}\\
MultiPASS (ours)&\textbf{79.22}& 0.925&0.919&0.934&\textbf{0.006}&\textbf{0.519}&0.950&0.949&0.950&\underline{0.001}&\underline{0.809}&0.974&0.974&0.973&\textbf{0.000}&\textbf{0.997}\\
\bottomrule
\end{tabular}}

\newcommand{\tabarcskintonebpc}{
\begin{tabular}{ccc|cc|cc|cc}
\toprule
FPR  & \multicolumn{2}{c|}{$10^{-4}$} & \multicolumn{2}{c|}{$10^{-3}$} & \multicolumn{2}{c}{$10^{-2}$} \\
\midrule
Network & TPR\textsubscript{*} & BPC\textsubscript{g}$(\uparrow)$ &  TPR\textsubscript{*} & BPC\textsubscript{g}$(\uparrow)$ &  TPR\textsubscript{*} & BPC\textsubscript{*}$(\uparrow)$\\
\midrule
Arcface \cite{deng2018arcface}& 0.953&0.000&0.974&0.000&0.976&0.000 \\
IVE(s)\cite{terhorst2019suppressing} &0.953&0.000&0.974&0.000&0.976&0.000\\
PASS-s (ours) & 0.934&\textbf{0.519}&0.950&\textbf{0.975}&0.973&\textbf{0.997}\\
MultiPASS (ours)& 0.934&\textbf{0.519}&0.950&0.808&0.973&\textbf{0.997}\\
\bottomrule
\end{tabular}}

\begin{table*}%
  \scriptsize
  \centering
\tabarcgender
\caption{\small \textit{Gender} bias analysis and accuracy (`Acc-g') of gender classifier for \textit{Arcface} descriptors, and their transformed counterparts on IJB-C. TPR: overall True Positive rate, TPR\textsubscript{m}: male-male TPR, TPR\textsubscript{f}: female-female TPR. \textbf{Bold}=Best, \underline{Underlined}=Second best} \vspace{-0.2cm}
\label{tab:arcg}
\end{table*}

\begin{table*}%
  \scriptsize
  \centering
\tabarcskintone
\caption{\small \textit{Skintone} bias analysis and accuracy (`Acc-st') of skintone classifier for \textit{Arcface} descriptors, and their transformed counterparts on IJB-C. TPR: overall True Positive rate, TPR\textsubscript{l}: light-light TPR, TPR\textsubscript{d}: dark-dark TPR. \textbf{Bold}=Best, \underline{Underlined}=Second best}%
\label{tab:arcst}
\vspace{-0.4cm}
\end{table*}
\newcommand{\tabcrygenderbpc}{
\begin{tabular}{cc|cc|cc|cc}
\toprule
FPR & & \multicolumn{2}{c|}{$10^{-5}$} & \multicolumn{2}{c|}{$10^{-4}$} & \multicolumn{2}{c}{$10^{-3}$} \\
\midrule
Method & Acc-g$(\downarrow)$& TPR & BPC\textsubscript{g} $(\uparrow)$ &  TPR& BPC\textsubscript{g}$(\uparrow)$ &  TPR & BPC\textsubscript{g}$(\uparrow)$\\
\midrule
Crystalface\cite{ranjan2019fast}&86.73& 0.833&0.000&0.910&0.000&0.951&0.000 \\
W/o hair\cite{albiero2020face}&86.04 & 0.589&-8.926&0.780&0.823&0.899&0.731\\
IVE(g)\cite{terhorst2019suppressing} &86.10& 0.833&\underline{0.833}&0.910&0.391&0.951&0.071\\
\midrule
PASS-g &\underline{80.54}& 0.761&\textbf{0.847}&0.839&\textbf{0.857}&0.910&\textbf{0.956} \\
MultiPASS &\textbf{76.31}& 0.708&0.383&0.809&\underline{0.823}&0.881&\underline{0.784}\\
\bottomrule
\end{tabular}}

\begin{table}%
  \scriptsize
  \centering
\tabcrygenderbpc
\caption{\small \textit{Gender} bias analysis and accuracy (`Acc-g') of gender classifier of \textit{Crystalface} descriptors, and their transformed counterparts on IJB-C. \textbf{Bold}=Best, \underline{Underlined}=Second best}%
\label{tab:cryg}
\vspace{-0.2cm}
\end{table}
\begin{figure}
{\centering
\vspace{-0.3cm}
\subfloat[]{\includegraphics[width = 0.5\linewidth]{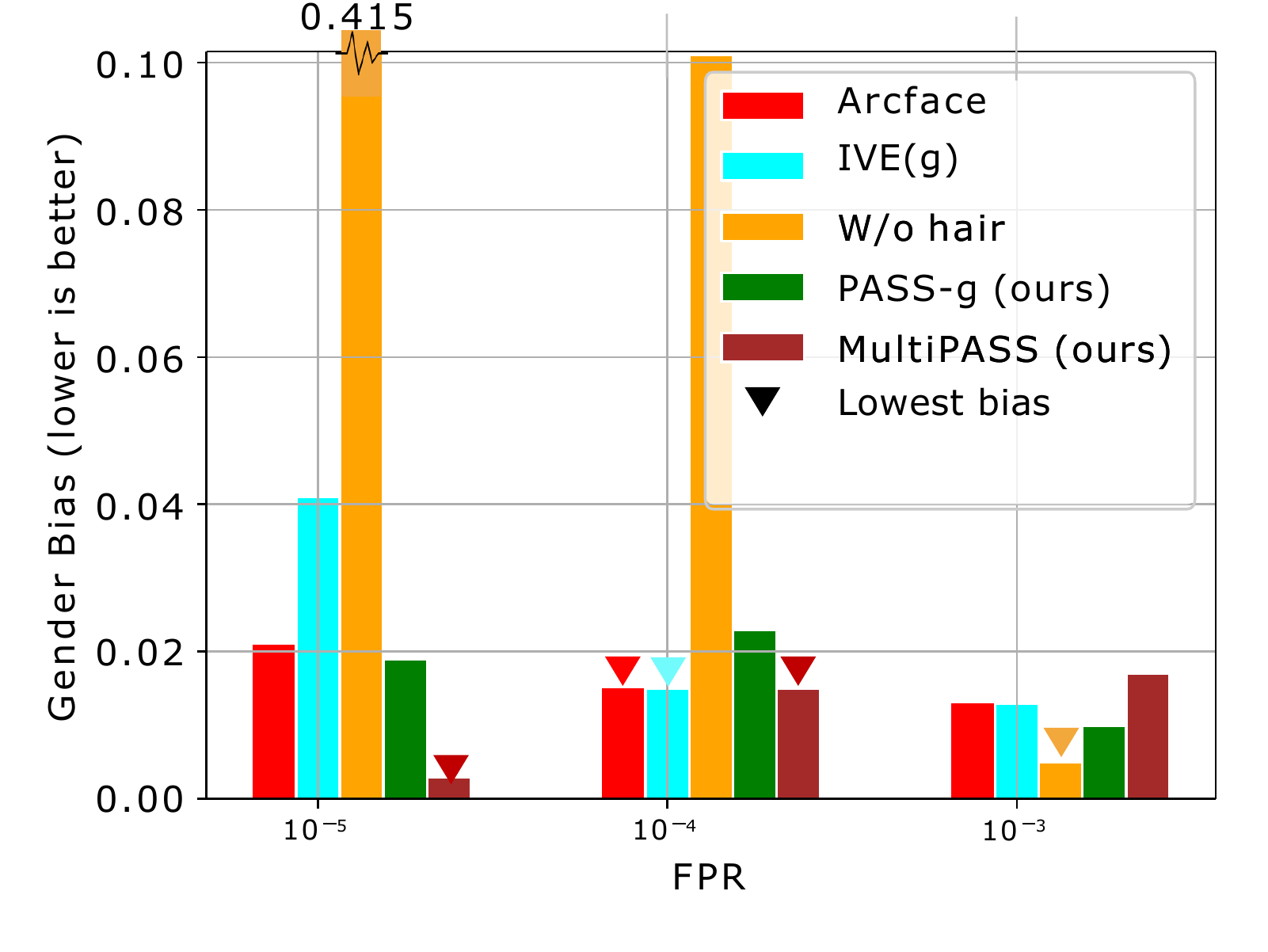}}
\subfloat[]{\includegraphics[width = 0.5\linewidth]{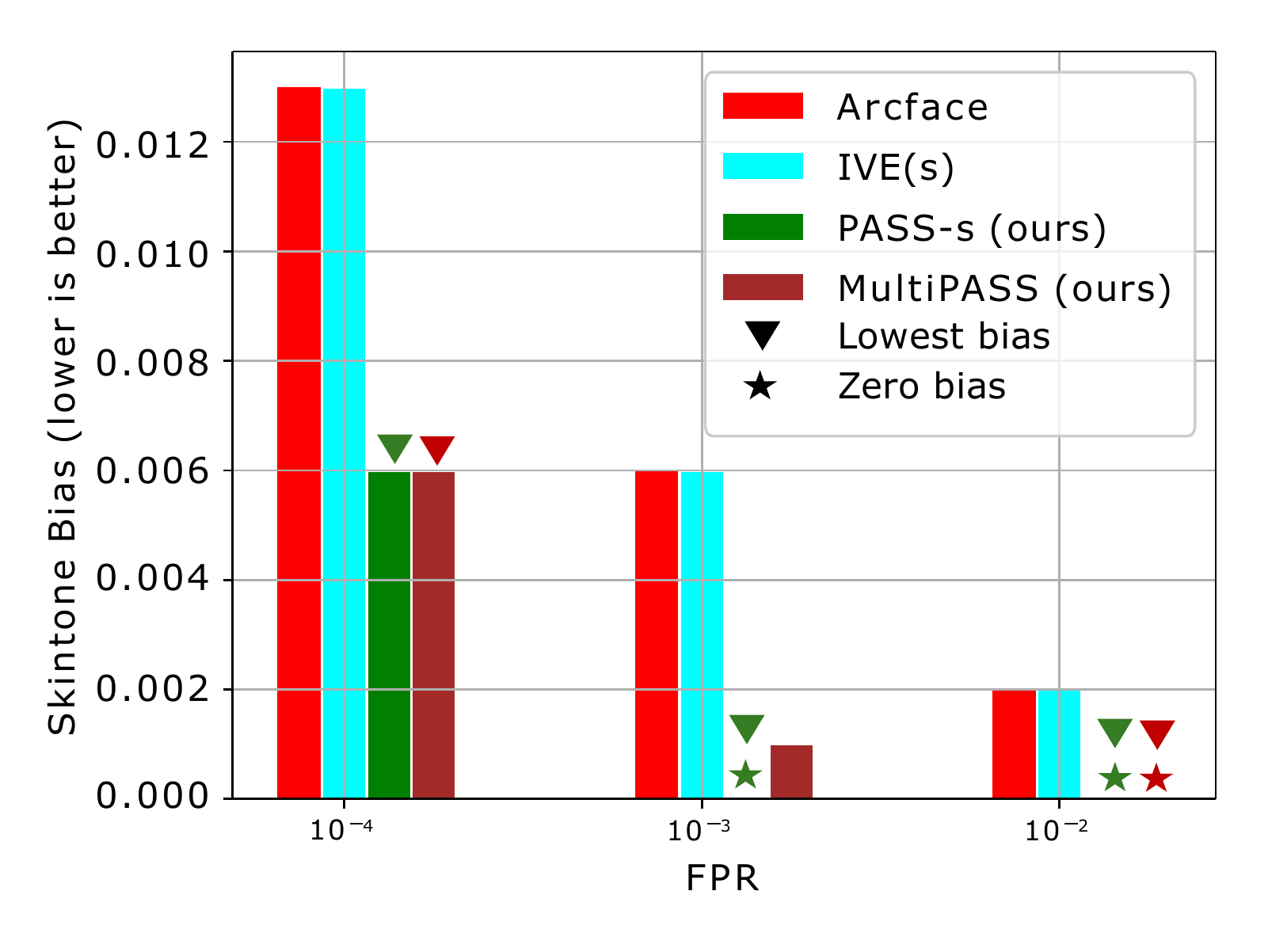}}

\caption{\small (a) Gender and (b) skintone bias in Arcface descriptors and their de-biased counterparts on IJB-C.}
\vspace{-0.35cm}
\label{fig:biasarc}
}
\end{figure}
We provide the gender-wise and skintone-wise verification TPRs and the corresponding bias on IJB-C for all the methods in Tables \ref{tab:arcg} and \ref{tab:arcst} respectively. From Fig \ref{fig:biasarc}, we infer that Arcface descriptors transformed using PASS/MultiPASS obtain lowest gender/skintone bias at most FPRs. Moreover, from Tables \ref{tab:arcg} and \ref{tab:arcst} , we also infer that \textit{PASS/MultiPASS-based frameworks obtain higher BPCs} (Eq \ref{eq:bpc}) \textit{than the baselines at most FPRs}. This shows that PASS variants are effective in reducing bias while maintaining high verification performance. We provide the gender-wise,  skintone-wise ROC plots (similar to the ROC curves in Fig \ref{fig:cfafroc}), along with overall verification plots in the supplementary material.

\vspace{-0.4cm} 
\subsubsection{End-to-end vs PASS}
\vspace{-0.3cm}
\newcommand{\tabcryskintonebpc}{
\begin{tabular}{cc|cc|cc|cc}
\toprule
FPR  && \multicolumn{2}{c|}{$10^{-4}$} & \multicolumn{2}{c|}{$10^{-3}$} & \multicolumn{2}{c}{$10^{-2}$} \\
\midrule
Method &Acc-st $(\downarrow)$& TPR & BPC\textsubscript{st}$(\uparrow)$ &  TPR & BPC\textsubscript{st}$(\uparrow)$ &  TPR & BPC\textsubscript{st}$(\uparrow)$\\
\midrule
Crystalface\cite{ranjan2019fast}& 89.30 & 0.910 & 0.000 & 0.951 & 0.000 & 0.974 & 0.000\\
IVE(s)\cite{terhorst2019suppressing} & 88.26 & 0.910 & -0.041 & 0.951 & -0.407 & 0.974 & -1.000\\
\midrule
PASS-s & \underline{83.84} & 0.844 & \underline{0.261} & 0.914 & \underline{0.702} & 0.919 & \underline{0.125} \\
MultiPASS & \textbf{79.44} & 0.809 & \textbf{0.639} & 0.881 & \textbf{0.927} & 0.968 & \textbf{0.994}\\
\bottomrule
\end{tabular}}
\begin{table}%
  \scriptsize
  \centering
\tabcryskintonebpc
\caption{\small \textit{Skintone} bias analysis and accuracy (`Acc-st') of skintone classifier for \textit{Crystalface} descriptors, and their transformed counterparts in IJB-C. \textbf{Bold}=Best, \underline{Underlined}=Second best}  
\label{tab:cryst}
\vspace{-0.6cm}
\end{table}

One subtlety when operating in an end-to-end fashion is that, in order to establish a baseline, one is generally required to retrain an entire face recognition system from scratch. Training such systems to achieve SOTA performance is technically challenging. Other works often report results using a weaker baseline system. For example, GAC~\cite{gac} uses a ResNet50 version of Arcface that achieves lower overall performance in IJB-C, than the original ArcFace, as shown in Table~\ref{tab:compsota}. Alternatively, PASS operates on pre-trained models, allowing us to start with an existing SOTA model, and maintaining nearly SOTA performance.
\vspace{-0.5cm}
\subsubsection{PASS with Crystalface}
\vspace{-0.3cm}
\begin{figure}
{\centering
\vspace{-0.3cm}
\subfloat[]{\includegraphics[width = 0.5\linewidth]{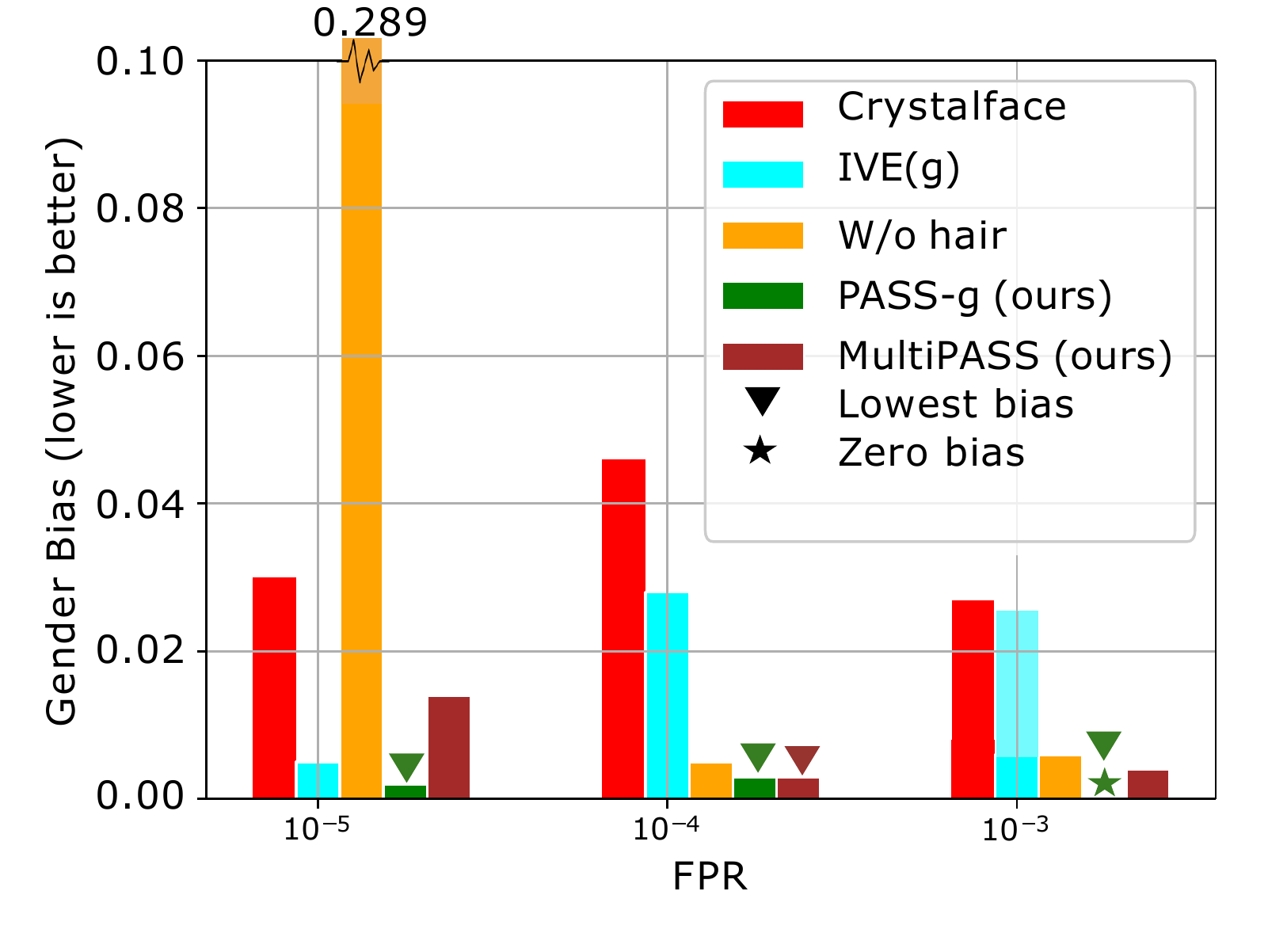}}
\subfloat[]{\includegraphics[width = 0.5\linewidth]{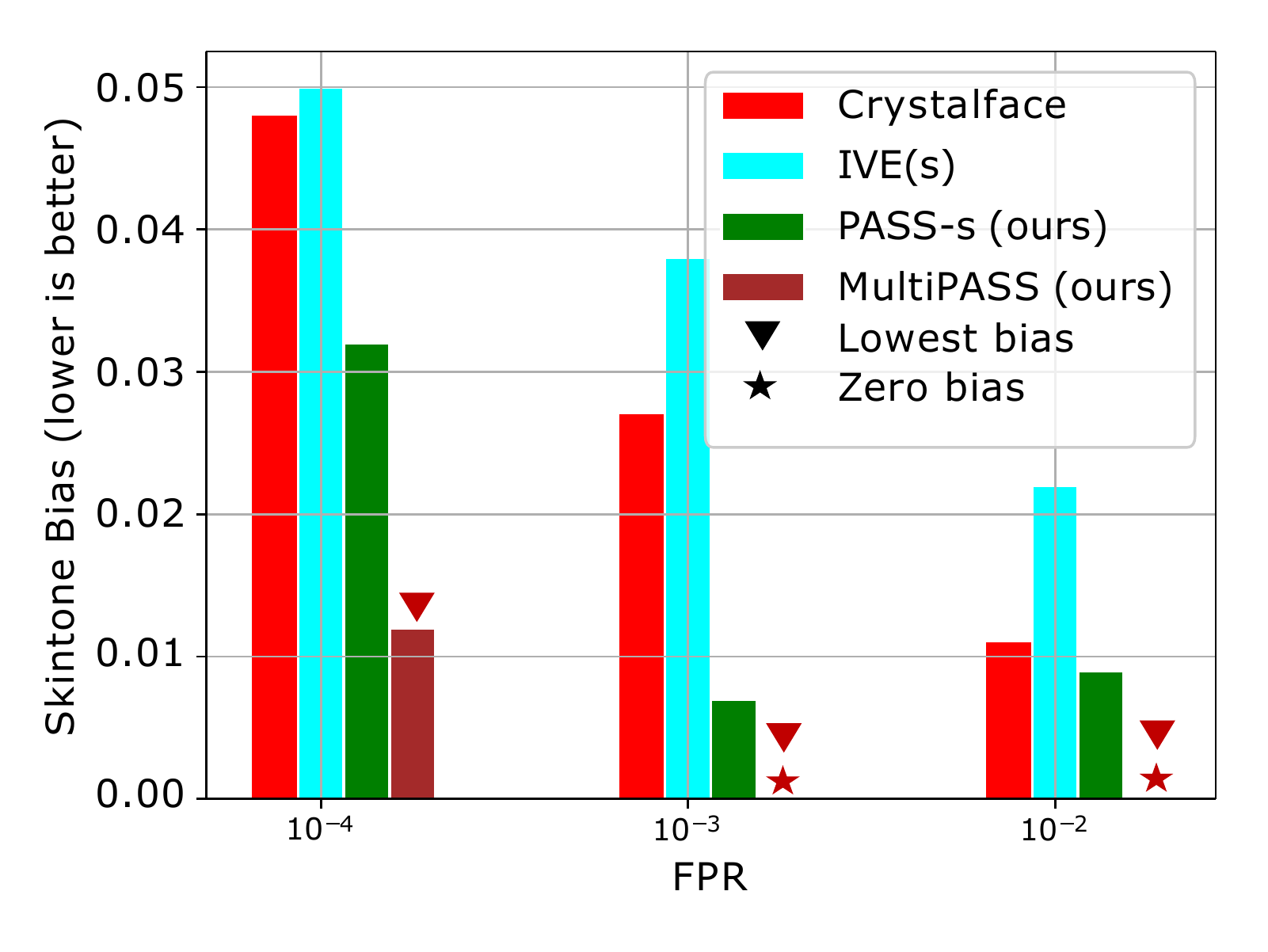}}

\caption{\small (a) Gender and (b) skintone bias in Crystalface descriptors and their de-biased counterparts on IJB-C.} \vspace{-0.4cm}
\label{fig:biascry}
}
\end{figure}
\begin{table}[]
    \centering
    \scriptsize
\vspace{0.4cm}
\begin{tabular}[h]{cccccc}
\hline
 Method/FPR & $10^{-5}$ & $10^{-4}$ & $10^{-3}$& Training method & Training attributes \\ 
\hline
Arcface \cite{deng2018arcface}(SOTA) & 92.9& 95.3& 97.4& -&-\\
\hline
Demo-ID\textsuperscript{+} \cite{gong2020jointly}&83.2& 89.4 &92.9&End-to-End&Age\\
Debface-ID\textsuperscript{+} \cite{gong2020jointly} & 82.0& 88.1 &89.5&End-to-End&Age,gender,race\\
GAC\textsuperscript{+} \cite{gac} &83.5&89.2&93.7&End-to-End&Race\\
\hline
PASS-s w/ AF & 88.1 & 93.4 & 95.0& Descriptor-based&Race\\
PASS-g w/ AF & 91.9 &94.6& 96.2&Descriptor-based&Gender\\
MultiPASS w/ AF &88.1 &93.4 &95.0 &Descriptor-based&Race, gender\\
\end{tabular}
\caption{IJB-C verification performance (TPR\% @ given FPR). AF refers to \textit{Arcface}.\textsuperscript{+} = Numbers copied from original paper. } \vspace{-0.5cm}
\label{tab:compsota}
\end{table}
\label{subsec:cryres}
To evaluate the generalizability of PASS and other baselines, we perform all of the aforementioned experiments on the Crystalface descriptors (mentioned in Sec. \ref{subsec:netnds}). We present the corresponding results of gender/skintone leakage in IJB-C in Tables \ref{tab:cryg} and \ref{tab:cryst}. We find that \textit{PASS and MultiPASS-transformed descriptors have the least gender/skintone predictability}. Similarly, \textit{Crystalface descriptors transformed with PASS/MultiPASS obtain the lowest bias} (Fig. \ref{fig:biascry}) and \textit{highest BPC values on IJB-C} (as shown in Tables \ref{tab:cryg} and \ref{tab:cryst}) \textit{at \textbf{all} FPRs, for both gender and skintone}. The hyperparameter information and detailed results for all the methods are provided in the supplementary material.

\newcommand{\tabcrygender}{
\begin{tabular}{cccc|ccc|ccc}
\toprule
FPR  & \multicolumn{3}{c|}{$10^{-5}$} & \multicolumn{3}{c|}{$10^{-4}$} & \multicolumn{3}{c}{$10^{-3}$}\\
\midrule
Network & TPR\textsubscript{m}& TPR\textsubscript{f}& Bias$(\downarrow)$  & TPR\textsubscript{m} & TPR\textsubscript{f} & Bias$(\downarrow)$ & TPR\textsubscript{m} & TPR\textsubscript{f} & Bias$(\downarrow)$\\
\midrule
Crystalface& 0.836&0.806&0.030&0.913&0.867&0.076&0.952&0.944&0.008\\
W/o hair& 0.424&0.713&0.289&0.779&0.774&0.005&0.881&0.875&0.006\\
IVE(g)& 0.818&0.813&0.005&0.912&0.884&0.028&0.952&0.946&0.006\\
\midrule
PASS-g& 0.751&0.749&0.002&0.831&0.828&0.003&0.922&0.922&0.00\\
MultiPASS&0.699&0.713&0.014&0.811&0.808&0.003&0.879&0.883&0.004\\
\bottomrule
\end{tabular}}

\newcommand{\tabcryskintone}{
\begin{tabular}{cccc|ccc|ccc}
\toprule
FPR  & \multicolumn{3}{c|}{$10^{-4}$} & \multicolumn{3}{c|}{$10^{-3}$} & \multicolumn{3}{c}{$10^{-2}$}\\
\midrule
Network & TPR\textsubscript{l}& TPR\textsubscript{d}& Bias$(\downarrow)$  & TPR\textsubscript{l} & TPR\textsubscript{d} & Bias$(\downarrow)$ & TPR\textsubscript{l} & TPR\textsubscript{d} & Bias$(\downarrow)$\\
\midrule
Crystalface & 0.912 & 0.864 & 0.048 & 0.948 & 0.921 & 0.027 & 0.974 & 0.963 & 0.011\\
IVE(s)& 0.912&0.862&0.05&0.949&0.911&0.038&0.975&0.953&0.022\\
\midrule
PASS-s& 0.850&0.818&0.032&0.913&0.906&0.007&0.962&0.953&0.009\\
MultiPASS& 0.826&0.838&\textbf{0.012}&0.907&0.907&\textbf{0.000}&0.953&0.953&\textbf{0.000}\\
\bottomrule
\end{tabular}}
\vspace{-0.25cm}
\subsection{OAT vs AET results}
\vspace{-0.2cm}
\label{sec:novcomp}
We train PASS-g systems with OAT and AET strategy on top of Arcface and Crystalface descriptors. We ensure that both OAT and AET approaches have the same number of classifiers ($K=3$ for Arcface, and $K=4$ for Crystalface) in ensemble $E$. We conduct the same gender-leakage experiment as done in Sec \ref{sec:leakage}, and report the gender classification accuracy of the trained MLP in Table \ref{tab:novcomp}. For both Arcface and Crystalface, \textit{MLP classifiers trained on descriptors from `PASS-g (OAT)' obtain lower accuracy than their AET counterparts}. Moreover, in Table \ref{tab:novcomp}, we find that the \textit{gender bias demonstrated by `PASS-g (OAT)' is lower than that of PASS-g (AET) at most FPRs}. In fact, from Table \ref{tab:novcomp}, it is clear that AET frameworks hardly reduce gender bias. Therefore, we conclude that our novel discriminator training strategy - OAT is an important component of PASS, and effectively removes sensitive attributes in descriptors.

\begin{table}
    \scriptsize
    \centering
    \begin{tabular}{cc|ccc|ccc|ccc}
        \toprule
        FPR \hspace{-6pt} & &\multicolumn{3}{c|}{$10^{-5}$} & \multicolumn{3}{c|}{$10^{-4}$} & \multicolumn{3}{c}{$10^{-3}$} \\
        \midrule
        Method \hspace{-6pt} & Acc-g & TPR\textsubscript{m} & TPR\textsubscript{f}& Bias & TPR\textsubscript{m} & TPR\textsubscript{f}& Bias & TPR\textsubscript{m} & TPR\textsubscript{f}& Bias \\
        \midrule
        Arcface \hspace{-6pt} & 82.06 & 0.921 & 0.900 & 0.021 & 0.962 & 0.947&\textbf{0.015} & 0.969 & 0.956 & 0.013 \\
        AET \hspace{-6pt} & 81.84 & 0.922 & 0.900 & 0.022 & 0.962 & 0.947 & \textbf{0.015} & 0.969 & 0.956 & 0.013 \\
        OAT \hspace{-6pt} & \textbf{73.65} & 0.900 & 0.881 & \textbf{0.019} & 0.948 & 0.925 & 0.023 & 0.957 & 0.947 & \textbf{0.010} \\
        \midrule
        Crystlfce \hspace{-6pt} & 86.73 & 0.836 & 0.806 & 0.030 & 0.913 & 0.867 & 0.046&0.952 & 0.924 & 0.028\\
        AET \hspace{-6pt} & 86.42 & 0.834 & 0.806 & 0.028 & 0.912 & 0.867 & 0.045 & 0.952 & 0.924 & 0.028\\
        OAT \hspace{-6pt} & \textbf{80.54} & 0.751 & 0.749 & \textbf{0.002} & 0.831 & 0.828 & \textbf{0.003} & 0.909 & 0.909 & \textbf{0.000}\\
        \bottomrule
    \end{tabular}
    \caption{Comparison of AET vs OAT strategies for gender bias reduction on Arcface (top) and Crystalface (bottom). Acc-g refers to gender classification accuracy (lower is better).} \vspace{-0.6cm}
    \label{tab:novcomp}
\end{table}
\vspace{-0.3cm}
\section{Conclusion}
\vspace{-0.2cm}
 We present an adversarial approach called PASS that can reduce the information of any protected attribute in face descriptors, while making them proficient in identity classification. Our approach allows the user to re-use the pre-computed descriptors for de-biasing them, without the need for expensive end-to-end training. In PASS, we also propose a novel discriminator training strategy called OAT to enforce removal of sensitive attributes and show that OAT is an important component of PASS. PASS can also be extended (as MultiPASS) to reduce the information of multiple attributes simultaneously. 
 
 
\vspace{-0.3cm}
\section*{Acknowledgement}
The authors would like to thank Dr. P. Jonathon Phillips (NIST) and Dr. Rajeev Ranjan (Amazon) for their helpful suggestions. This research is based upon work supported by a MURI from the Army Research Office under the Grant No. W911NF-17-1-0304. This is part of the collaboration between US DOD, UK MOD and UK Engineering and Physical Research Council (EPSRC) under the Multidisciplinary University Research Initiative. Carlos D. Castillo was supported by funding provided by National Eye Institute Grant R01EY029692-03.
{\small
\bibliographystyle{ieee_fullname}

}
\newpage
\appendix
\renewcommand{\thesection}{A\arabic{section}}  
\setcounter{table}{0}
\renewcommand{\thetable}{A\arabic{table}}  
\setcounter{figure}{0}
\renewcommand{\thefigure}{A\arabic{figure}}
\section*{Supplementary material}

In this supplementary material, we provide information about the following:
1. Relation between attribute predictability and bias (Sec. \ref{sec:prednbias}), 2. Detailed algorithm (pseudocode) for PASS and MultiPASS (Sec. \ref{sec:algs}), 3. Hyperparameters used for training PASS and MultiPASS systems (Sec. \ref{sec:hps}), 4. Hyperparameters for training IVE systems (Sec. \ref{sec:hpive}), 5. Our hair-obscuring pipeline (similar to \cite{albiero2020face}) (Sec. \ref{sec:bowyer}), 6. Detailed results (including verification plots) for de-biasing methods applied on Arcface/Crystalface descriptors (Sec. \ref{sec:res}), 7. Ablation study for PASS systems (Sec. \ref{sec:ablation}), 8. Effect of training a discriminative embedding (TPE\cite{Swami_2016_triplet}) on face descriptors and their PASS counterpart (Sec. \ref{sec:tpe}), 9. Advantages of deploying PASS system over end-to-end training (Sec. \ref{sec:advpass}), 10. Discussion about the trade-off between bias reduction and drop in verification performance (Sec. \ref{sec:tradeoff}).

\section{Relation between predictability and bias}
\label{sec:prednbias}
In the Section 3 of the main paper, we hypothesize that \textit{reducing the ability to predict protected attributes (gender and skintones) in face descriptors will reduce gender/skintone bias in face verification tasks}. This hypothesis is built on the results of \cite{gong2020jointly}, which shows that adversarially removing sensitive information from face representations reduces bias. In the context of gender/skintone bias, we conduct additional experiments to provide the reasoning for this hypothesis. We compare the gender and skintone predictability (i.e. ability to classify an gender/skintone) of face descriptors extracted from Arcface and Crystalface networks and analyze the corresponding bias demonstrated by these networks.\\ 
\textbf{Evaluating gender bias and predictability:} Using the IJB-C dataset, we first build a training set with 60k images (30k males and females). Similarly, we construct a test set of 20k images (10k males and females). The images for training and testing are selected randomly, and the face descriptors are extracted using the pre-trained networks (Arcface or Crystalface). There is no overlap between the identities in training and testing set. Subsequently, we train an MLP classifier on face descriptors of the training set to classify gender and evaluate it on the test descriptors. This is done for both Arcface and Crystalface descriptors. The MLP classifier is a two hidden layer MLP with 128 and 64 hidden units respectively with SELU activations, followed by a sigmoid activated output layer. The gender classification accuracy is reported in Table \ref{tab:hypgen}. Using the gender-wise verification results in Figure 2(a) in the main paper, we also compute the gender bias at every FPR and present it in Table \ref{tab:hypgen}. \\
\newcommand{\tabhypgender}{
\begin{tabular}{ccccccccccc}
\toprule
FPR  & &\multicolumn{3}{c|}{$10^{-5}$} & \multicolumn{3}{c|}{$10^{-4}$} & \multicolumn{3}{c}{$10^{-3}$} \\
\midrule
Network & Acc-g & TPR\textsubscript{m} & TPR\textsubscript{f}& Bias & TPR\textsubscript{m} & TPR\textsubscript{f}& Bias &  TPR\textsubscript{m} & TPR\textsubscript{f}& Bias \\
\midrule
Arcface &\red{82.06}& 0.921&0.900 & \textbf{0.021} &0.962&0.947 &\textbf{0.015}&0.969&0.956 & 0.013 \\
Crystalface & 86.73& 0.836 &0.806 & 0.030&0.913&0.867&0.046&0.952&0.944&0.008 \\
\bottomrule
\vspace{-6pt}
\end{tabular}}
\newcommand{\tabhypskintone}{
\begin{tabular}{ccccccccccc}
\toprule
FPR  & &\multicolumn{3}{c|}{$10^{-4}$} & \multicolumn{3}{c|}{$10^{-3}$} & \multicolumn{3}{c}{$10^{-2}$} \\
\midrule
Network & Acc-s & TPR\textsubscript{l} & TPR\textsubscript{d}& Bias & TPR\textsubscript{l} & TPR\textsubscript{d}& Bias &  TPR\textsubscript{l} & TPR\textsubscript{d}& Bias \\
\midrule
Arcface &\red{87.15}&0.951&0.938&\textbf{0.013}&0.974&0.968&\textbf{0.006}&0.976&0.974&\textbf{0.002} \\
Crystalface &89.30&0.912&0.864 &0.048 &0.948 &0.921 & 0.027 &0.974 &0.963&0.011\\
\bottomrule
\vspace{-6pt}
\end{tabular}}
\begin{table}%
  \scriptsize
  \centering
  \tabhypgender%
  \caption{\small Gender bias in IJB-C verification - Arcface vs Crystalface. Acc-g $=$ performance of MLP classifier in predicting Gender.}
  \label{tab:hypgen}
  \vspace{-11pt}
\end{table}
\begin{table}%
  \scriptsize
  \centering
  \tabhypskintone%
  \caption{\small Skintone bias in IJB-C verification - Arcface vs Crystalface. Acc-s $=$ performance of MLP classifier in predicting Skintone.}
  \label{tab:hypst}
  \vspace{-15pt}
\end{table}
\textbf{Evaluating skintone bias and predictability:} We follow the same experimental setup for skintone. The only difference is that the training and testing sets are balanced in terms of skintone (dark, medium and light) and the MLP has three output nodes corresponding to light, medium, and dark skintones. The skintone classification accuracy is reported in Table \ref{tab:hypst}. Using the skintone-wise verification results in Figure 2(b) in the main paper, we also compute the skintone bias at every FPR and present it in Table \ref{tab:hypst}.

From the results in Tables \ref{tab:hypgen} and \ref{tab:hypst}, we find that Arcface descriptors have lower gender/skintone predictability than Crystalface descriptors. Moreover, the Arcface descriptors also demonstrate lower gender/skintone bias than their Crystalface counterparts at most FPRs (Tables \ref{tab:hypgen} and \ref{tab:hypst}). From this, we infer that \textit{face descriptors with low gender/skintone predictability appear to demonstrate lower gender/skintone bias in face verification}, thus forming the basis of our initial hypothesis. Therefore, we propose techniques and construct baselines to reduce the predictability of gender and skintone in face descriptors while making them proficient in identity classification. \\
\textbf{Why reduce predictability of protected attributes?} Reducing predictability of a protected attribute from a face descriptor to zero implies that no information about that attribute is present in the descriptor. This also implies that no information about the attribute is used to represent identity. Thus, following from the data processing inequality~\cite{cover2012elements}, any prediction that is a function of the descriptor is independent of the protected attribute.

\section{PASS and MultiPASS algorithm}
\label{sec:algs}
In section 4.1.1 of the main paper, we explain the components of our proposed adversarial PASS system and discuss the stage-wise training procedure in section 4.1.2 (main paper). Here, we present the detailed algorithm for PASS in Algorithm \ref{alg:pass}. 
\begin{algorithm}[t]
\algsetup{linenosize=\small}
 \small
\floatname{algorithm}{Algorithm}
\caption{PASS}
\label{alg:pass}
\begin{algorithmic}[1]
\STATE \textbf{Required}: $N_{ep}$: Number of training episodes
\STATE \textbf{Required}: $\lambda, K, T_{fc}, A^*, T_{deb}, T_{atrain}, T_{plat}, T_{ep}$\\
\STATE \textbf{Required} Learning rates: $\alpha_1, \alpha_2, \alpha_3$
\FOR {$i$ in \textbf{range}($N_{ep}$)}
\STATE Begin \textbf{Stage 1} (initial training of $M$ and $C$)
\IF{$i$ == 0}
\STATE Initialize $\phi_M$ and $\phi_C$ with random weights
\FOR {$n$ in \textbf{range}($T_{fc}$)}
\STATE $\phi_M\longleftarrow \phi_M - \alpha_1\nabla_{\phi_M}L_{class}(\phi_M,\phi_C)$
\STATE $\phi_C\longleftarrow \phi_C - \alpha_1\nabla_{\phi_C}L_{class}(\phi_M,\phi_C)$
\ENDFOR
\ENDIF
\STATE Begin \textbf{Stage 2} (initial training of $E$)
\IF {$i$ mod $T_{ep}$ == 0}
\STATE Initialize $\phi_E$ with random weights
\FOR{$n$ in \textbf{range}($T_{atrain}$)}
\STATE $\phi_E \longleftarrow \phi_E - \alpha_2\nabla_{\phi_E}L_{att}(\phi_M, \phi_E)$
\ENDFOR
\ENDIF
\STATE Begin \textbf{Stage 3} (update $M$ and $C$)
\FOR{$n$ in \textbf{range}($T_{deb}$)}
\STATE $\phi_M \longleftarrow \phi_M - \alpha_3 \nabla_{\phi_M}L_{br}(\phi_C, \phi_M, \phi_E)$
\STATE $\phi_C \longleftarrow \phi_C - \alpha_3 \nabla_{\phi_C}L_{br}(\phi_C, \phi_M, \phi_E)$
\ENDFOR
\STATE Begin \textbf{Stage 4} (update $E_k$)
\STATE $k$ = $i$ mod $K$
\FOR{$n$ in \textbf{range}($T_{plat}$)}
\STATE Compute validation attribute prediction accuracy $A$ of $E_k$
\IF{$A>A^*$}
\STATE break
\ENDIF
\STATE $\phi_{E_{k}} \longleftarrow \phi_{E_{k}} - \alpha_2 \nabla_{\phi_{E_k}} L^{(E_k)}_{att}(\phi_M, \phi_{E_{k}})$
\ENDFOR
\ENDFOR
\end{algorithmic}
\end{algorithm}

\begin{algorithm}[t]
\algsetup{linenosize=\small}
 \small
\floatname{algorithm}{Algorithm}
\caption{MultiPASS}
\label{alg:multipass}
\begin{algorithmic}[1]
\STATE \textbf{Required}: $N_{ep}$: Number of training episodes
\STATE \textbf{Required}:$\lambda_a, \lambda_b, K_a, K_b, T_{fc}, A^{*}_1,A^{*}_2$
\STATE \textbf{Required}:$T_{deb}, T^{(a)}_{atrain}, T^{(b)}_{atrain}, T_{plat}, T_{ep}$
\STATE \textbf{Required} Learning rates: $\alpha_1, \alpha_2, \alpha_3$
\FOR {$i$ in \textbf{range}($N_{ep}$)}
\STATE Begin \textbf{Stage 1} (initial training of $M$ and $C$)
\IF{$i$ == 0}
\STATE Initialize $\phi_M$ and $\phi_C$ with random weights
\FOR {$n$ in \textbf{range}($T_{fc}$)}
\STATE $\phi_M\longleftarrow \phi_M - \alpha_1\nabla_{\phi_M}L_{class}(\phi_M,\phi_C)$
\STATE $\phi_C\longleftarrow \phi_C - \alpha_1\nabla_{\phi_C}L_{class}(\phi_M,\phi_C)$
\ENDFOR
\ENDIF
\STATE Begin \textbf{Stage 2} (initial training of $E^{(a)}, E^{(b)}$)
\IF {$i$ mod $T_{ep}$ == 0}
\STATE Initialize $\phi_{E^{(a)}},\phi_{E^{(b)}}$ with random weights
\FOR{$n$ in \textbf{range}($T^{(a)}_{atrain}$)}
\STATE $\phi_{E^{(a)}} \longleftarrow \phi_{E^{(a)}} - \alpha_2\nabla_{\phi_E}L^{(a)}_{att}(\phi_M, \phi_{E^{(a)}})$
\ENDFOR
\FOR{$n$ in \textbf{range}($T^{(b)}_{atrain}$)}
\STATE $\phi_{E^{(b)}} \longleftarrow \phi_{E^{(b)}} - \alpha_2\nabla_{\phi_E}L^{(b)}_{att}(\phi_M, \phi_{E^{(b)}})$
\ENDFOR
\ENDIF
\STATE Begin \textbf{Stage 3} (update $M$ and $C$)
\FOR{$n$ in \textbf{range}($T_{deb}$)}
\STATE $\phi_M \longleftarrow \phi_M - \alpha_3 \nabla_{\phi_M}L_{br}(\phi_C, \phi_M, \phi_{E^{(a)}},  \phi_{E^{(b)}})$
\STATE $\phi_C \longleftarrow \phi_C - \alpha_3 \nabla_{\phi_C}L_{br}(\phi_C, \phi_M, \phi_{E^{(a)}},  \phi_{E^{(b)}})$
\ENDFOR
\STATE Begin \textbf{Stage 4} (update $E^{(a)}_{k_a},E^{(b)}_{k_b}$)
\STATE $k_a$ = $i$ mod $K_a$
\STATE $k_b$ = $i$ mod $K_b$
\FOR{$n$ in \textbf{range}($T_{plat}$)}
\STATE Compute validation attribute prediction accuracy $A_1$ of $E^{(a)}_{k_a}$ and $A_2$ of $E^{(b)}_{k_b}$ 
\IF{$A_1>A^{*}_{1}$ and $A_2>A^{*}_{2}$}
\STATE break
\ENDIF
\STATE $\phi_{E^{(a)}_{k_a}} \longleftarrow \phi_{E^{(a)}_{k_a}} - \alpha_2 \nabla_{\phi_{E^{(a)}_{k_a}}} L^{(E^{(a)}_{k_a})}_{att}(\phi_M, \phi_{E^{(a)}_{k_a}})$
\STATE $\phi_{E^{(b)}_{k_b}} \longleftarrow \phi_{E^{(b)}_{k_b}} - \alpha_2 \nabla_{\phi_{E^{(b)}_{k_b}}} L^{(E^{(b)}_{k_b})}_{att}(\phi_M, \phi_{E^{(b)}_{k_b}})$
\ENDFOR
\ENDFOR
\end{algorithmic}
\end{algorithm}
Following this, we extend PASS to MultiPASS by reducing the information of two attributes simultaneously: Attribute $a$, with $N^{(a)}_{att}$ categories and attribute $b$, with $N^{(b)}_{att}$ categories. The detailed algorithm for training MultiPASS is provided in Algorithm \ref{alg:multipass}. We include two ensembles of discriminators in MultiPASS: one for attribute $a$, denoted as $E^{(a)}$ and one for attribute $b$, denoted as $E^{(b)}$. Let $E^{(a)}$ and $E^{(b)}$ consist of $K_a$ and $K_b$ adversary classifiers respectively. The weights for all the classifiers in $E^{(a)}$ are collectively denoted as $\phi_{E^{(a)}}$ and those for $E^{(b)}$ are denoted as $\phi_{E^{(b)}}$.  The stage 1 training for model $M$ in MultiPASS is same as that in PASS. \\
\textbf{Stage 2}: In stage 2, we train both $E^{(a)}$ (for $T^{(a)}_{atrain}$ iterations) and $E^{(b)}$ (for $T^{(b)}_{atrain}$ iterations). An adversarial classifier $E^{(a)}_k$  in $E^{(a)}$ is trained with a standard cross entropy classification loss $L^{E^{(a)}_k}_{att}$
\begin{equation}
    L^{E^{(a)}_k}_{att} = -\sum_{i=1}^{N^{(a)}_{att}} y_{a,i} \text{log } y_{a,i}^{(k)}.  
\end{equation}
Here $\mathbf{y_a}$ denotes the one hot label with respect to attribute $a$. $\mathbf{y^{(k)}_a}$ is the softmaxed output from the $k^{th}$ adversary classifier in ensemble $E^{(a)}$. The classification loss $L^{(a)}_{att}$ (in line 17 of Algorithm \ref{alg:multipass}) for the entire ensemble $E^{(a)}_k$ is computed by summing up $L^{E^{(a)}_k}_{att}$ as follows:
\begin{equation}
     L^{(a)}_{att} = \sum^{K_a}_{k=1}L^{E^{(a)}_k}_{att}
\end{equation}
We train the classifiers in ensemble $E^{(b)}$ in a similar way. \\
\textbf{Stage 3}: Subsequently, we train model $M$ for $T_{deb}$ iterations to generate $f_{out}$ to classify identities (similar to stage 3 in Algorithm \ref{alg:pass}), while reducing the information of attributes $a$ and $b$ simultaneously. $f_{out}$ from $M$ is provided to both $E^{(a)}$ and $E^{(b)}$ for computing debiasing losses $L^{(a)}_{deb}$ and $L^{(b)}_{deb}$ (See Eq. 14 in main paper). This is used to compute the bias reducing classification loss $L_{br}$ (Eq 15 in the main paper).\\
\textbf{Stage 4}: After stage 3, we update the adversary classifiers in $E^{(a)}$ and $E^{(b)}$. Using our proposed OAT strategy we choose one classifier $E^{(a)}_{k_a}$ in $E^{(a)}$ and $E^{(b)}_{k_b}$ in $E^{(b)}$ (Lines 29 and 30 in Algorithm \ref{alg:multipass}). We train them for $T_{plat}$ iterations or until $E^{(a)}_{k_a}$ reaches a threshold accuracy of $A^{*}_1$ and $E^{(b)}_{k_b}$ reaches a threshold accuracy of $A^{*}_2$  on the validation set. We run stages 3 and 4 alternatively, for $T_{ep}$ episodes, after which we re-initialize and re-train all the models in $E^{(a)}$ and $E^{(b)}$ (as done in stage 2). 
\section{Hyperparameters for PASS and MultiPASS}
\label{sec:hps}
We provide the hyperparameters used to train PASS-g and PASS-s systems on Arcface and Crystalface descriptors in Table \ref{tab:hppass}. 

In our MultiPASS framework, we use attribute $a$ as gender ($N^{(a)}_{att}=2$, male/female), and attribute $b$ as race ($N^{(b)}_{att}=4$, Caucasian/Indian/Asian/African). Thus $E^{(a)}$ is an ensemble of gender classifiers and $E^{(a)}$ is an ensemble of race classifiers. Note that, we train MultiPASS on BUPTBalancedFace which consists of race labels, since we currently do not have a large training dataset with skintone labels. The hyperparameters for MultiPASS systems are provided in Table \ref{tab:hpmultipass}. We use a batch size of 400 in all the experiments.
\begin{table}%
  \small
  \centering
\begin{tabular}{cccc|cc}
\toprule
Network  & &\multicolumn{2}{c|}{Arcface} & \multicolumn{2}{c}{Crystalface}\\
\midrule
Hyperparameter & Stage & PASS-g & PASS-s& PASS-g & PASS-s \\
\midrule
$\lambda$ & 3&10&10&1&10\\
$K$ & 2, 3, 4 &3&2&4&2\\
$T_{fc}$ & 1  &10000&10000&16000&16000\\
$T_{deb}$ & 3 &1200&1200&1200&1200\\
$T_{atrain}$ & 2 &30000&30000&30000&30000\\
$T_{plat}$ & 4  &2000&2000&2000&2000\\
$A^*$ &  4&0.95&0.95&0.90&0.95\\
$\alpha_1$ & 1 &$10^{-2}$&$10^{-2}$&$10^{-2}$&$10^{-2}$\\
$\alpha_2$ &  2,4&$10^{-3}$&$10^{-3}$&$10^{-3}$&$10^{-3}$\\
$\alpha_3$ &  3&$10^{-4}$&$10^{-4}$&$10^{-4}$&$10^{-4}$\\
$T_{ep}$ & 3,4 &40&40&40&40\\
\bottomrule
\vspace{-6pt}
\end{tabular}
\caption{\small  Hyperparameters for training PASS-g and PASS-s on Arcface and Crystalface descriptors}
  \label{tab:hppass}
  \vspace{-11pt}
\end{table}
\begin{table}%
  \small
  \centering
\begin{tabular}{cccc}
\toprule
Hyperparameter & Stage & Arcface &  Crystalface \\
\midrule
$\lambda_a$ & 3&10&1\\
$\lambda_b$ & 3&10&10\\
$K_a$ & 2, 3, 4 &3&4\\
$K_b$ & 2, 3, 4 &2&2\\
$T_{fc}$ & 1  &10000&16000\\
$T_{deb}$ & 3 &1200&1200\\
$T^{(a)}_{atrain}$ & 2 &30000&30000\\
$T^{(b)}_{atrain}$ & 2 &30000&30000\\
$T_{plat}$ & 4  &2000&2000\\
$A^{*}_1$ &  4&0.95&0.90\\
$A^{*}_2$ &  4&0.95&0.95\\
$\alpha_1$ & 1 &$10^{-2}$&$10^{-2}$\\
$\alpha_2$ &  2,4&$10^{-3}$&$10^{-3}$\\
$\alpha_3$ &  3&$10^{-4}$&$10^{-4}$\\
$T_{ep}$ & 3,4&40&40\\
\bottomrule
\vspace{-6pt}
\end{tabular}
\caption{\small Hyperparameters for training MultiPASS on Arcface and Crystalface descriptors}
  \label{tab:hpmultipass}
  \vspace{-11pt}
\end{table}

\section{Hyperparameters for IVE(g) and IVE(s)}
\label{sec:hpive}
IVE \cite{terhorst2019suppressing} is an attribute suppression algorithm that uses a decision tree ensemble to score each variable in face representations with respect to their importance for a specific recognition task. Variables affecting attribute classification in a significant way are then excluded from the representation. Each step of exclusion removes $n_e$ variables from the representation. The algorithm runs for $n_s$ steps, thus resulting in removal of $n_s \times n_e$ variables from the representation. We train IVE(g) by using face descriptors of MS1M dataset, extracted using a pre-trained netowrk (Arcface or Crystalface). The gender labels are obtained using \cite{ranjan2017all}. \\

We follow the same experimental setup for training IVE(s). The only difference is that the training dataset for training IVE(s) is BUPT-BalancedFace \cite{wang2020mitigating}. The official implementation for training IVE is publicly available \cite{ivecode}. In all of our IVE experiments, we use the parameters values mentioned in the code, i.e. $n_s=20$ and $n_e=5$, thus resulting in 100 eliminations. Since face descriptors from Arcface or Crystalface are 512-dimensional, the trained IVE(s/g) framework transforms the input descriptors for test images into $512-100=412$ dimensional descriptors. These descriptors are then used to perform face verification.
\newcommand{\tabcrygenderbpcsupp}{
\begin{tabular}{c|ccccc|ccccc|ccccc}
\toprule
FPR &  \multicolumn{5}{c|}{$10^{-5}$} & \multicolumn{5}{c|}{$10^{-4}$} & \multicolumn{5}{c}{$10^{-3}$} \\
\midrule
Network & TPR\textsubscript{m}&TPR\textsubscript{f}&TPR & Bias ($\downarrow$)& BPC\textsubscript{g} $(\uparrow)$& TPR\textsubscript{m}&TPR\textsubscript{f}&TPR & Bias ($\downarrow$)& BPC\textsubscript{g} $(\uparrow)$& TPR\textsubscript{m}&TPR\textsubscript{f}&TPR & Bias ($\downarrow$)& BPC\textsubscript{g} $(\uparrow)$ \\
\midrule
Crystalface\cite{ranjan2019fast}&0.836 &0.806&0.833&0.030&0.000&0.913&0.867&0.910&0.046&0.000&0.952&0.924&0.951&0.028&0.000 \\
W/o hair\cite{albiero2020face}&0.424 &0.713&0.589&0.289&-8.926&0.774&0.779&0.809&0.005&0.780&0.881&0.875&0.899&0.006&0.731\\
IVE(g)\cite{terhorst2019suppressing} &0.818&0.813& 0.833&\underline{0.005}&\underline{0.833}&0.912&0.884&0.910&0.028&0.391&0.952&0.926&0.951&0.026&0.071\\
\midrule
PASS-g &0.751 &0.749&0.761&\textbf{0.002}&\textbf{0.847}&0.831&0.828&0.839&\textbf{0.003}&\textbf{0.857}&0.909&0.909&0.910&\textbf{0.00}&\textbf{0.956} \\
MultiPASS &0.699&0.713& 0.708&0.014&0.383&0.811&0.808&0.809&\textbf{0.003}&\underline{0.823}&0.879&0.883&0.881&\underline{0.004}&\underline{0.784}\\
\bottomrule
\end{tabular}}
\begin{table*}%
  \scriptsize
  \centering
\tabcrygenderbpcsupp
\caption{\small \textit{Gender} bias analysis of \textit{Crystalface} descriptors, and their transformed counterparts on IJB-C. TPR: overall True Positive rate, TPR\textsubscript{m}: male-male TPR, TPR\textsubscript{f}: female-female TPR. \textbf{Bold}=Best, \underline{Underlined}=Second best}
\label{tab:crygfull}
\end{table*}
\newcommand{\tabcryskintonebpcsupp}{
\begin{tabular}{c|ccccc|ccccc|ccccc}
\toprule
FPR  & \multicolumn{5}{c|}{$10^{-4}$} & \multicolumn{5}{c|}{$10^{-3}$} & \multicolumn{5}{c}{$10^{-2}$} \\
\midrule
Network & TPR\textsubscript{l}&TPR\textsubscript{d}&TPR & Bias ($\downarrow$)& BPC\textsubscript{st}$(\uparrow)$ & TPR\textsubscript{l}&TPR\textsubscript{d}&TPR & Bias ($\downarrow$)& BPC\textsubscript{st}$(\uparrow)$& TPR\textsubscript{l}&TPR\textsubscript{d}&TPR & Bias ($\downarrow$)& BPC\textsubscript{st}$(\uparrow)$\\
\midrule
Crystalface\cite{ranjan2019fast} &0.912&0.864& 0.910 &0.048& 0.000 &0.948 &0.921&0.951 & 0.027&0.000 &0.974 &0.963&0.974 &0.011& 0.000\\
IVE(s)\cite{terhorst2019suppressing} &0.912&0.862& 0.910 &0.050& -0.041 &0.949&0.911& 0.951 &0.038& -0.407 &0.975&0.953& 0.974 &0.022& -1.000\\
\midrule
PASS-s &0.850&0.818& 0.844 &\underline{0.032}& \underline{0.261} & 0.913&0.906&0.914 &\underline{0.007}& \underline{0.702} &0.962&0.953& 0.919 &\underline{0.009}& \underline{0.125} \\
MultiPASS &0.826&0.838& 0.809 &\textbf{0.012}& \textbf{0.639} &0.907&0.907& 0.881 &\textbf{0.000}& \textbf{0.927} &0.953 &0.953&0.968 &\textbf{0.000}& \textbf{0.994}\\
\bottomrule
\end{tabular}}
\begin{table*}%
  \scriptsize
  \centering
\tabcryskintonebpcsupp
\caption{\small \textit{Skintone} bias analysis of \textit{Crystalface} descriptors, and their transformed counterparts on IJB-C. TPR: overall True Positive rate, TPR\textsubscript{l}: light-light TPR, TPR\textsubscript{d}: dark-dark TPR. \textbf{Bold}=Best, \underline{Underlined}=Second best}  
\label{tab:crystfull}
\end{table*}
\begin{table*}
\scriptsize
\centering
\begin{tabular}{cccccc|ccccc|ccccc}
\toprule
FPR & \multicolumn{5}{c|}{$10^{-4}$}& \multicolumn{5}{c|}{$10^{-3}$} & \multicolumn{5}{c}{$10^{-2}$}\\
\midrule
Method& TPR\textsubscript{l} & TPR\textsubscript{med}& TPR\textsubscript{d}& Avg & STD ($\downarrow$)& TPR\textsubscript{l} & TPR\textsubscript{med}& TPR\textsubscript{d}& Avg & STD ($\downarrow$)& TPR\textsubscript{l} & TPR\textsubscript{med}& TPR\textsubscript{d}& Avg & STD ($\downarrow$)\\
\midrule
Crystalface &0.912&0.912&0.864&0.896&0.023 &0.948&0.939&0.921&0.936&0.011 &0.974&0.964&0.963&0.967&0.005 \\
IVE(s) &0.912&0.899&0.862&0.891& 0.021&0.949&0.946&0.911&0.935&0.017 &0.975&0.968&0.953&0.965&0.009\\
\midrule
PASS-s (ours) &0.850&0.861&0.818&0.843&\underline{0.018} &0.913&0.909&0.906&0.909&\underline{0.003} &0.962&0.957&0.953&0.957&\underline{0.004}\\
MultiPASS (ours) &0.826&0.838&0.838&0.834&\textbf{0.006} &0.907&0.908&0.907&0.907&\textbf{0.0005} &0.953&0.952&0.953&0.953&\textbf{0.0005}\\
\bottomrule
\end{tabular}
\caption{\small Average and Standard deviation (STD) among the verification TPRs of light-light pairs, medium-medium pairs and dark-dark pairs. TPR: overall True Positive rate, TPR\textsubscript{l}: light-light TPR, TPR\textsubscript{med}: medium-medium TPR, TPR\textsubscript{d}: dark-dark TPR. \textbf{Bold}=Best, \underline{Underlined}=Second best}
\label{tab:std}
\end{table*}
\begin{table*}
\scriptsize
\centering
\begin{tabular}{cccccc|ccccc|ccccc}
\toprule
FPR & \multicolumn{5}{c|}{$10^{-5}$}& \multicolumn{5}{c|}{$10^{-4}$} & \multicolumn{5}{c}{$10^{-3}$}\\
\midrule
Method& TPR\textsubscript{m} & TPR\textsubscript{f}& TPR& Bias $(\downarrow)$ & BPC\textsubscript{g} ($\uparrow$)& TPR\textsubscript{m} & TPR\textsubscript{f}& TPR& Bias $(\downarrow)$& BPC\textsubscript{g} ($\uparrow$)& TPR\textsubscript{m} & TPR\textsubscript{f}& TPR& Bias $(\downarrow)$& BPC\textsubscript{g} ($\uparrow$)\\
\midrule
Crystalface + TPE & 0.883 & 0.838 &0.875 &0.045 &0.000 &0.925 & 0.891 &0.924 & 0.034 &0.000 &0.962&0.939& 0.959&0.023 &0.000 \\
PASS-g + TPE &0.797& 0.764&0.800&\textbf{0.033} &\textbf{0.181} & 0.875 & 0.843&0.875 & \textbf{0.032}&\textbf{0.006} &0.929&0.915&0.930&\textbf{0.014}&\textbf{0.361}\\
\bottomrule
\end{tabular}
\caption{ IJB-C 1:1 verification results after applying TPE on face descriptors from Crystalface and its PASS-g counterpart. TPR: overall True Positive rate, TPR\textsubscript{m}: male-male TPR, TPR\textsubscript{f}: female-female TPR.}
\label{tab:tpe}
\end{table*}

\section{Hair obscuring - Similar to \cite{albiero2020face}}
\label{sec:bowyer}
\begin{figure}
\centering
{\includegraphics[width=0.9\linewidth]{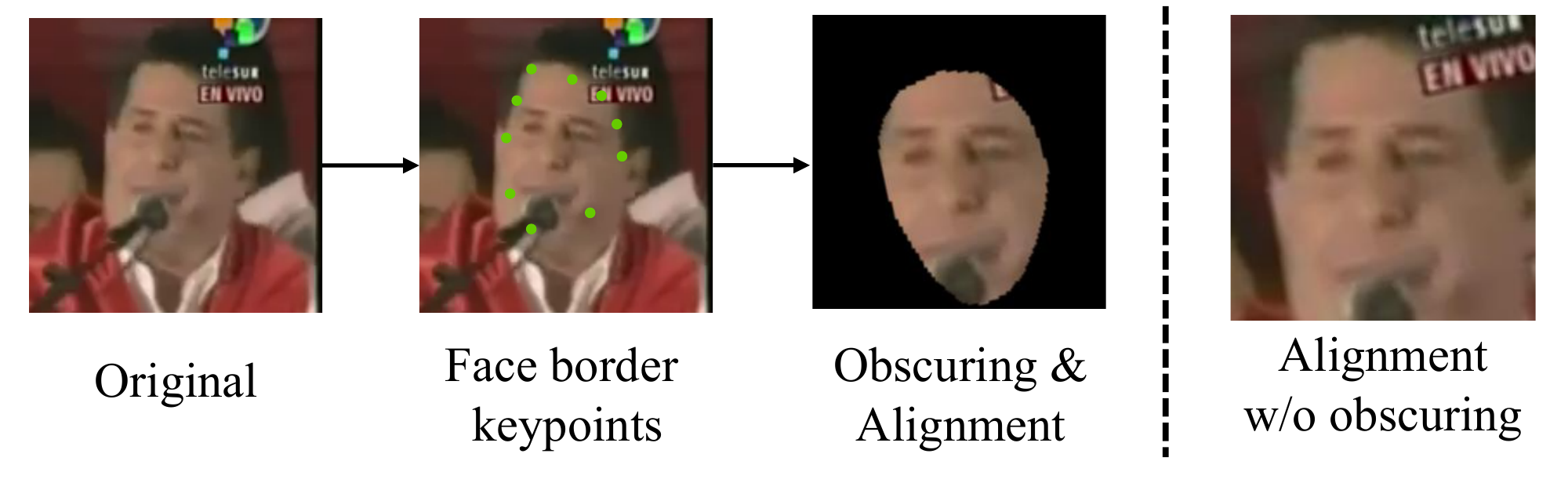}}
\caption{\small Our method for obscuring hair (Similar to \cite{albiero2020face}). On the right, we show an aligned image without obscuring hair.}
\label{fig:bowyerbaseline}
\vspace{-6pt}
\end{figure}

In \cite{albiero2020face}, it is shown that after obscuring hair in facial images, the resulting face descriptors extracted using Arcface demonstrate lower gender bias.  However, such experiments are only performed on datasets with clean frontal faces in MORPH \cite{ricanek2006morph} and Notre-Dame \cite{phillips2005overview} datasets. The authors used a segmentation network \cite{yu2018bisenet} to obscure the hair. But, in complex datasets, e.g., IJB-C containing varied and cluttered poses, segmenting out hair region is non-trivial and hard to perform. Instead, we compute the face border keypoints using \cite{ranjan2017all} and obscure all the regions outside the polygon formed by these keypoints. Our hair obscuring pipeline is presented in Fig \ref{fig:bowyerbaseline}. Note that, \cite{albiero2020face} proposes hair-obscuring as a possible approach to specifically mitigate gender-bias, and not skintone bias. So, we do not evaluate the effect of hair-obscuring while analyzing skintone bias.

\section{Detailed results}
\label{sec:res}
\subsection{PASS with Arcface}
\begin{figure}
{
\centering
\subfloat[]{\includegraphics[width=0.5\linewidth]{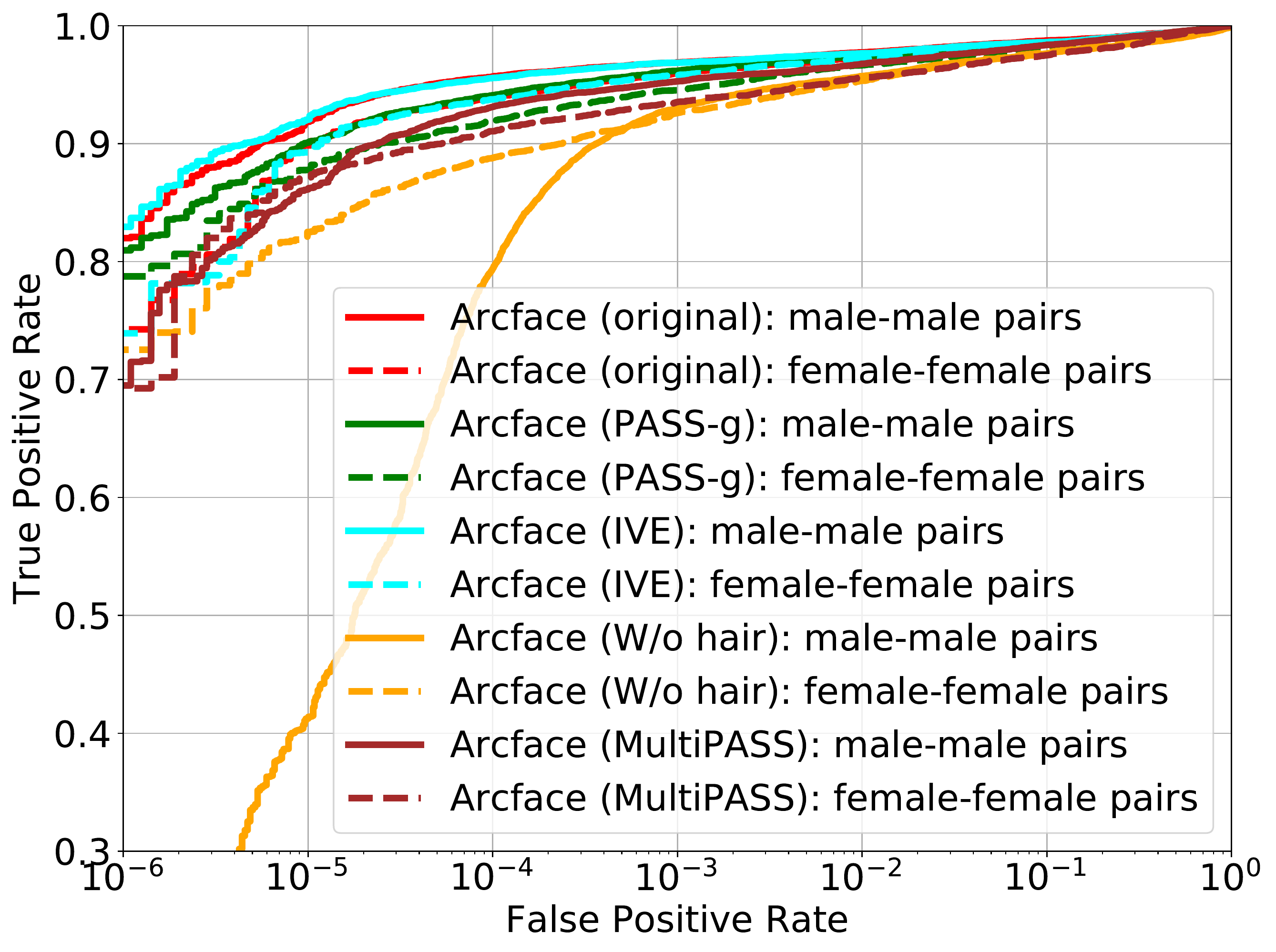}\label{fig:gwarcface}}
\subfloat[]{\includegraphics[width=0.5\linewidth]{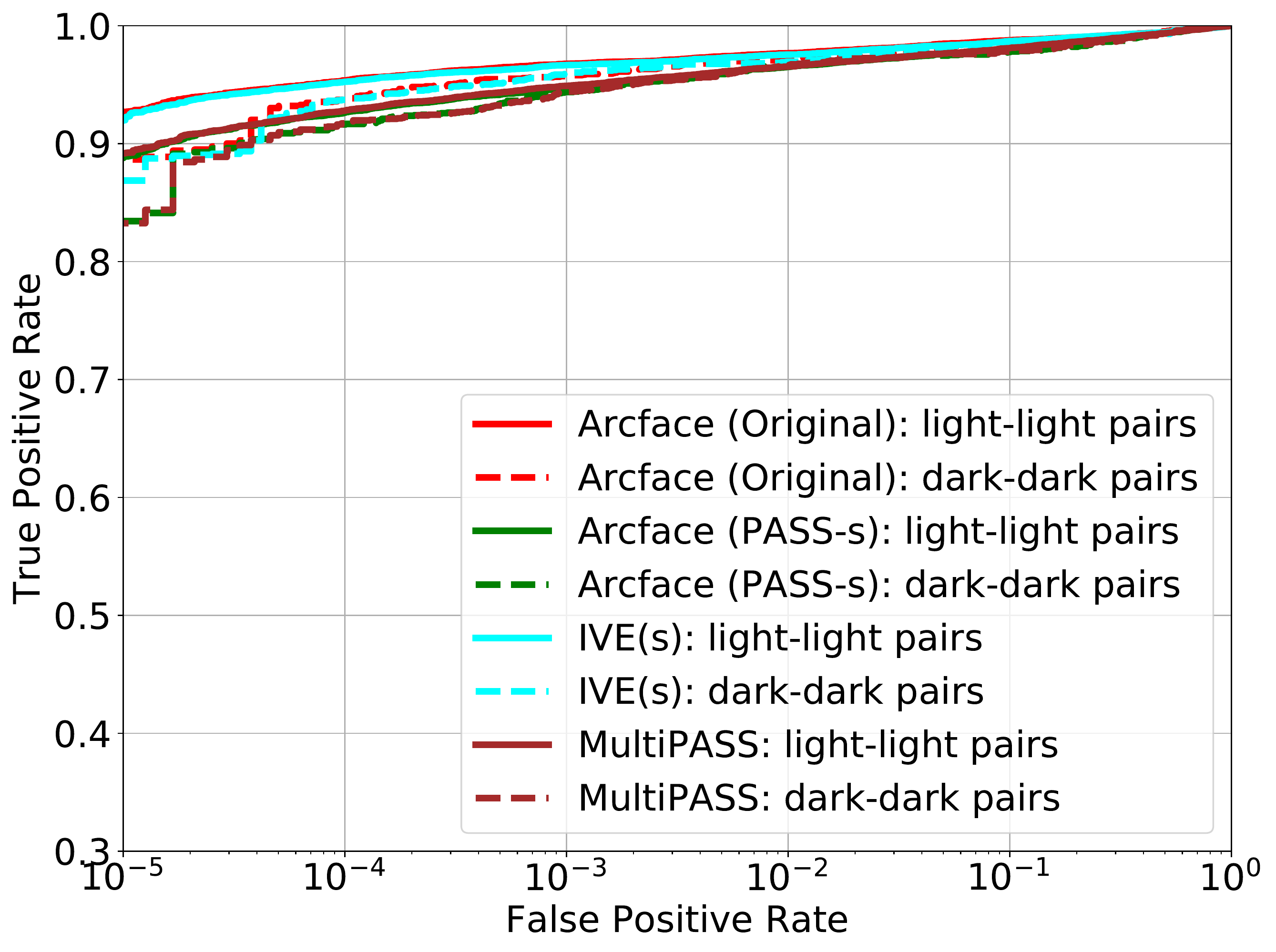}}
\caption{\small (a.) Gender-wise and (b.) Skintone-wise verification plots for Arcface descriptors and their de-biased counterparts on IJB-C}
\label{fig:gwstwarcface}
}
\end{figure}

\begin{figure}
{
\centering
\subfloat[]{\includegraphics[width=0.5\linewidth,height=3.15cm]{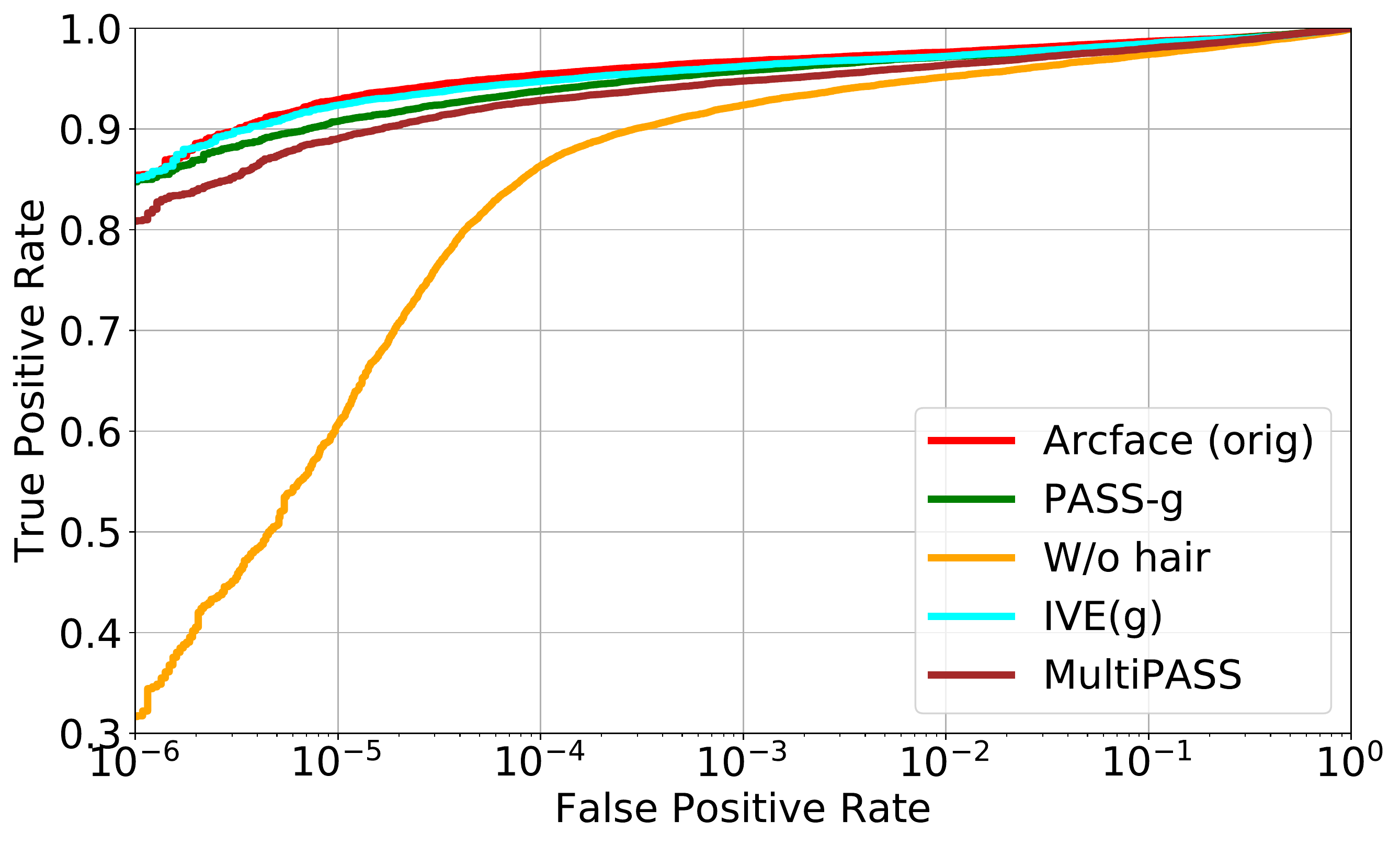}}
\subfloat[]{\includegraphics[width=0.5\linewidth,height=3.15cm]{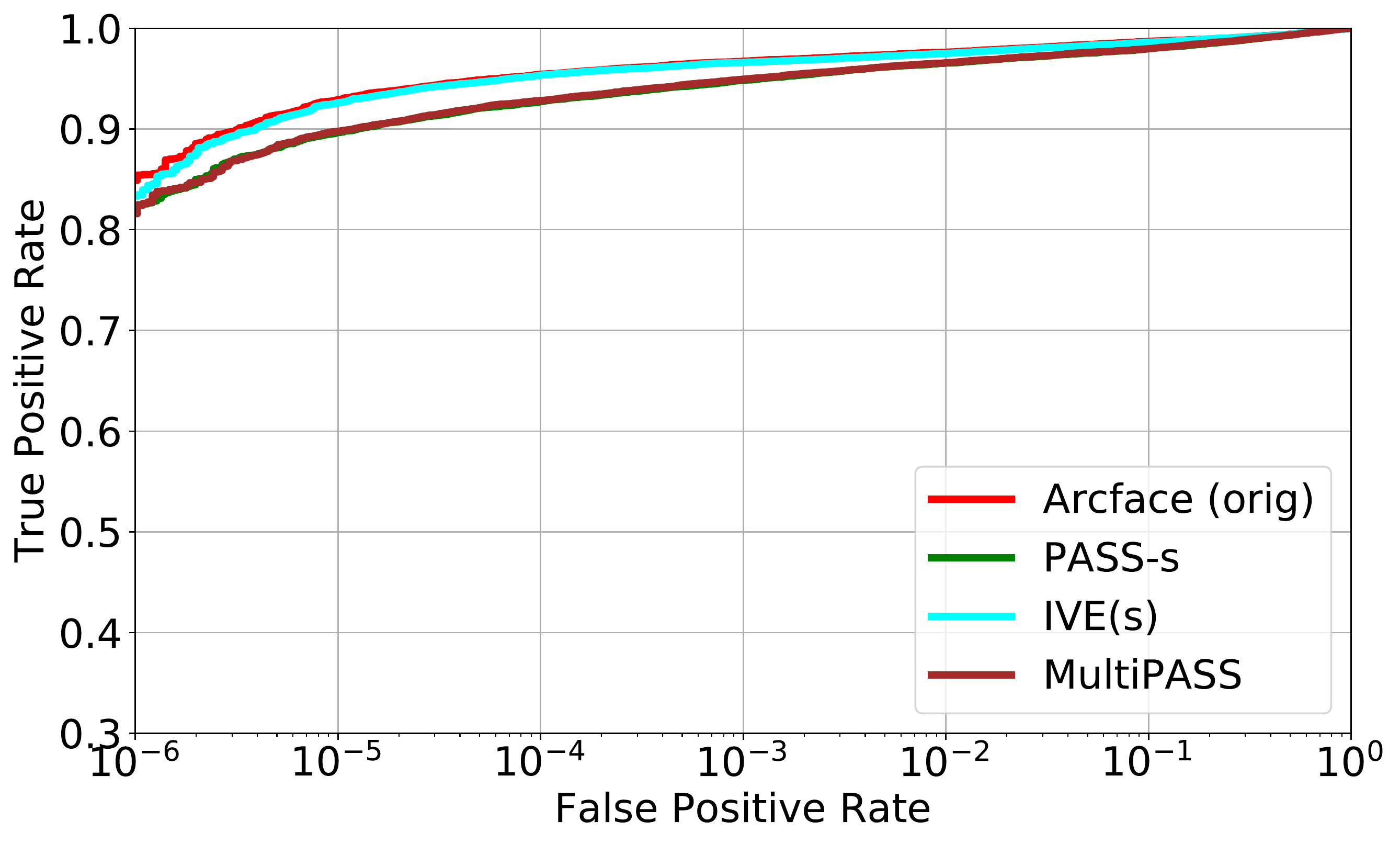}}
\caption{\small Overall IJB-C verification plots of Arcface along with  (a.) Gender-debiasing algorithms, (b.) Skintone-debiasing algorithms.}
\label{fig:overallarcface}
}
\end{figure}

For PASS/MultiPASS systems trained on Arcface descriptors, we provide the gender-wise and skintone-wise results in Table 2 and 3 respectively in the main paper. We also present the gender and skintone bias in Figure 6 in the main paper, and show that the PASS/MultiPASS systems outperform the IVE and hair-obscuring baselines at most FPRs. Here, we provide the gender-wise and skintone-wise verification plots for all the methods used to de-bias Arcface descriptors in Figure \ref{fig:gwstwarcface}. Additionally, we also provide the overall verification plots in Figure \ref{fig:overallarcface}.
\begin{figure}
\centering
{\includegraphics[width=\linewidth]{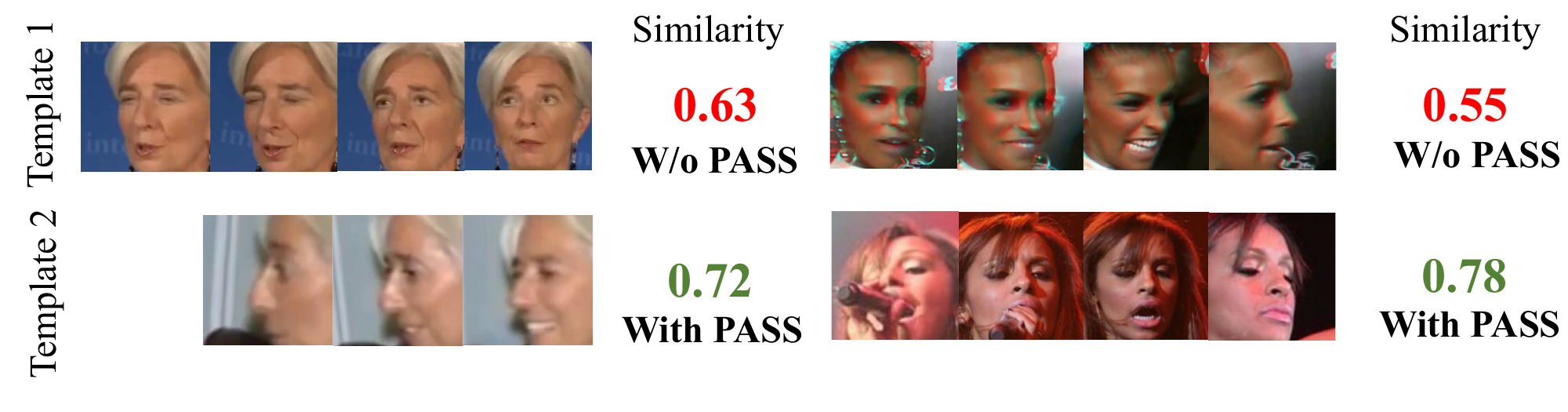}}
\caption{\small Examples of templates in IJB-C verification for which the average cosine similarity improved after PASS transformation.}
\label{fig:improved}
\vspace{-12pt}
\end{figure}

Although the main aim of using PASS-g is to reduce gender predictability in face descriptors, we find (in Fig. \ref{fig:gwarcface}) that the performance of female-female verification improves between FPR $10^{-5}$ and $10^{-6}$. In fact, we find several examples of template pairs which are verified between these FPRs, for both Arcface descriptors and their PASS-g counterparts. In such pairs, we find the average cosine similarity of images in templates that belong to the same female identity increases after the face descriptors are transformed using PASS-g.  We show two examples of such templates in Fig \ref{fig:improved}. 
\subsection{PASS with Crystalface}
\begin{figure}
{
\centering
\subfloat[]{\includegraphics[width=0.5\linewidth]{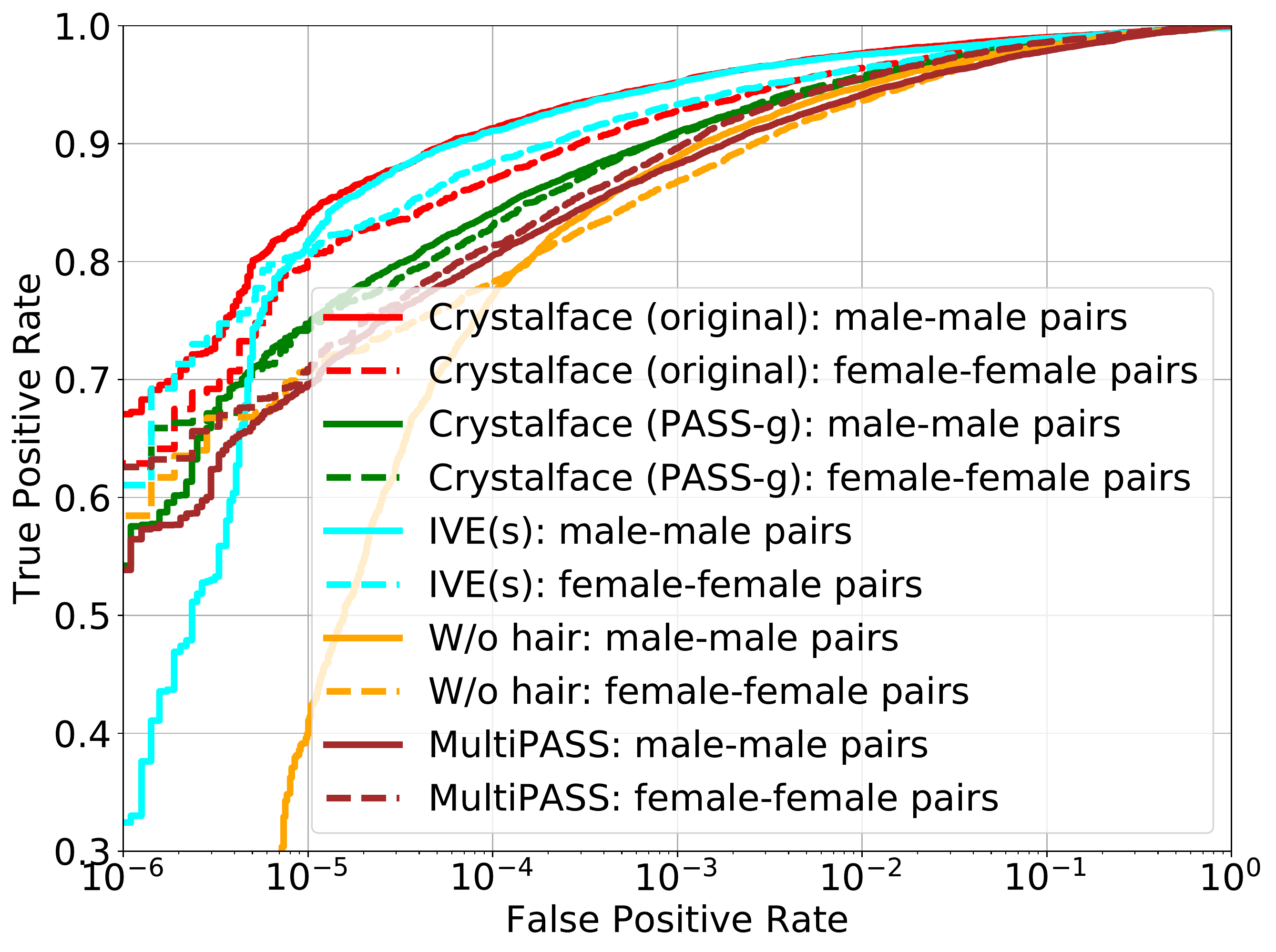}}
\subfloat[]{\includegraphics[width=0.5\linewidth]{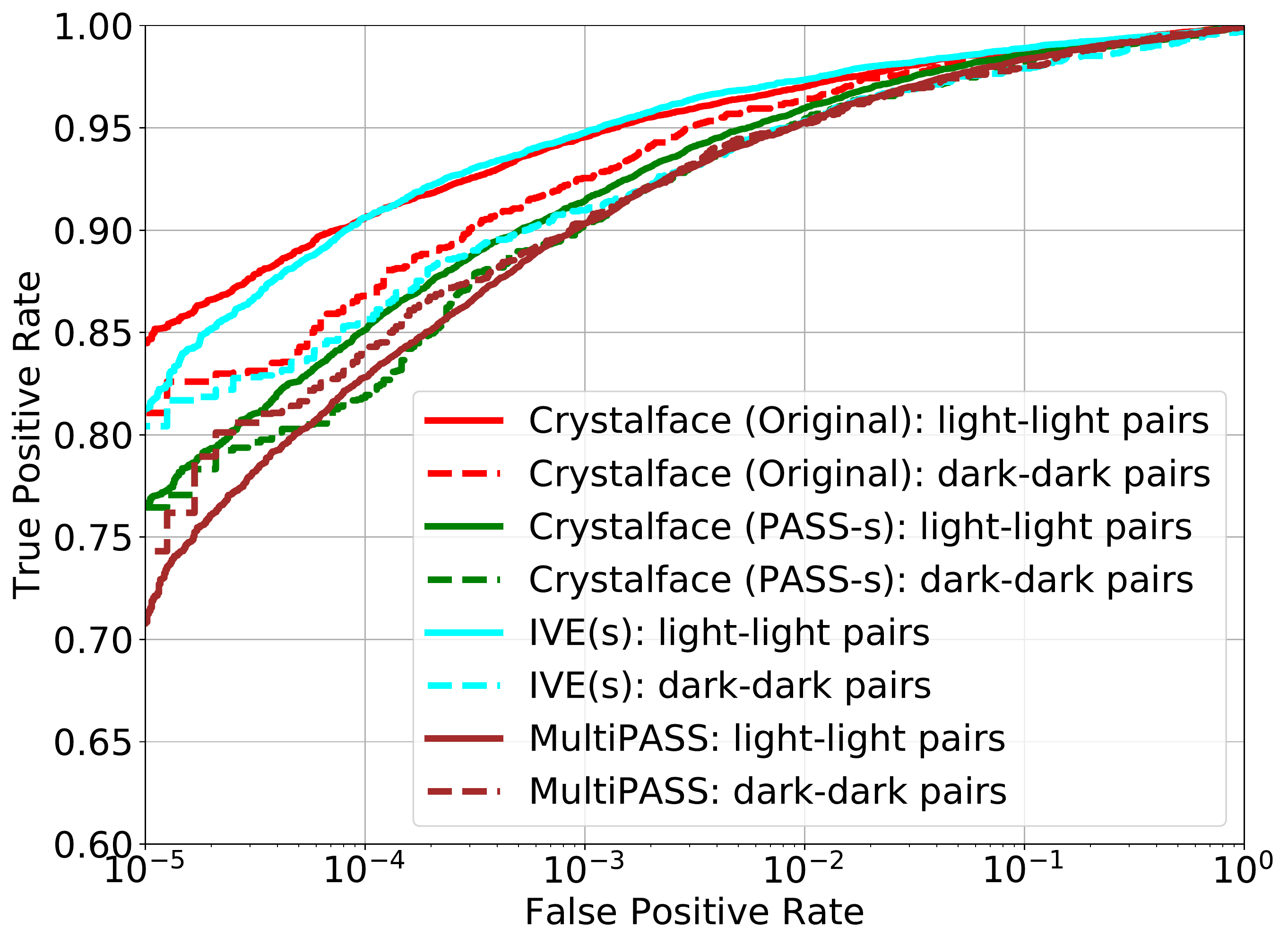}}
\caption{\small (a.) Gender-wise and (b.) Skintone-wise verification plots for Arcface descriptors and their de-biased counterparts on IJB-C}
\label{fig:gwstwcry}
}
\end{figure}

\begin{figure}
{
\centering
\subfloat[]{\includegraphics[width=0.5\linewidth,height=3.15cm]{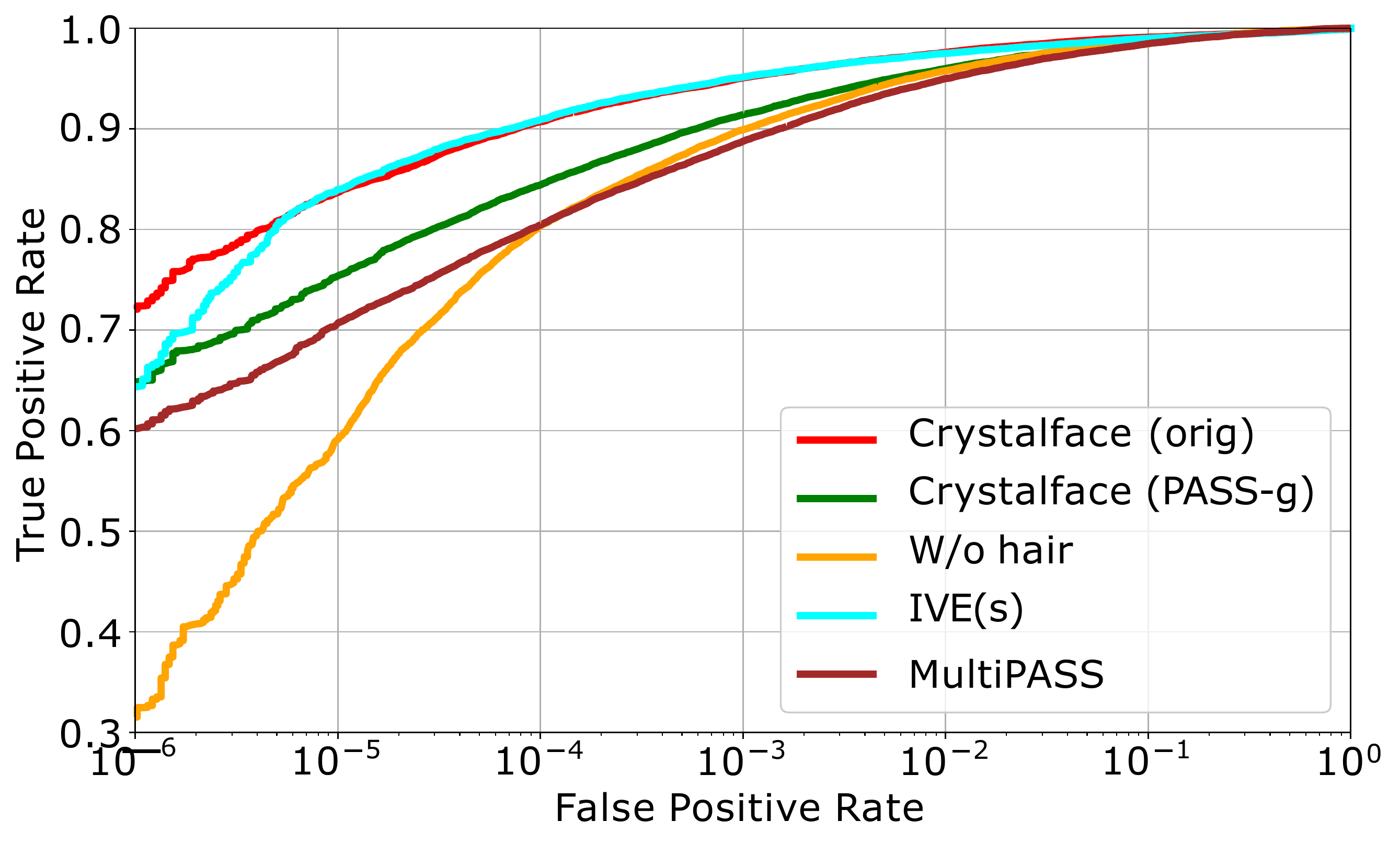}}
\subfloat[]{\includegraphics[width=0.5\linewidth,height=3.15cm]{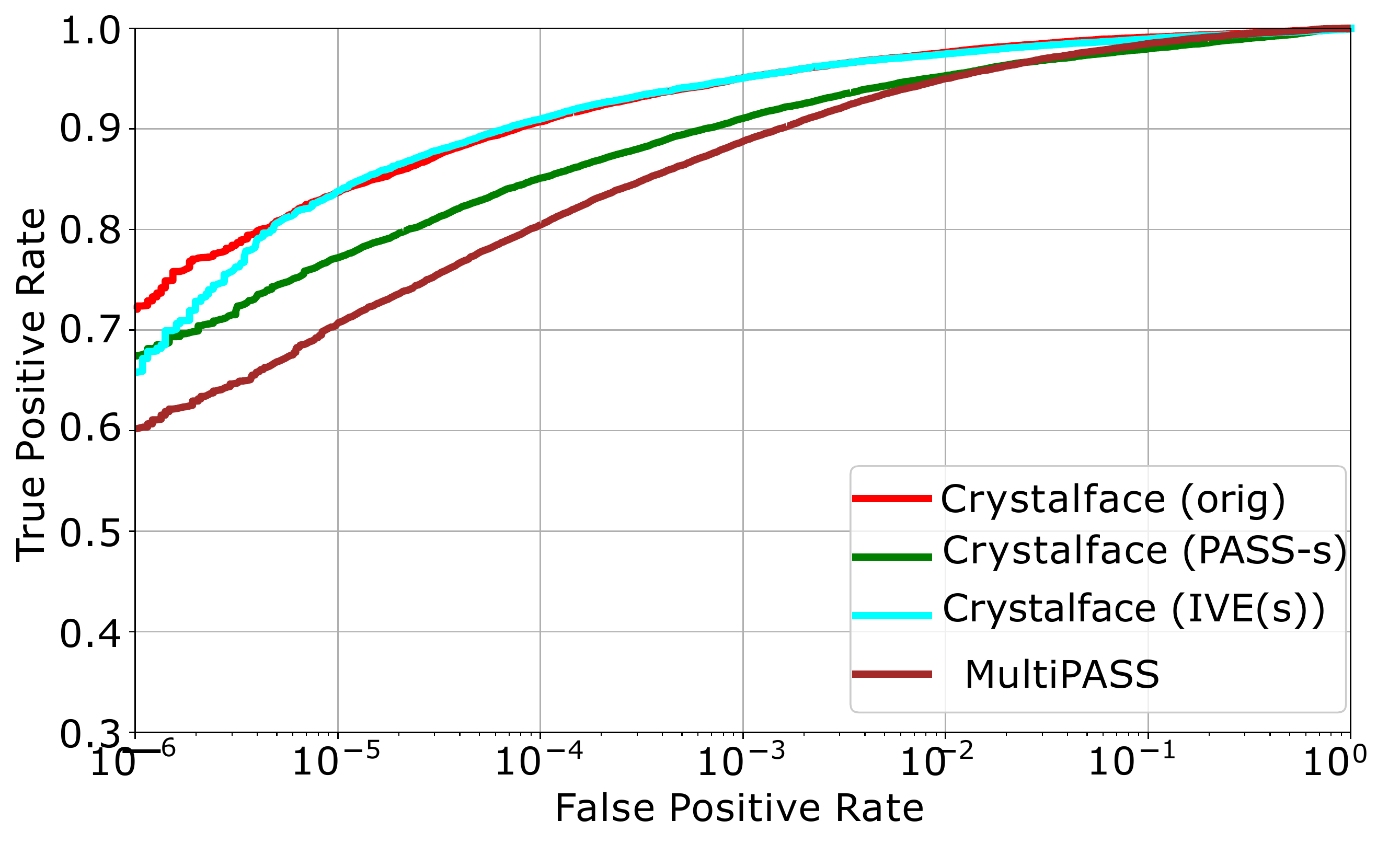}}
\caption{\small Overall IJB-C verification plots of Crystalface along with  (a.) Gender-debiasing algorithms, (b.) Skintone-debiasing algorithms.}
\label{fig:overallcrystalface}
}
\end{figure}

It can be inferred from Tables \ref{tab:hypgen} and \ref{tab:hypst} that descriptors from Crystalface demonstrate higher gender/skintone bias than those from Arcface. Therefore, we believe that de-biasing Crystalface descriptors is a better testing ground for de-biasing algorithms like PASS/MultiPASS. Moreover, this helps us assess the generalizability of  proposed PASS/MultiPASS systems. We provide the BPC values and overall TPRs of all the approaches for de-biasing Crystalface descriptors in Table 4 (for gender) and Table 5 (for skintone) in the main paper, and show that PASS/MultiPASS systems achieve higher BPC values than the baselines. Here, we provide the gender-wise and skintone-wise verification TPRs (along with the corresponding bias values) in Tables \ref{tab:crygfull} and \ref{tab:crystfull} respectively. Moreover, we provide the gender-wise and skintone-wise verification plots for all the methods in Figure \ref{fig:gwstwcry}. Also, we provide the overall verification plots for all the methods in Figure \ref{fig:overallcrystalface}. It should be noted in Tables \ref{tab:crygfull} and \ref{tab:crystfull} that although IVE achieves higher overall TPRs, it hardly reduces bias, thus obtaining lower BPC values than PASS/MultiPASS systems.
\subsection{OAT v/s AET}
\begin{figure}
{\centering
\vspace{-0.4cm}
\subfloat[]{\includegraphics[width = 0.5\linewidth]{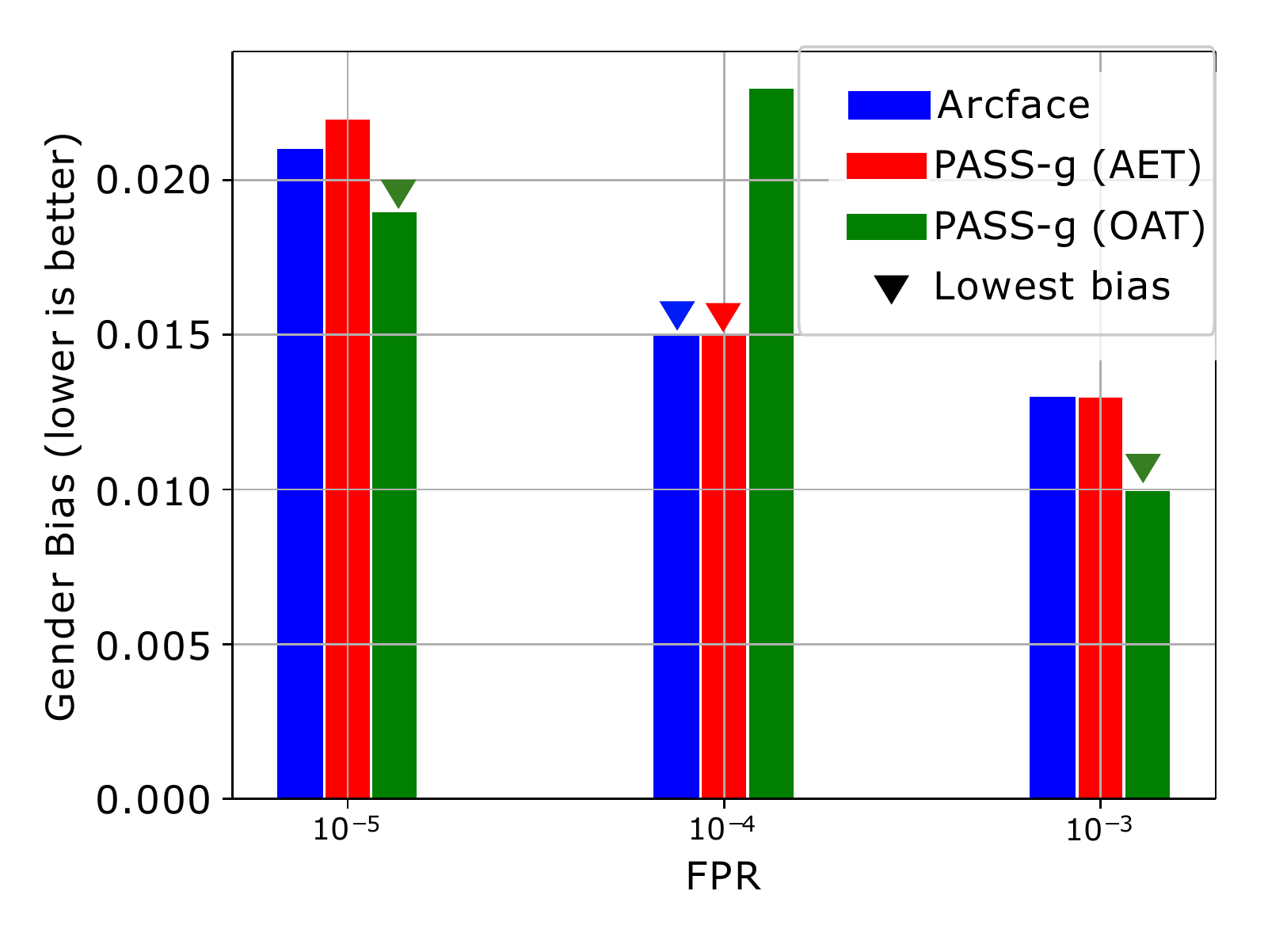}}
\subfloat[]{\includegraphics[width = 0.5\linewidth]{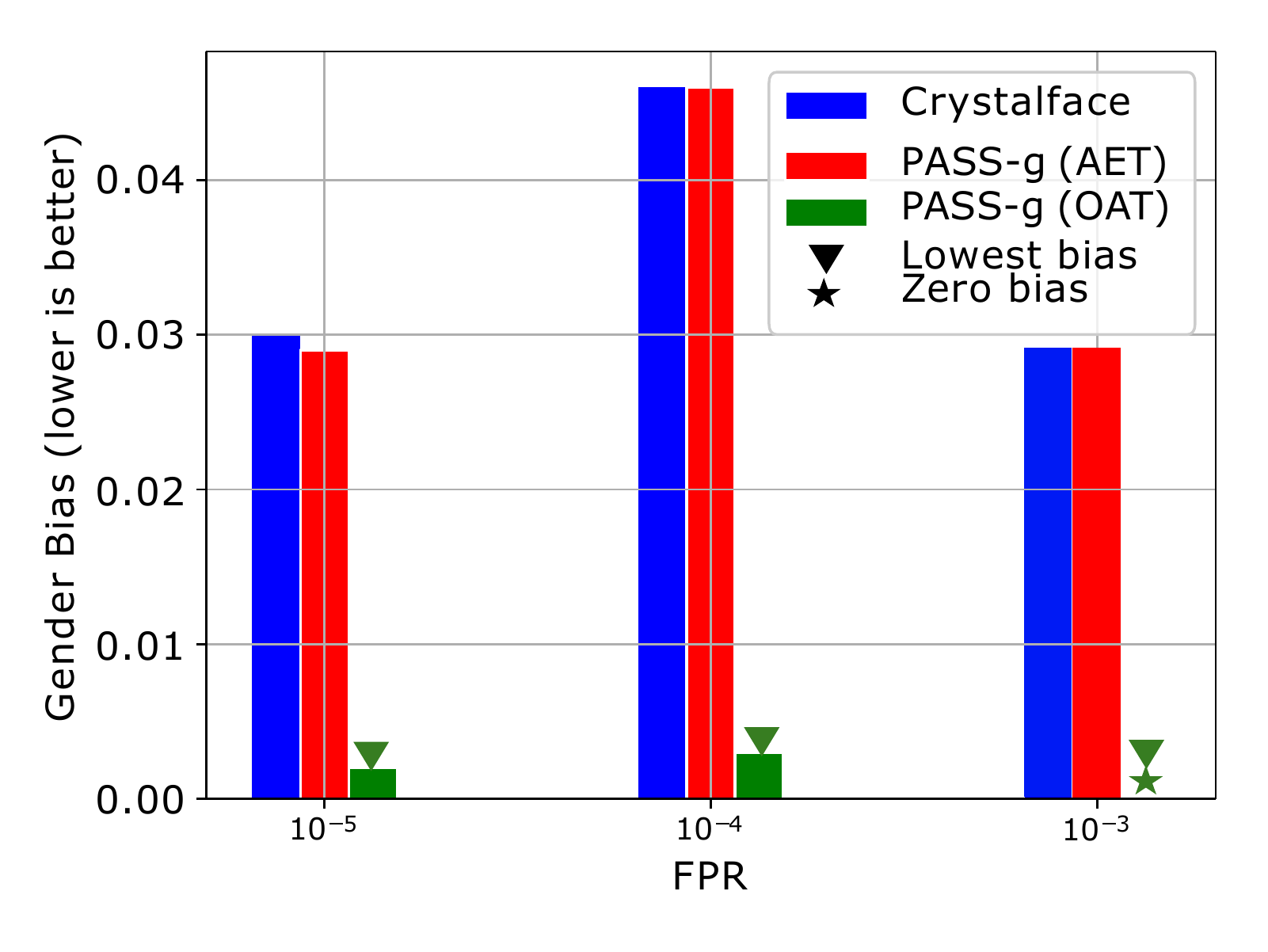}}
\caption{\small Comparison of bias for AET vs OAT in gender reduction on (a) Arcface, (b) Crystalface.} \vspace{-0.33cm}
\label{fig:novcomp}
}
\end{figure}

In Figure~\ref{fig:novcomp}, we visualize the results presented in Table 7 in the main paper.

\subsection{Results with multiple skintones}
\begin{figure}
{
\centering
\subfloat[Crystalface]{\includegraphics[width=0.5\linewidth]{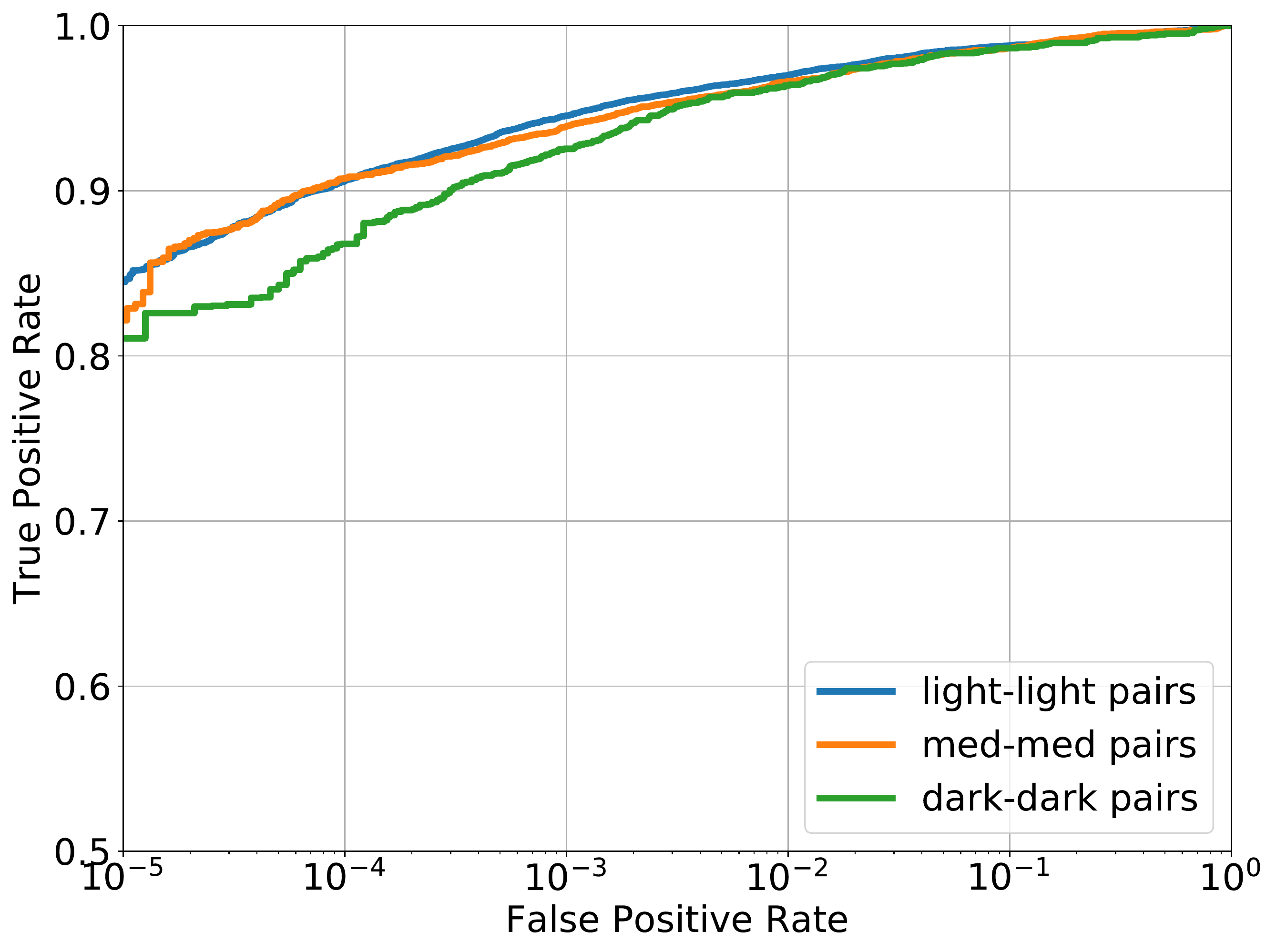}}
\subfloat[IVE(s) on Crystalface]{\includegraphics[width=0.5\linewidth]{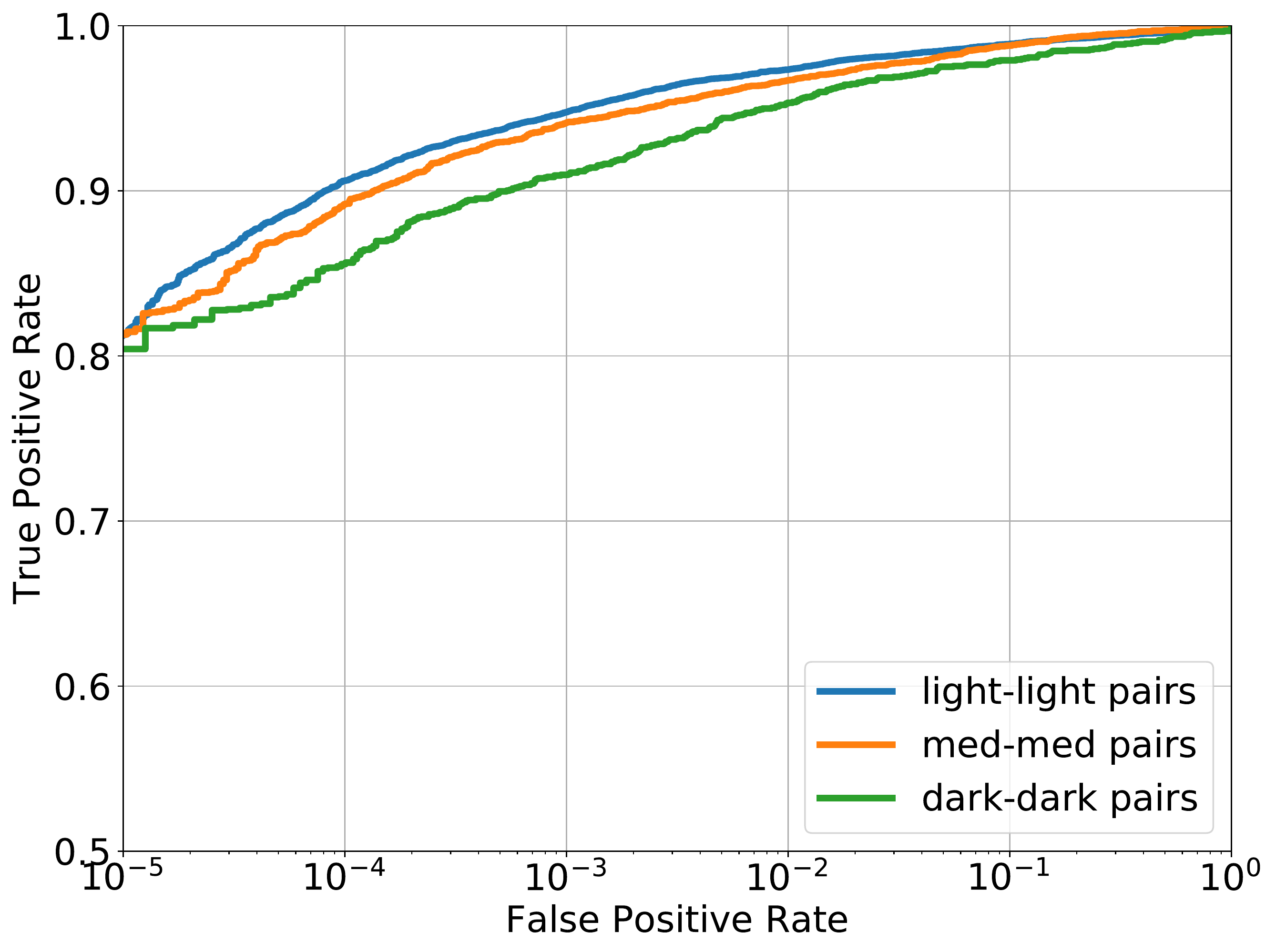}}\\
\subfloat[PASS-s on Crystalface]{\includegraphics[width=0.5\linewidth]{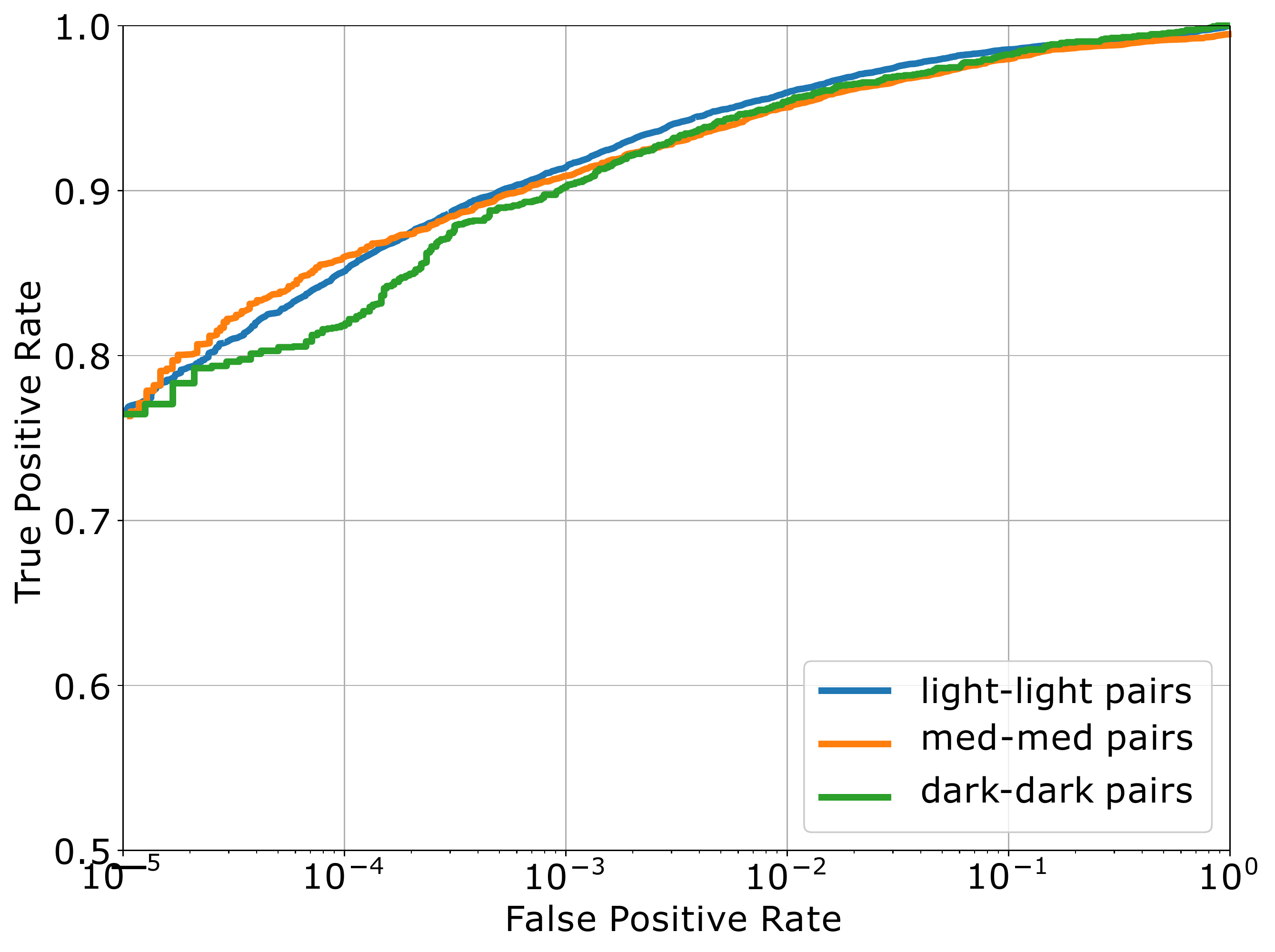}}
\subfloat[MultiPASS on Crystalface]{\includegraphics[width=0.5\linewidth]{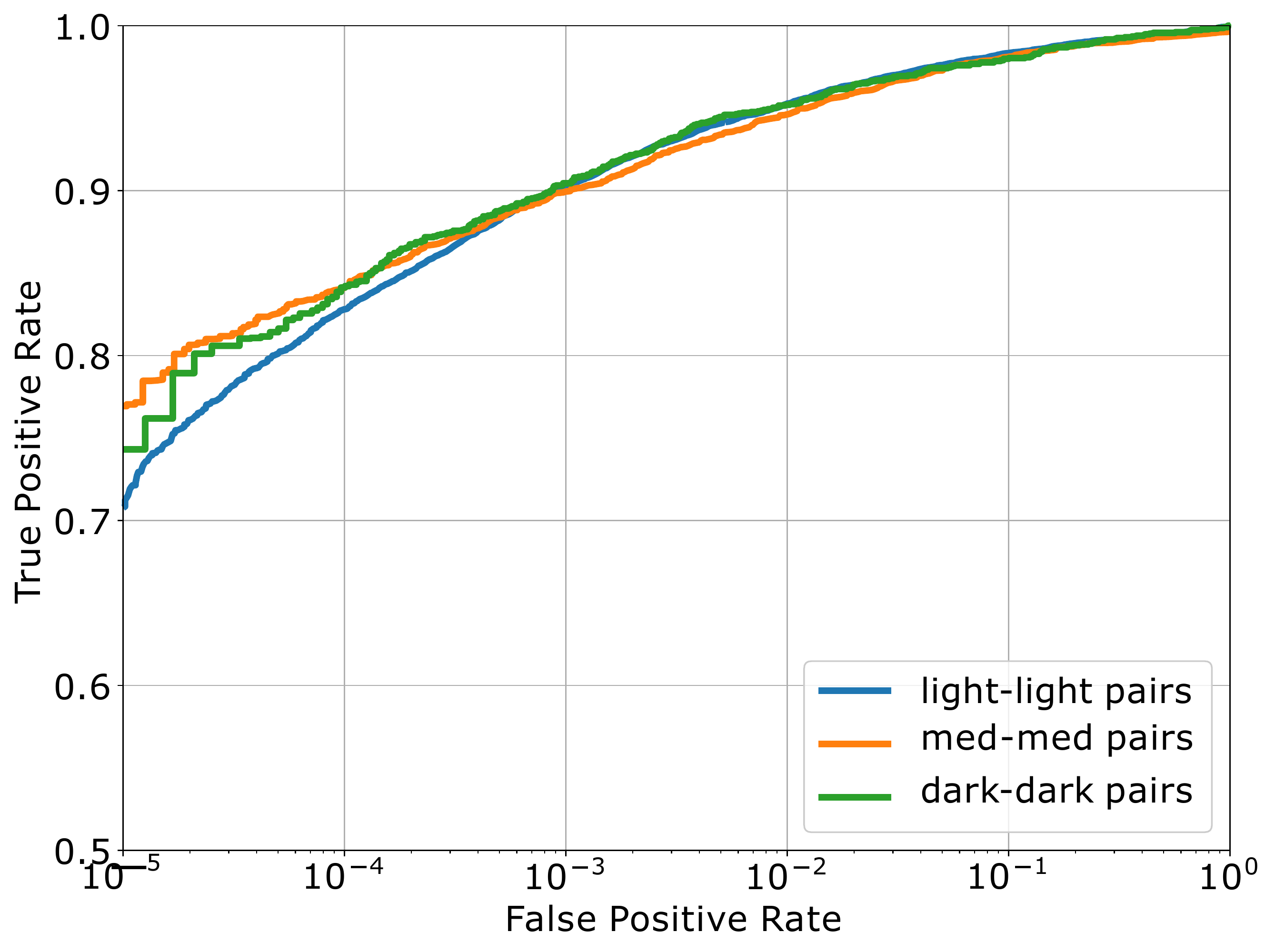}}
\caption{\small Skintone-wise verification plots for all three skintones on the IJB-C dataset for Crystalface descriptors and their skintone-debiased counterparts}
\label{fig:lmd}
}
\end{figure}

In Equations 1 and 2 in the main paper, we define bias as the absolute difference between the verification TPRs of two groups at a given FPR. However, it possible that a sensitive attribute consists of more than two categories. For instance, the skintone attribute consists of three categories: Light, medium, dark. In the main paper, we chose to define bias as the difference between the verification TPRs of light-light and dark-dark pairs at a given FPR. However, as shown in \cite{wang2020mitigating}, we can also define bias as the standard deviation (STD) among the verification TPRs of light-light pairs, medium-medium pairs and dark-dark pairs. In Table \ref{tab:std}, we report these STD values for our PASS-s and MultiPASS systems (and the corresponding baselines) trained on Crystalface descriptors, along with the average of the TPRs obtained for the three skintone categories. \textit{We find that our proposed PASS-s/MultiPASS systems obtain considerably lower STD than existing baselines, thus mitigating skintone bias.} We also provide the skintone-wise verification plots for all three skintones (light, medium and dark) on IJB-C dataset in Figure \ref{fig:lmd}
\section{Ablation experiments: Effect of $K, \lambda$ in PASS}
\label{sec:ablation}
\begin{figure}
{
\centering
\subfloat[PASS-g on Arcface ($\lambda=10$)]{\includegraphics[width=0.5\linewidth]{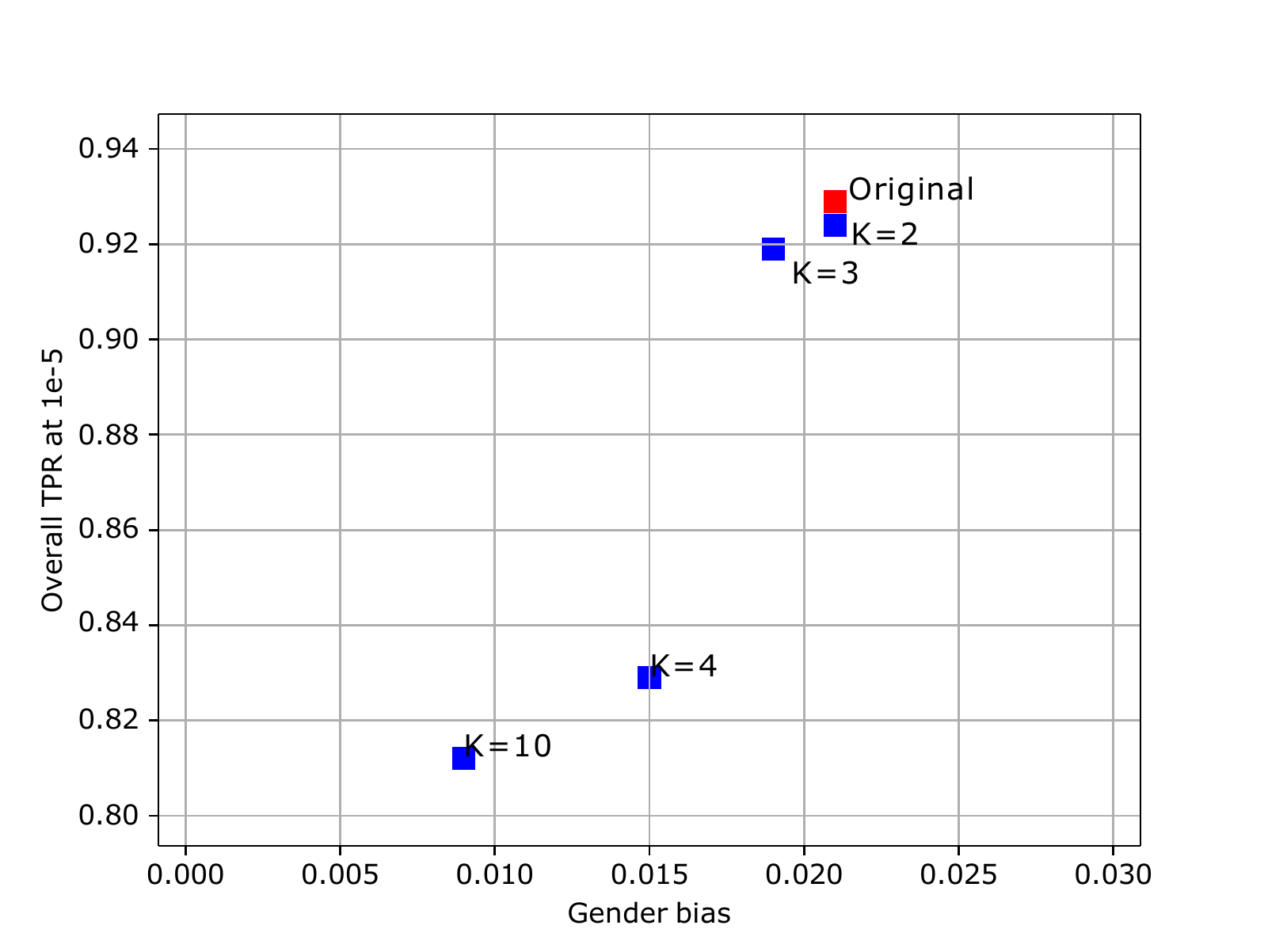}\label{fig:karcg}}
\subfloat[PASS-g on Crystalface ($\lambda=1$)]{\includegraphics[width=0.5\linewidth]{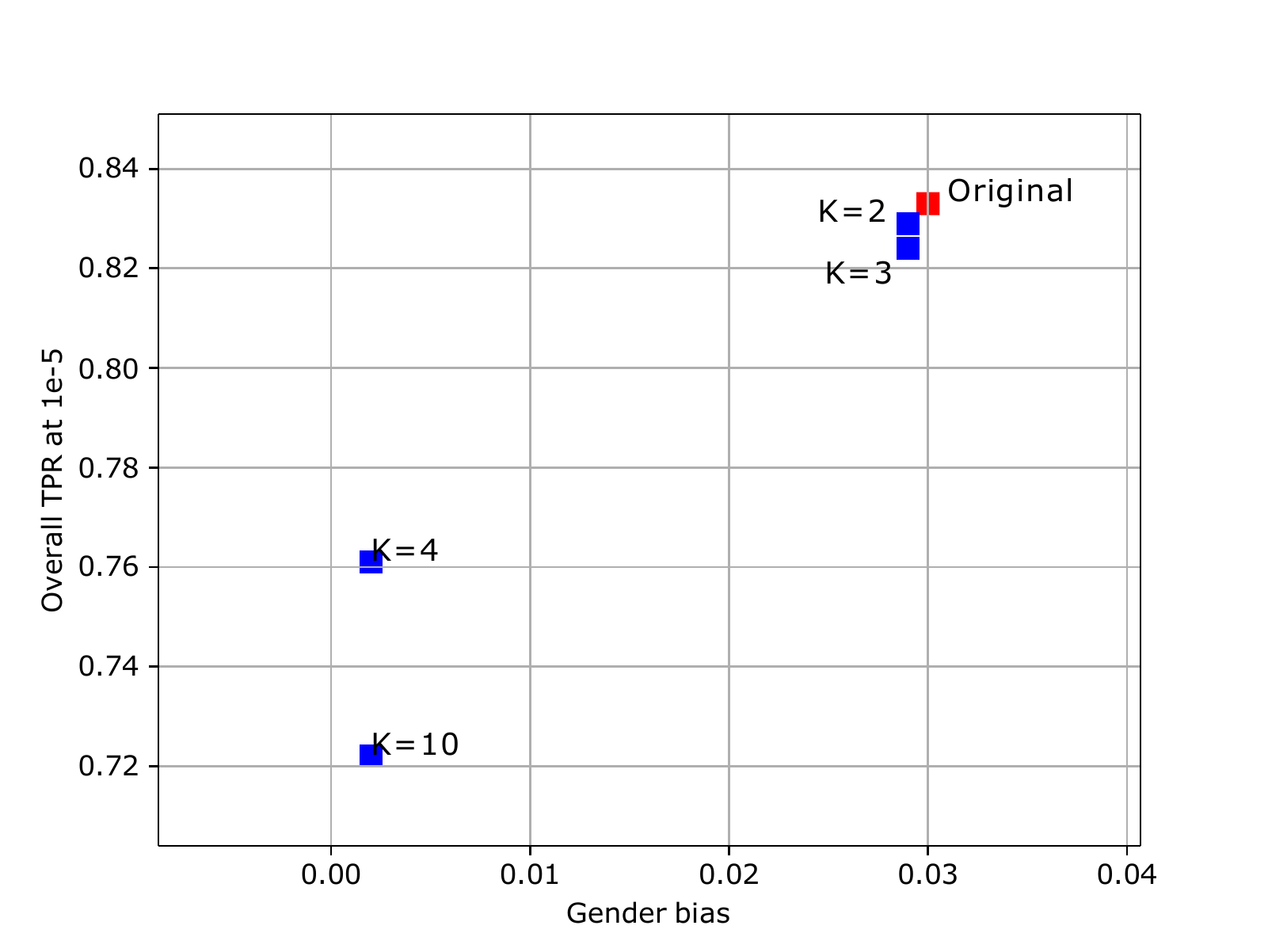}\label{fig:kcryg}}\\
\subfloat[PASS-s on Arcface ($\lambda=10$)]{\includegraphics[width=0.5\linewidth]{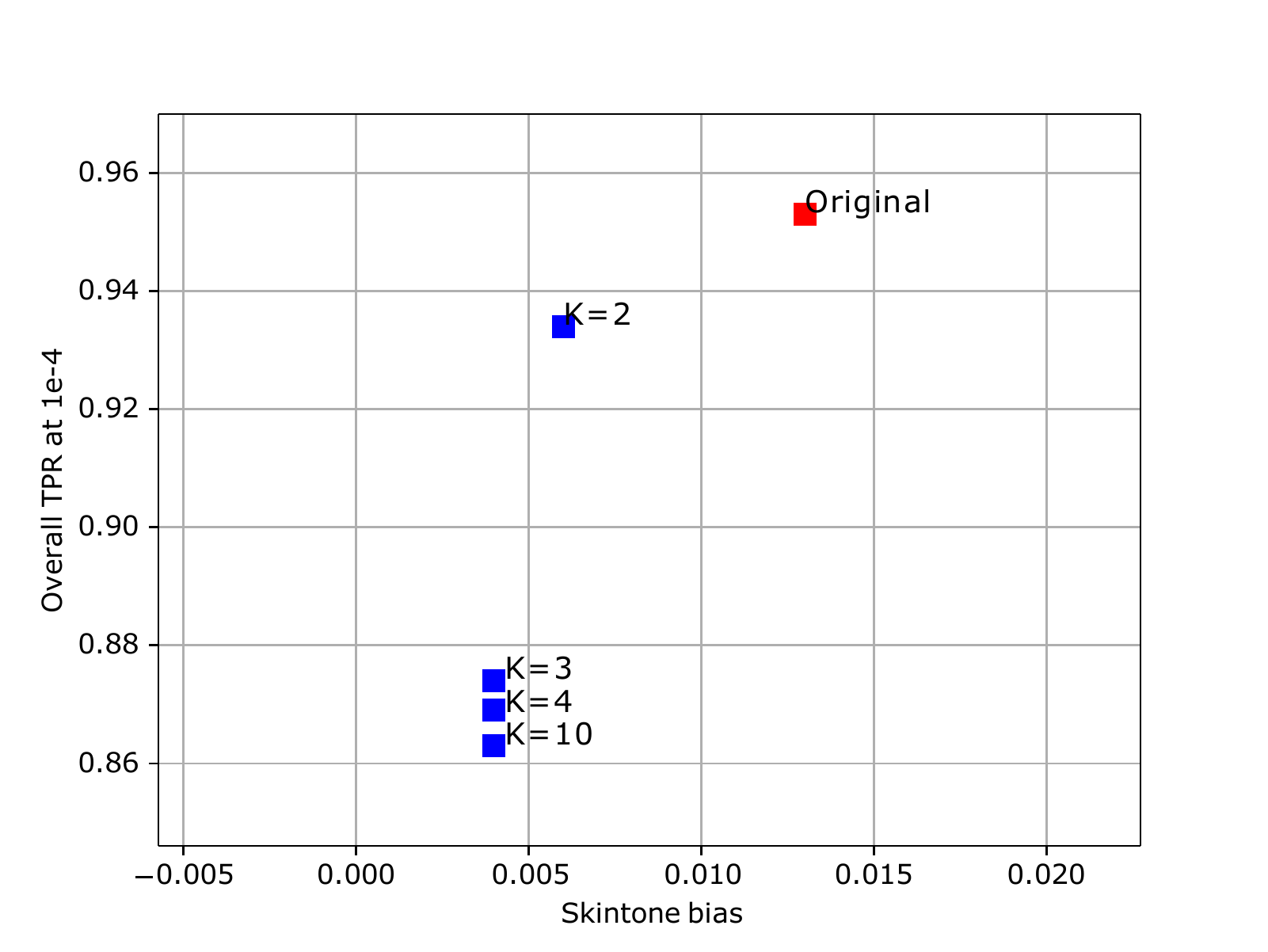}\label{fig:karcst}}
\subfloat[PASS-s on Crystalface ($\lambda=10$)]{\includegraphics[width=0.5\linewidth]{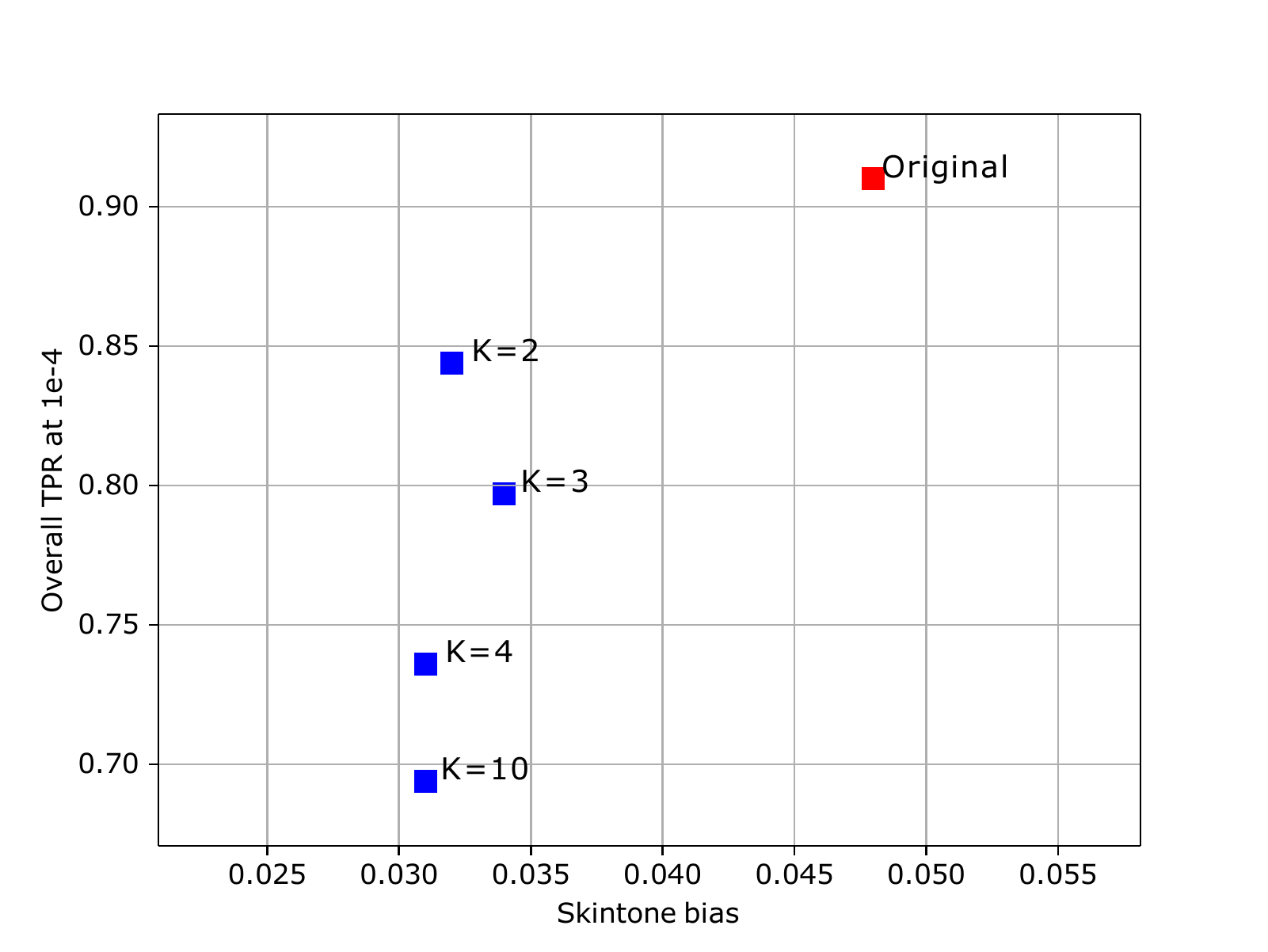}\label{fig:kcryst}}
\caption{\small Effect of varying $K$ (number of adversary classifiers in the ensemble $E$) in PASS systems }
\label{fig:abk}
}
\end{figure}

\begin{figure}
{
\centering
\subfloat[PASS-g on Arcface ($K=3$)]{\includegraphics[width=0.5\linewidth]{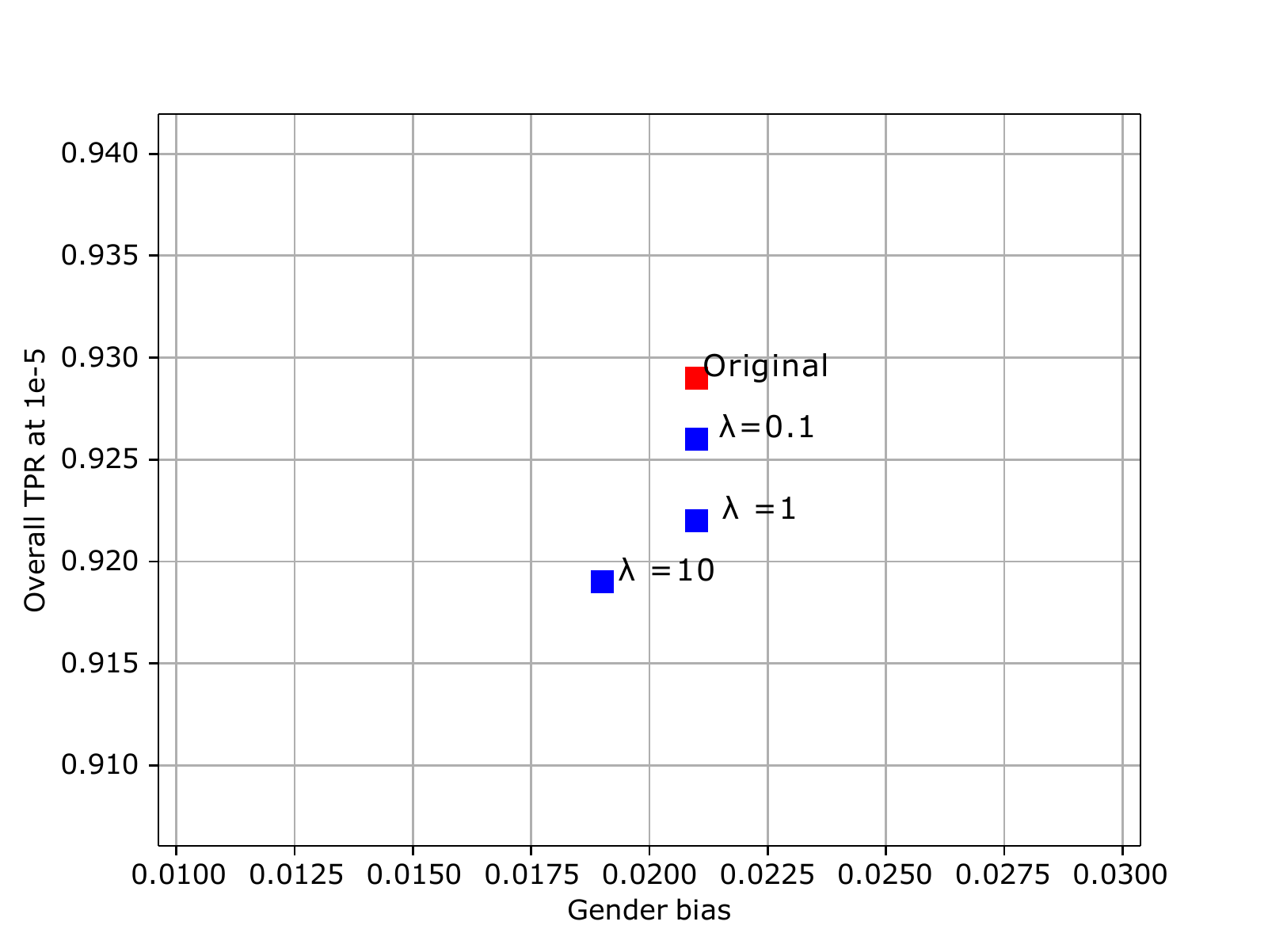}\label{fig:lamarcg}}
\subfloat[PASS-g on Crystalface ($K=4$)]{\includegraphics[width=0.5\linewidth]{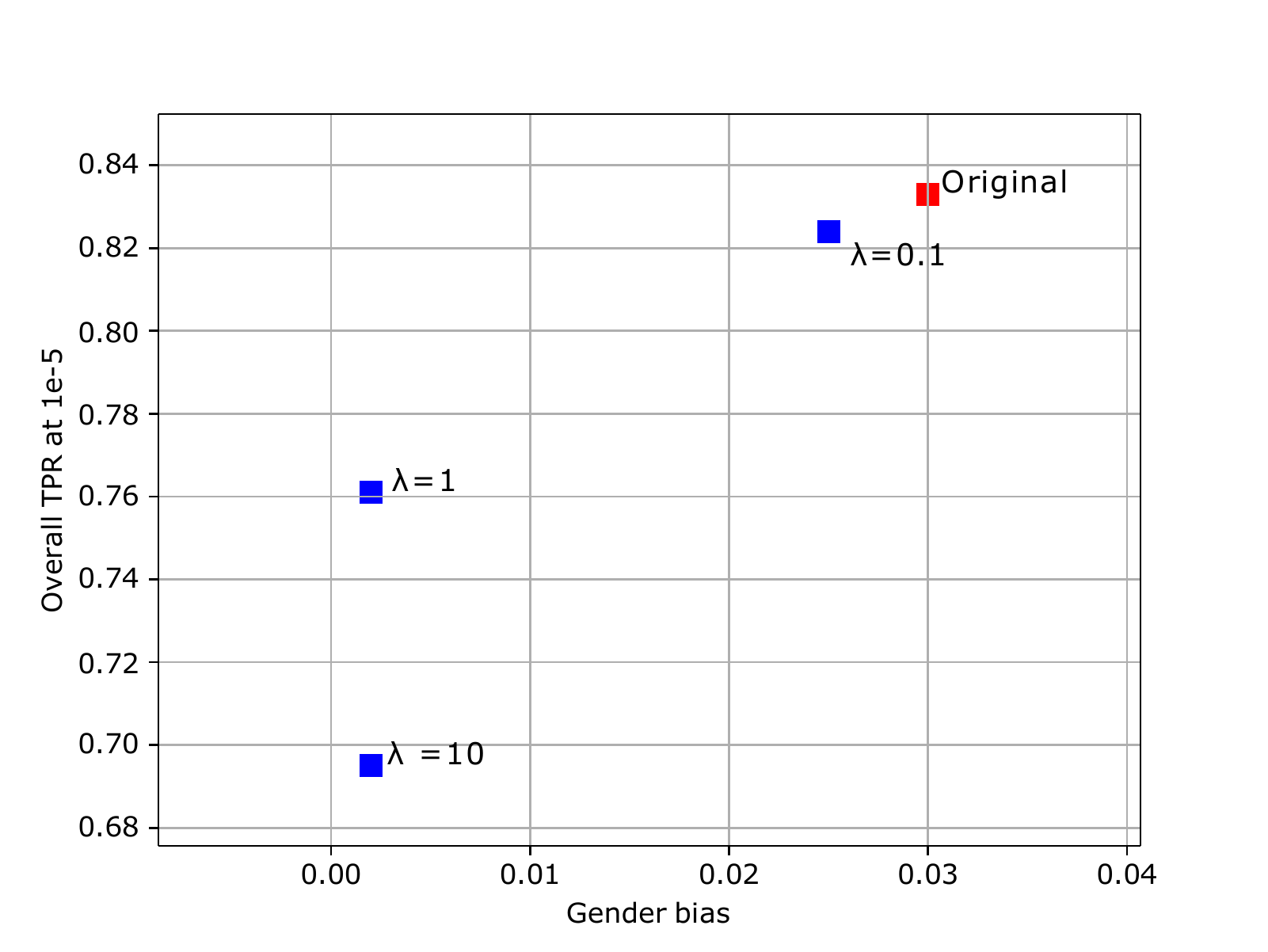}\label{fig:lamcryg}}\\
\subfloat[PASS-s on Arcface ($K=2$)]{\includegraphics[width=0.5\linewidth]{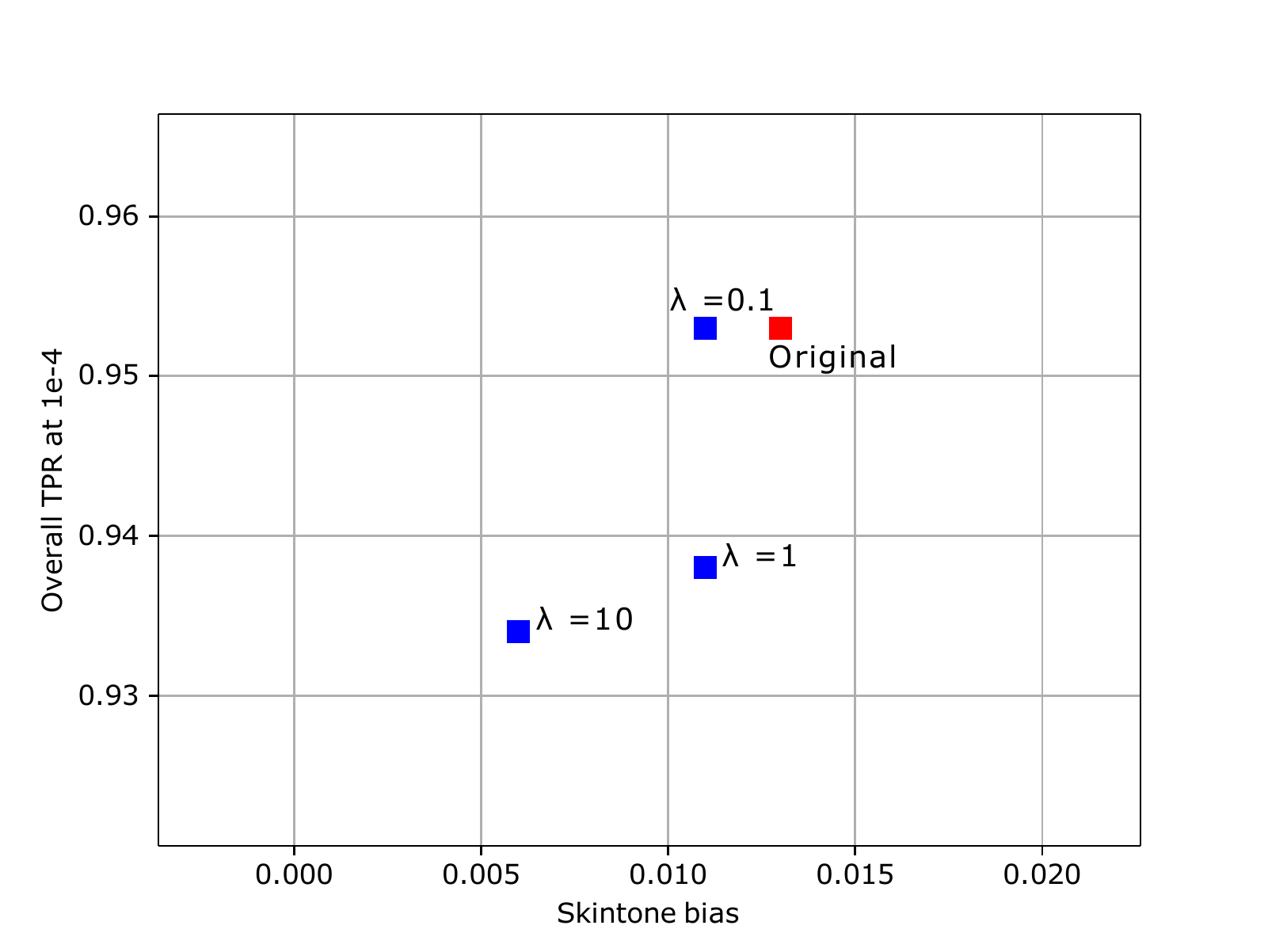}\label{fig:lamarcst}}
\subfloat[PASS-s on Crystalface ($K=2$)]{\includegraphics[width=0.5\linewidth]{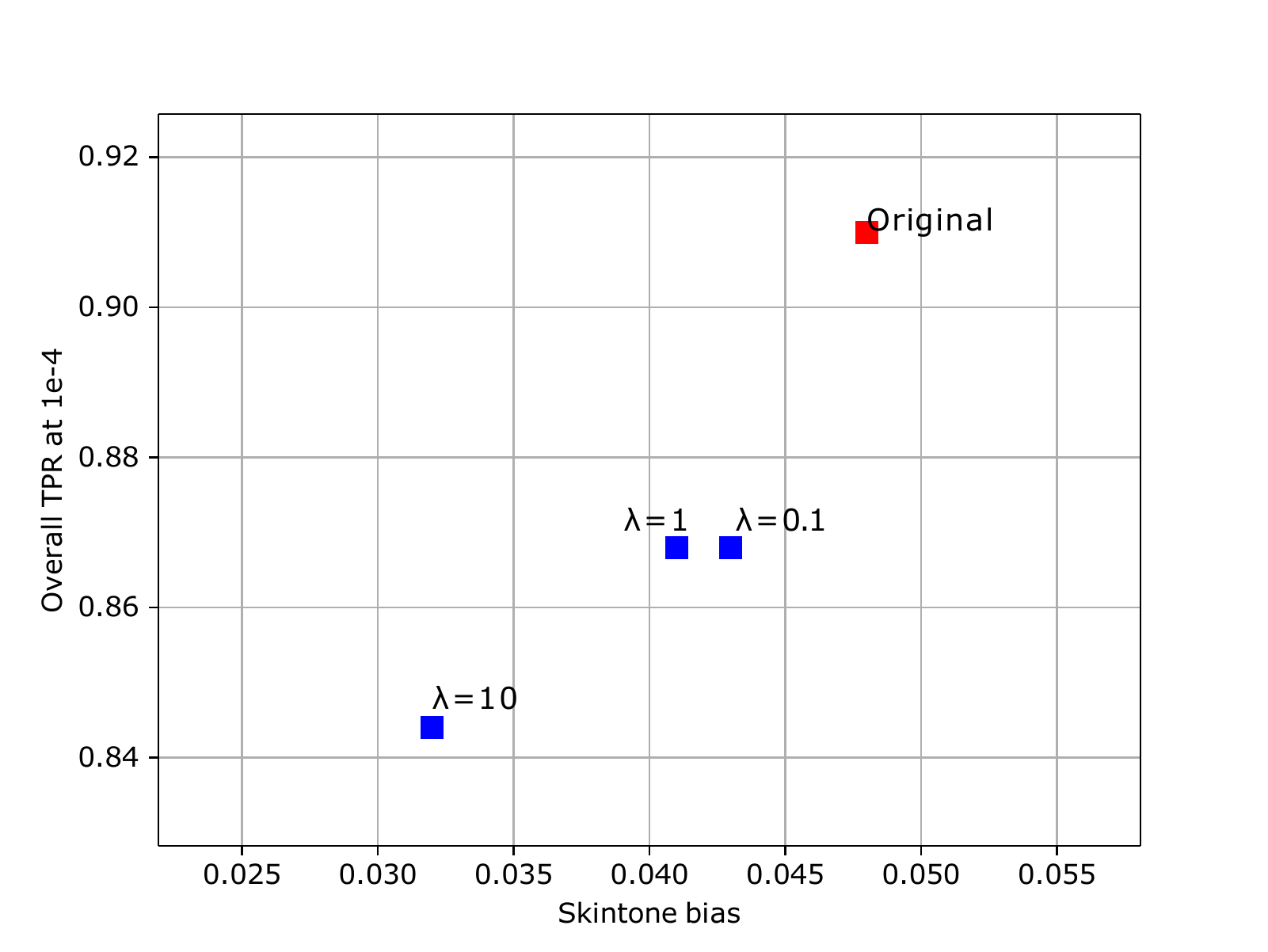}\label{fig:lamcryst}}
\caption{\small Effect of varying $\lambda$ (weight for $L_{deb}$) in PASS systems}
\label{fig:ablam}
}
\end{figure}

In Eq. 11 of the main paper, we combined a classification loss $L_{class}$ and an adversarial de-biasing loss $L_{deb}$ to compute a bias reducing classification loss $L_{br}$ as follows:
\begin{equation}
L_{br} = L_{class}+\lambda L_{deb}
\label{eq:lbrsupp}
\end{equation}
$L_{deb}$ is computed using an ensemble of $K$ attribute classifiers that act as adversaries to model $M$. $\lambda$ is the weight applied on this de-biasing loss. Here, we evaluate two hyperparameters used to train the PASS framework : (a) the number of attribute classifiers $K$ in the ensemble $E$ used to compute $L_{deb}$ (Eq. 10 in main paper). (b) the weight $\lambda$ for $L_{deb}$ defined in Eq. \ref{eq:lbrsupp} here. We analyze how changing these hyperparameters in PASS-g and PASS-s systems vary the resultant gender bias reduction and verification performance at a fixed FPR in the IJB-C dataset. We perform these experiments on PASS-g and PASS-s trained on both Arcface and Crystalface descriptors. For evaluating the PASS-g systems, we report the gender bias and verification TPR at FPR=$10^{-5}$. For evaluating PASS-s systems, we report the skintone bias and verification TPR at FPR=$10^{-4}$. (See Fig. \ref{fig:abk} and \ref{fig:ablam})  \\

\textbf{Varying K (number of adversary classifier in the ensemble)} : We experiment with $K=2,3,4$ and $10$, while fixing all the other parameters. The ablation results for PASS-g systems are presented in Figures \ref{fig:karcg} (for Arcface) and \ref{fig:kcryg} (for Crystalface). The results for PASS-s systems trained on Arcface descriptors are presented in Figure  \ref{fig:karcst} and those for Crystalface descriptors are presented in Figure \ref{fig:kcryst}. We find that for both PASS-s and PASS-g systems, increasing $K$ generally lowers the corresponding bias but also reduces the verification performance.\\
\begin{figure}
{
\centering
\subfloat[]{\includegraphics[width=0.5\linewidth, height=3.15cm]{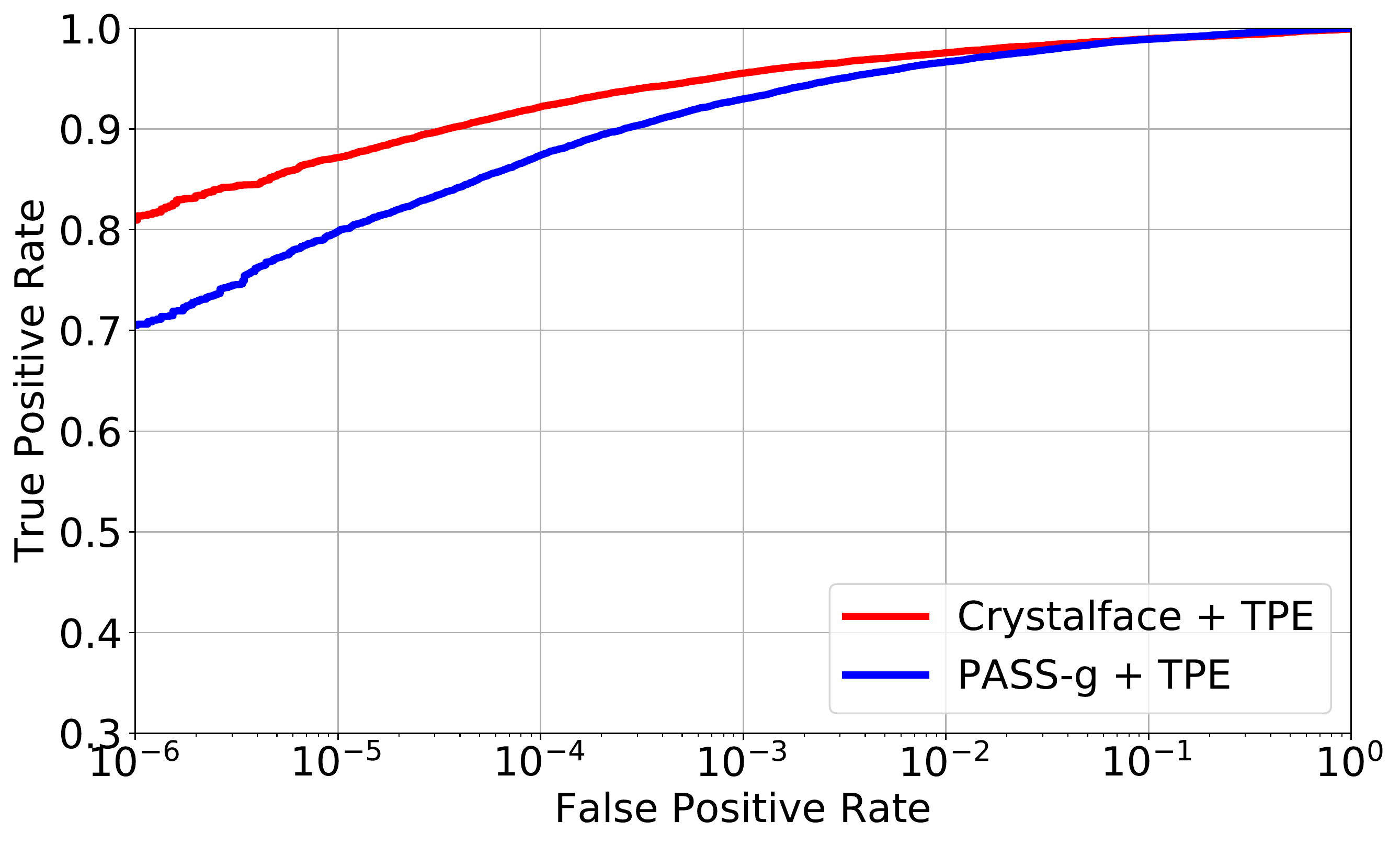}}
\subfloat[]{\includegraphics[width=0.5\linewidth]{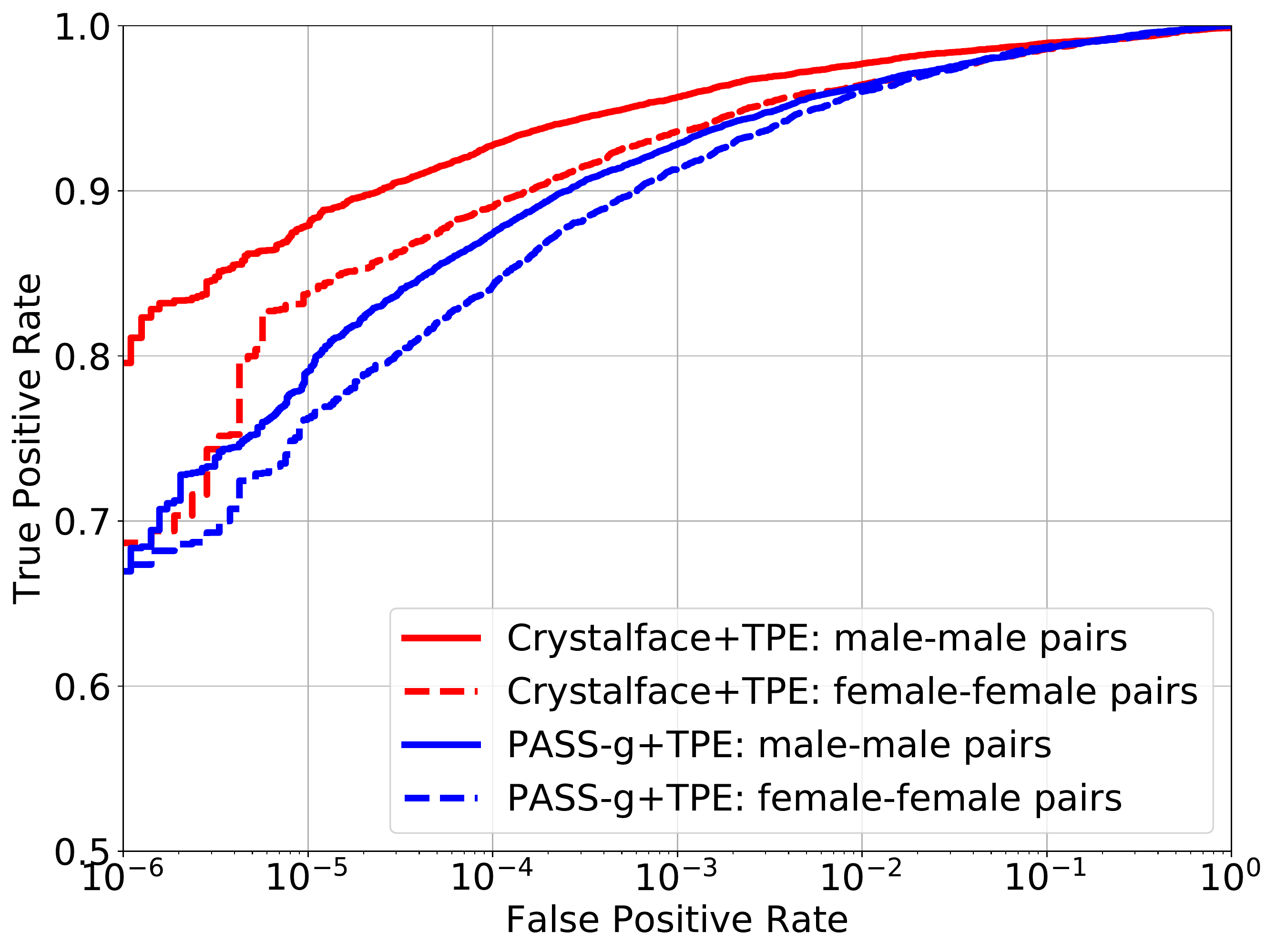}}\\
\subfloat[]{\includegraphics[width=0.5\linewidth]{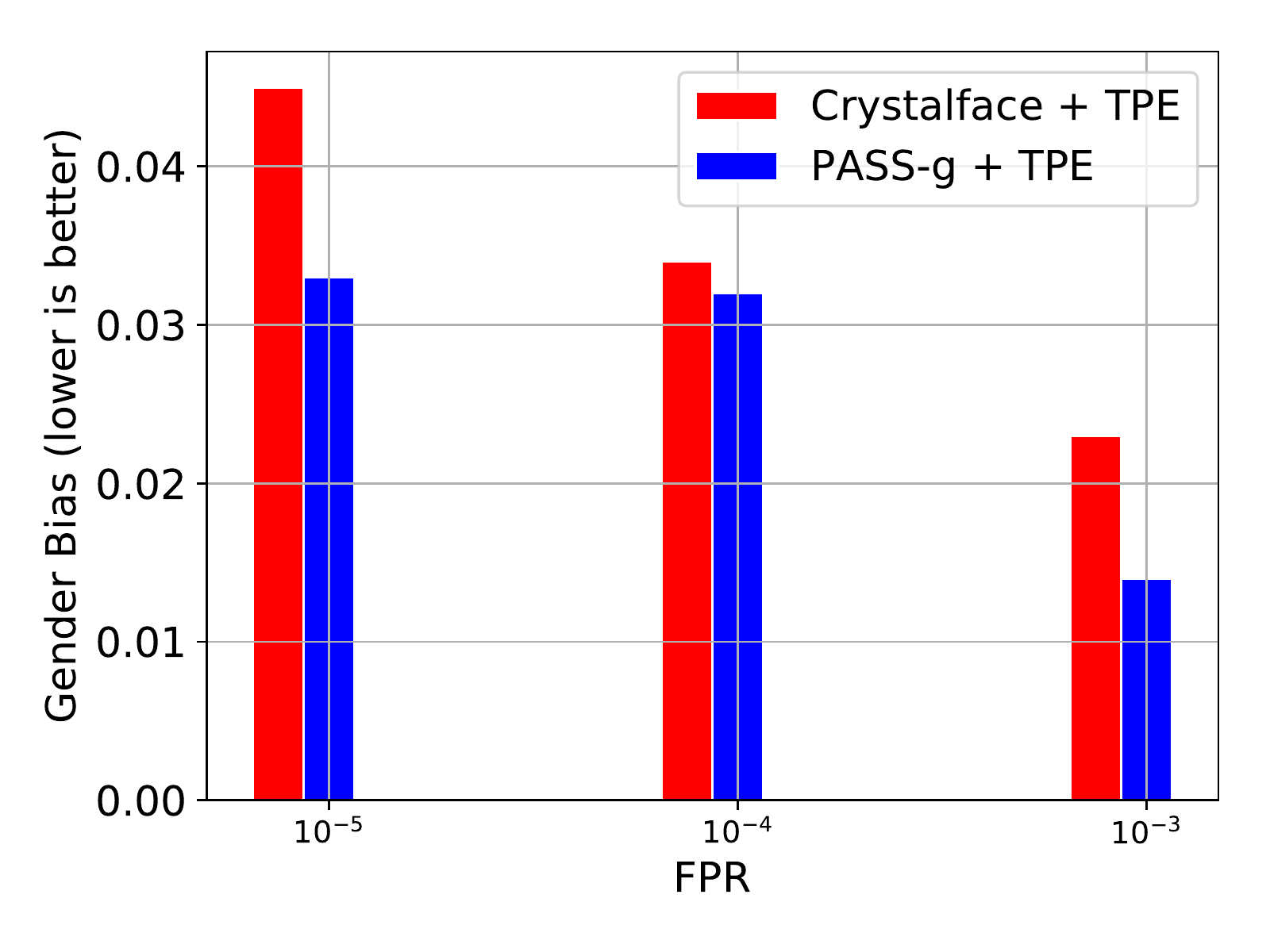}}
\caption{\small (a.) Overall IJB-C verification plots, (b.) Gender-wise IJB-C verification plots, (c.) Associated gender bias for Crystalface descriptors and its PASS-g counterpart after applying TPE}
\label{fig:tpe}
}
\end{figure}
\begin{figure*}
{
\centering
\subfloat[]{\includegraphics[width=0.333\linewidth]{img/teaser_small.pdf}\label{fig:teasersupp}}
\subfloat[]{\includegraphics[width=0.333\linewidth]{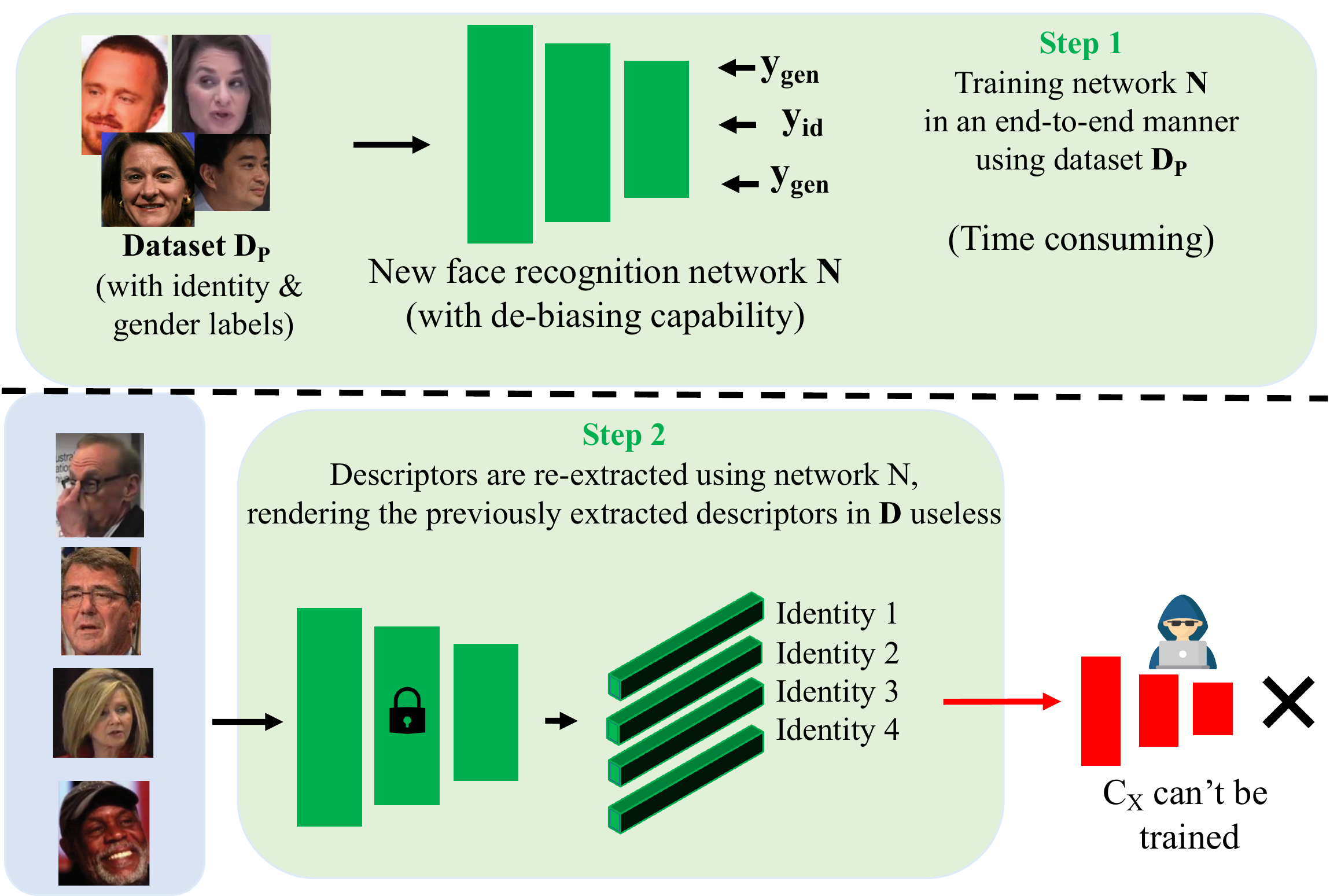}\label{fig:e2e}}
\subfloat[]{\includegraphics[width=0.333\linewidth]{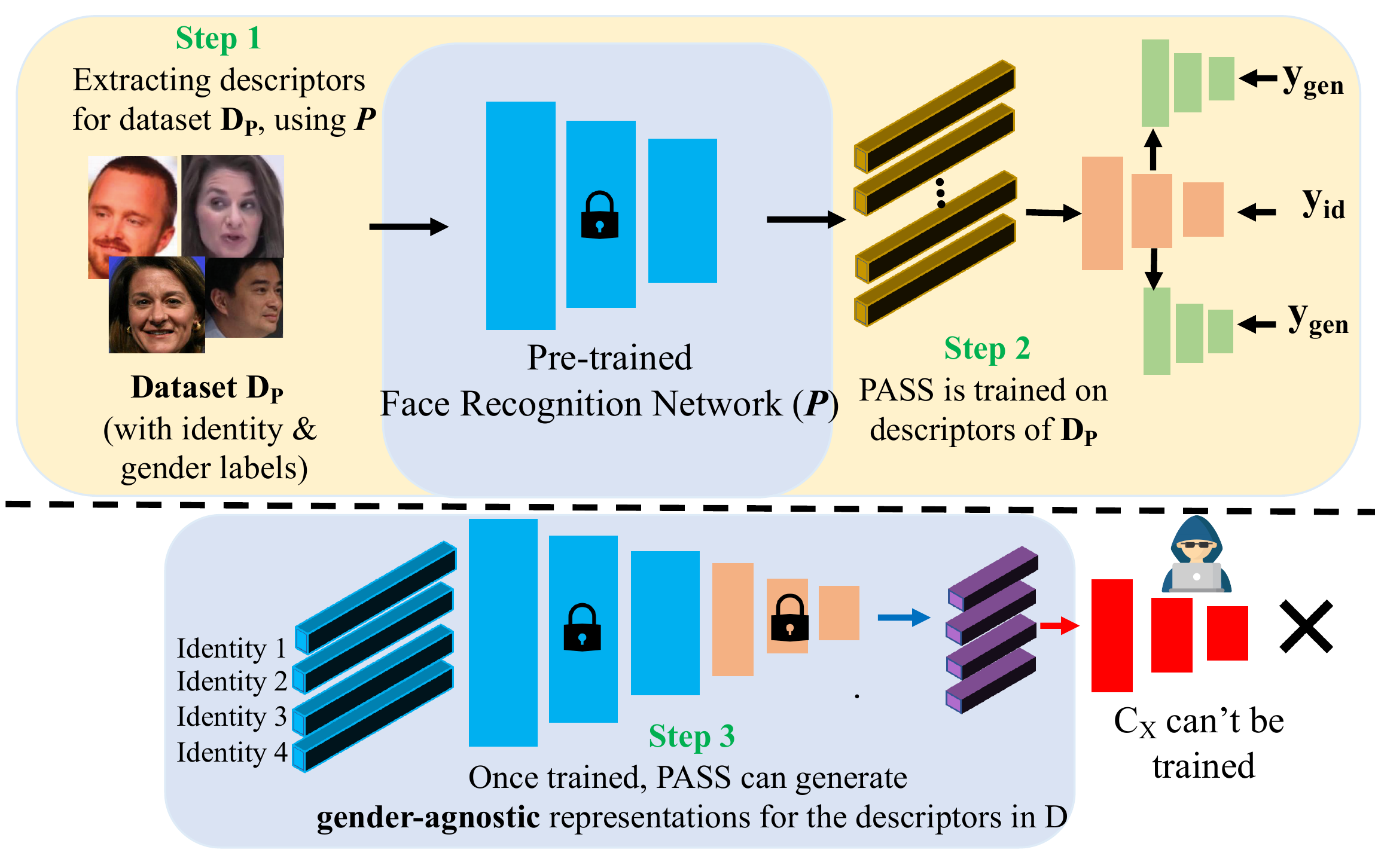}\label{fig:reuse}}

\caption{\small (a.) Example of a scenario where an agent $C_X$ can cause privacy breach in a private database $D$ that contains a pre-trained face recognition network $P$ and face descriptors of four identities extracted using $P$. (b.) Training an end-to-end de-biasing system does not allow us to re-use the pre-computed descriptors in $D$. (c) PASS can be train on top of descriptors from $P$ and can re-use the pre-computed descriptors in $D$ to generate their gender-agnostic representations. }
\label{fig:passadv}
}
\end{figure*}

\textbf{Varying $\lambda$ (weight for $L_{deb}$):} We experiment with $\lambda = 0.1, 1, 10$ for training the PASS-s framework on Arcface and Crystalface descriptors. All the other hyperparameters remain fixed. The results are presented in Fig. \ref{fig:ablam}. For both PASS-g and PASS-s systems, we find that as we keep on increasing the value of $\lambda$, the associated bias generally decreases and the verification TPR keeps decreasing.

\section{Additional experiment: Effect of TPE}
\label{sec:tpe}
In \cite{ranjan2019fast}, the face descriptors from Crystalface are not directly used for verification. Instead, the descriptors undergo triplet probabilistic embedding (TPE) \cite{Swami_2016_triplet} for generating a template representation of a given identity. TPE is an embedding learned to generate more discriminative, low-dimensional representations of given input descriptors, that have been shown to achieve better verification results. We apply TPE on the descriptors obtained using Crystalface and find that \textit{TPE improves the overall verification performance, but it also increases gender bias at all FPRs} (`Crystalface + TPE' in Table \ref{tab:tpe}). We analyze if applying TPE on PASS-g descriptors has the same effect. We learn a TPE matrix using Crystalface descriptors transformed with PASS-g. We apply this TPE matrix to transform the PASS-g descriptors extracted for the test (IJB-C) dataset, the results for which are presented in Table \ref{tab:tpe} (`PASS-g + TPE'). From Table \ref{tab:tpe} and Figure \ref{fig:tpe}, we can infer that the g\textit{ender bias in the verification results obtained after applying TPE on PASS-g transformed descriptors is lower than when TPE is applied on original face descriptors of Crystalface.}\\

To learn a triplet probabilistic embedding $W_{cf}$, we use the descriptors from Crystalface (extracted for UMD-Faces \cite{bansal2017umdfaces} dataset). This embedding $W_{cf} \in \mathbb{R}^{512\times128}$ is then used to transform the 512 dimensional IJB-C \cite{maze2018iarpa} descriptors (extracted using Crystalface) to obtain 128-dimensional face descriptors, which are used for 1:1 face verification. The results of this experiment are provided in `Crystalface + TPE' in Table \ref{tab:tpe}. We perform the same experiment with the PASS-g transformed descriptors of Crystalface, where a new TPE matrix  $W'_{cf} \in \mathbb{R}^{256\times128}$ is learned and used to transform the IJB-C descriptors before performing 1:1 verification.\\

For training both, $W_{cf}$ and $W'_{cf}$, we use a fixed learning rate of $2.5 \times 10^{-3}$ and a batch size of 32. The training for computing such a matrix using the descriptors from Crystalface (or its PASS-g counterpart) generally converges after 10k iterations.  For a given set of descriptors, we compute its TPE matrix ten times and finally compute the  average of the resulting matrices. We use this matrix to transform the test descriptors. More details about TPE are provided in \cite{Swami_2016_triplet}.

Note that, unlike Crystalface \cite{ranjan2019fast}, Arcface \cite{deng2018arcface} does not mention applying TPE on the face descriptors and therefore we do not apply TPE on PASS-based systems that are trained on Arcface.

\section{Advantages of PASS over end-to-end systems}
\label{sec:advpass}
In section 5.4.3 of the main paper, we explained how PASS/MultiPASS systems outperform end-to-end bias mitigation methods like \cite{gong2020jointly} and \cite{gac} in terms of overall face verification performance. Apart from this, the PASS/MultiPASS system is easier to deploy than end-to-end pipelines. 
\begin{table}%
  \scriptsize
  \centering
\begin{tabular}{cccc}
\toprule
Method & Training & Backbone &  \#Params w/o final classif\textsuperscript{n} layer\\
\midrule
Debface-ID\cite{gong2020jointly}&End-to-end&ResNet-52&10.99 million\\
Demo-ID\cite{gong2020jointly}&End-to-end&ResNet-52&10.99 million\\
GAC\cite{gac}&End-to-end&ResNet-52&10.99 million\\
\midrule
PASS-g w/ AF& Descriptor-based & MLP & 254,336\\
PASS-s w/ AF& Descriptor-based & MLP & 213,504\\
MultiPASS w/ AF& Descriptor-based & MLP & 336,768\\
PASS-g w/ CF& Descriptor-based & MLP & 295,424\\
PASS-s w/ CF& Descriptor-based & MLP & 213,504\\
MultiPASS w/ CF& Descriptor-based & MLP & 377,856\\
\bottomrule
\vspace{-6pt}
\end{tabular}
\caption{\small Number of trainable parameters in end-to-end and PASS-based methods. AF=Arcface, CF=Crystalface}
  \label{tab:params}
  \vspace{-11pt}
\end{table}

\textbf{Training time}: Most end-to-end bias-mitigation techniques (\cite{gong2020jointly} and \cite{gac}) use a ResNet architecture, for this reason training such frameworks likely takes a long time. In contrast, our descriptor-based PASS/MultiPASS systems (which are composed of MLPs) have fewer trainable parameters. In Table \ref{tab:params}, we compare the number of trainable parameters (excluding the final identity classification layer) of PASS-based systems and other end-to-end debiasing approaches. Since PASS/MultiPASS systems have fewer trainable parameters, the training is relatively fast.

Note that we recognize that convolution layers and linear layers differ in number of floating-point operations per weight, however, we use number of weights here as a rough proxy for computation time.

\textbf{Re-using pre-computed descriptors: }We go back to the example scenario described in Fig 1 of the main paper (and here in Fig \ref{fig:teasersupp}). Suppose a malicious agent $X$ has gained access to a private database $D$ (blue) which consists of a pre-trained network $P$ and face descriptors of four identities. The agent can use $P$ to extract descriptors (red) for a gender-labeled dataset $D_X$ (Step 1). Using these descriptors, the agent can train a gender classifier $C_X$ (Step 2). Using the trained $C_X$, the agent can predict the gender of the descriptors in $D$ (Step 3) and thus cause privacy breach.

Let's say we apply an end-to-end bias mitigation technique to prevent such privacy breach (Fig \ref{fig:e2e}). We first need to train a network $N$ on a dataset with identity and gender labels. This step is time consuming. Also, once $N$ is trained, we need to re-extract the face descriptors for the four identities using $N$. Thus, the pre-computed descriptors in $D$ cannot be re-used.

Instead, suppose that we deploy PASS-g for this task (Fig \ref{fig:reuse}). We can use the pre-trained network $P$ to first extract face descriptors for a dataset with identity and gender labels. Using these descriptors, we can train a PASS-g system. Once trained, PASS-g can be quickly applied to the pre-computed descriptors
to generate their gender agnostic representations. This re-use of existing descriptors is not possible using an end-to-end de-biasing system.
Thus, compared to end-to-end de-biasing methods, PASS allows easier deployment.
\section{A discussion about bias reduction and drop in verification performance}
\label{sec:tradeoff}
Although PASS/MultiPASS systems are trained to reduce sensitive information from face descriptors while maintaining their identity classification capability, it is clear from Figures \ref{fig:overallarcface} and \ref{fig:overallcrystalface} that reducing information of sensitive attributes in face descriptors leads to a slight drop in verification performance. This is not unexpected because attributes like gender and race/skintone are entangled with identity \cite{dhar2020attributes}, and are integral to it . Hence, reducing the information of such attributes is expected to slightly reduce the face descriptors' ability to classify identities. In fact, several works that reduce information of sensitive attributes demonstrate a drop in overall performance of the system. For instance, \cite{bortolato2020learning} proposes a method to suppress gender in face representations while performing the task of face recognition. Although this method successfully enhances gender privacy in the representations, it also leads to a slight drop in face recognition performance. Similarly, \cite{wu2018towards} proposes a method to perform activity recognition while reducing sensitive identity information. However, this leads to a slight drop in the target task of activity recognition. Also, \cite{8869910} proposes a GAN-based framework to generate a dataset that is fair (neutral) in terms of gender and skintone, while performing the target task of predicting attractiveness. While this method reduces the gender/skintone bias in attractiveness prediction, this also leads to a slight drop in the attractiveness prediction accuracy.

\end{document}